\documentclass{article}

\usepackage{fullpage}
\usepackage{latexsym}
\usepackage{amsmath,amssymb,stmaryrd,graphicx}
\usepackage{multirow}
\usepackage{color}
\usepackage{url}
\usepackage[round]{natbib}
\usepackage{ifthen}

\newcommand\sD{\ensuremath{\mathcal{D}}}

\newcommand\sO{\ensuremath{\mathcal{O}}}
\newcommand\sP{\ensuremath{\mathcal{P}}}

\newcommand\sR{\ensuremath{\mathcal{R}}}

\newcommand\sT{\ensuremath{\mathcal{T}}}

\newcommand\sV{\ensuremath{\mathcal{V}}}

\newcommand\sZ{\ensuremath{\mathcal{Z}}}
\newcommand\ba{\ensuremath{\mathbf{a}}}

\newcommand\be{\ensuremath{\mathbf{e}}}

\newcommand\bi{\ensuremath{\mathbf{i}}}
\newcommand\bj{\ensuremath{\mathbf{j}}}

\newcommand\bq{\ensuremath{\mathbf{q}}}

\newcommand\bs{\ensuremath{\mathbf{s}}}

\newcommand\bx{\ensuremath{\mathbf{x}}}

\newcommand\bA{\ensuremath{\mathbf{A}}}

\newcommand\bM{\ensuremath{\mathbf{M}}}

\newcommand\bX{\ensuremath{\mathbf{X}}}

\newcommand\T{\text}

\newcommand\Fig[4]{\begin{figure}[ht] \begin{center} \includegraphics[scale=#2]{#1} \end{center} \caption{\label{fig:#3} #4} \end{figure}}
\newcommand\FigTop[4]{\begin{figure}[t] \begin{center} \includegraphics[scale=#2]{#1} \end{center} \caption{\label{fig:#3} #4} \end{figure}}
\newcommand\FigStar[4]{\begin{figure*}[ht] \begin{center} \includegraphics[scale=#2]{#1} \end{center} \caption{\label{fig:#3} #4} \end{figure*}}
\newcommand\aside[1]{\quad\text{[#1]}}

\newcommand\argmin{\mathop{\text{argmin}}}
\newcommand\argmax{\mathop{\text{argmax}}}
\newcommand\p[1]{\ensuremath{\left( #1 \right)}} % Parenthesis ()
\newcommand\pa[1]{\ensuremath{\left\langle #1 \right\rangle}} % {}
\newcommand\pb[1]{\ensuremath{\left[ #1 \right]}} % []
\newcommand\pc[1]{\ensuremath{\left\{ #1 \right\}}} % {}
\newcommand\half{\ensuremath{\frac{1}{2}}}
\newcommand\R{\ensuremath{\mathbb{R}}} % Real numbers
\newcommand\eqdef{\ensuremath{\stackrel{\rm def}{=}}} % Equal by definition
\newcommand\refeqn[1]{(\ref{eqn:#1})}
\newcommand\refeqns[2]{(\ref{eqn:#1}) and (\ref{eqn:#2})}
\newcommand\refchp[1]{Chapter~\ref{chp:#1}}
\newcommand\refsec[1]{Section~\ref{sec:#1}}

\newcommand\reffig[1]{Figure~\ref{fig:#1}}

\newcommand\reftab[1]{Table~\ref{tab:#1}}
\newcommand\refapp[1]{Appendix~\ref{sec:#1}}

\newcommand\Section[2]{\section{#2}\label{sec:#1}}
\newcommand\Subsection[2]{\subsection{#2}\label{sec:#1}}
\newcommand\Subsubsection[2]{\subsubsection{#2}\label{sec:#1}}
\ifthenelse{\isundefined{\definition}}{\newtheorem{definition}{Definition}}{}
\ifthenelse{\isundefined{\assumption}}{\newtheorem{assumption}{Assumption}}{}
\ifthenelse{\isundefined{\proposition}}{\newtheorem{proposition}{Proposition}}{}
\ifthenelse{\isundefined{\theorem}}{\newtheorem{theorem}{Theorem}}{}
\ifthenelse{\isundefined{\lemma}}{\newtheorem{lemma}{Lemma}}{}
\ifthenelse{\isundefined{\corollary}}{\newtheorem{corollary}{Corollary}}{}
\newcommand\E{\ensuremath{\mathbb{E}}} % Expectation
\newcommand\theTitle{Learning Dependency-Based Compositional Semantics}

\newcommand\ind{\textcolor{white}{$-$}}

\newcommand\store{\sigma}
\newcommand\bstore{{\mathbf\sigma}}

\newcommand\Abstract{\alpha}

\newcommand\den[1]{{\llbracket #1 \rrbracket}_w} % Denotation
\newcommand\denabs[1]{{\llbracket #1 \rrbracket}_{\Abstract(w)}} % Denotation

\newcommand\geo{\text{\sc Geo}}
\newcommand\job{\text{\sc Jobs}}

\newcommand\nl[1]{\T{\it #1}} % Language
\newcommand\wl[1]{\text{\tt #1}} % World
\newcommand\ts[1]{\text{\sc #1}}

\newcommand\bolde{\mathbf{e}}

\newcommand\SigmaRel{\Sigma} % Summation
\newcommand\X{\ts{x}} % Execute
\newcommand\ExtractRel{\ts{e}} % Extraction
\newcommand\CompareRel{\ts{c}} % Superlatives/comparatives
\newcommand\QuantRel{\ts{q}} % Quantification/negation
\newcommand\nil{\text{\o}}
\newcommand\nilthree{\text{\o}}
\newcommand\arr[1]{[#1]}
\newcommand\syn[1]{\pa{#1}}
\newcommand\sem[1]{\langle\!\langle #1 \rangle\!\rangle}
\newcommand\densyn[1]{\den{\syn{#1}}}
\newcommand\dtrue{d_{\ts{t}}}
\newcommand\dfalse{d_{\ts{f}}}

\newcommand\JoinProjOp[3]{#2 \bowtie_{#1}^{-\nil} #3}
\newcommand\joinOp[1]{\bowtie_{#1}}
\newcommand\joinProjOp[1]{\bowtie_{#1}^{-\nil}}
\newcommand\ProjOp[2]{#2[#1]}
\newcommand\SumOp[1]{\Sigma\p{#1}}
\newcommand\ConcatOp[2]{+_{#1}\p{#2}}

\newcommand\with[1]{\{#1\}}

\newcommand\state{\wl{state}}

\newcommand\Count{\wl{count}}
\newcommand\Argmin{\wl{argmin}}
\newcommand\Argmax{\wl{argmax}}
\newcommand\average{\wl{average}}

\newcommand\entity{\star}
\newcommand\sTtuple{\sT_\ts{tuple}}

\newcommand\sVsettuple{\sV_{\{\ts{tuple}\}}}

\newcommand\Mark{\bM}

\newcommand\cand{T}
\newcommand\lcand{T^\swarrow}
\newcommand\rcand{T^\searrow}
\newcommand\descrip[1]{#1}

\newcommand\JoinRel[2]{\ensuremath{\substack{#1 \\ #2}}}
\newcommand\cto{\!:\!}
\newcommand\arity[1]{\ts{Arity}(#1)}
\newcommand\arityw[1]{\ts{Arity}_w(#1)}

\newcommand\PredCount{\ts{PredHit}}
\newcommand\Pred{\ts{Pred}}
\newcommand\PredRel{\ts{PredRel}}
\newcommand\PredPred{\ts{PredRelPred}}
\newcommand\WordTriggerPred{\ts{TriggerPred}}
\newcommand\WordTraceAll{\ts{Trace}*}
\newcommand\WordTracePred{\ts{TracePred}}
\newcommand\WordTraceRel{\ts{TraceRel}}
\newcommand\WordTracePredRel{\ts{TracePredRel}}

\newcommand\Lb{L_\ts{b}}
\newcommand\Lbp{L_{\ts{b}+\ts{p}}}

\newcommand\LogZ{\bA}

\newif\ifjournal
\journaltrue

\newcommand\paper{article}
\newcommand\chp{section}
\newcommand\SecOne\Section
\newcommand\SecTwo\Subsection
\newcommand\SecThree\Subsubsection
\newcommand\SecFour[2]{\paragraph{#2}}
\renewcommand\refchp\refsec

\begin{document}

\title{\theTitle}
%\author{Percy Liang \\ UC Berkeley}
%\author{Michael I. Jordan \\ UC Berkeley}
%\author{Dan Klein \\ UC Berkeley}
\author{Percy Liang \\
  UC Berkeley \\
  pliang@cs.berkeley.edu  \and
  Michael I. Jordan \\
  UC Berkeley \\
  jordan@cs.berkeley.edu \and
  Dan Klein \\
  UC Berkeley \\
  klein@cs.berkeley.edu }
\maketitle
\begin{abstract}
Suppose we want to build a system that answers a natural language question
by representing its semantics as a logical form and computing the answer 
given a structured database of facts. The core part of such a system is 
the semantic parser that maps questions to logical forms.  Semantic parsers 
are typically trained from examples of questions annotated with their target 
logical forms, but this type of annotation is expensive.

Our goal is to learn a semantic parser from question-answer pairs instead,
where the logical form is modeled as a latent variable.  Motivated by this 
challenging learning problem, we develop a new semantic formalism,
dependency-based compositional semantics (DCS), which has favorable 
linguistic, statistical, and computational properties.  We define a 
log-linear distribution over DCS logical forms and estimate the parameters 
using a simple procedure that alternates between beam search and numerical 
optimization.  On two standard semantic parsing benchmarks, our system 
outperforms all existing state-of-the-art systems, despite using no 
annotated logical forms.
\end{abstract}

\SecOne{introduction}{Introduction}

% Problem statement
% Percy
%A major challenge in NLP is building systems
%to understand language with increasing degrees of semantic sophistication
%but with decreasing amounts of human supervision.
%Towards this goal, we focus on the concrete task of building a system to answer
%questions given a structured database of facts.
%As a running example, consider the domain of US geography (\reffig{introExample}).
%We chose this task because it is a practical application,
%requires non-trivial modeling of semantics,
%and has a clear evaluation metric.

% Mike
%One of the major challenges in many areas of NLP is to develop models
%that exhibit full coverage of linguistic phenomena of interest but which 
%require minimal human annotation.  This tension is particularly evident in 
%semantics, where there seems to be little alternative to making significant
%usage of human annotation in developing models but where the effort needed 
%to obtain broad coverage in real-world domains is generally prohibitive.  
%In this \paper we tackle this problem, developing a question-answering 
%system that represents questions as logical forms and obtains answers
%by evaluating these logical forms against a structured database but which 
%requires no human annotation of logical forms.  Instead, our system learns 
%its semantic representations from a training corpus of question-answer 
%pairs, inferring logical forms as latent structures.

% Abstract - vague
One of the major challenges in NLP is building systems
that both handle complex linguistic phenomena and require minimal human effort.
The difficulty of achieving both criteria
is particularly evident in training semantic parsers,
where annotating linguistic expressions with their associated logical forms
is expensive but seemingly unavoidable.
In this \paper, we overcome these limitations
by developing new techniques that can learn rich semantic representations from weak supervision.

\Fig{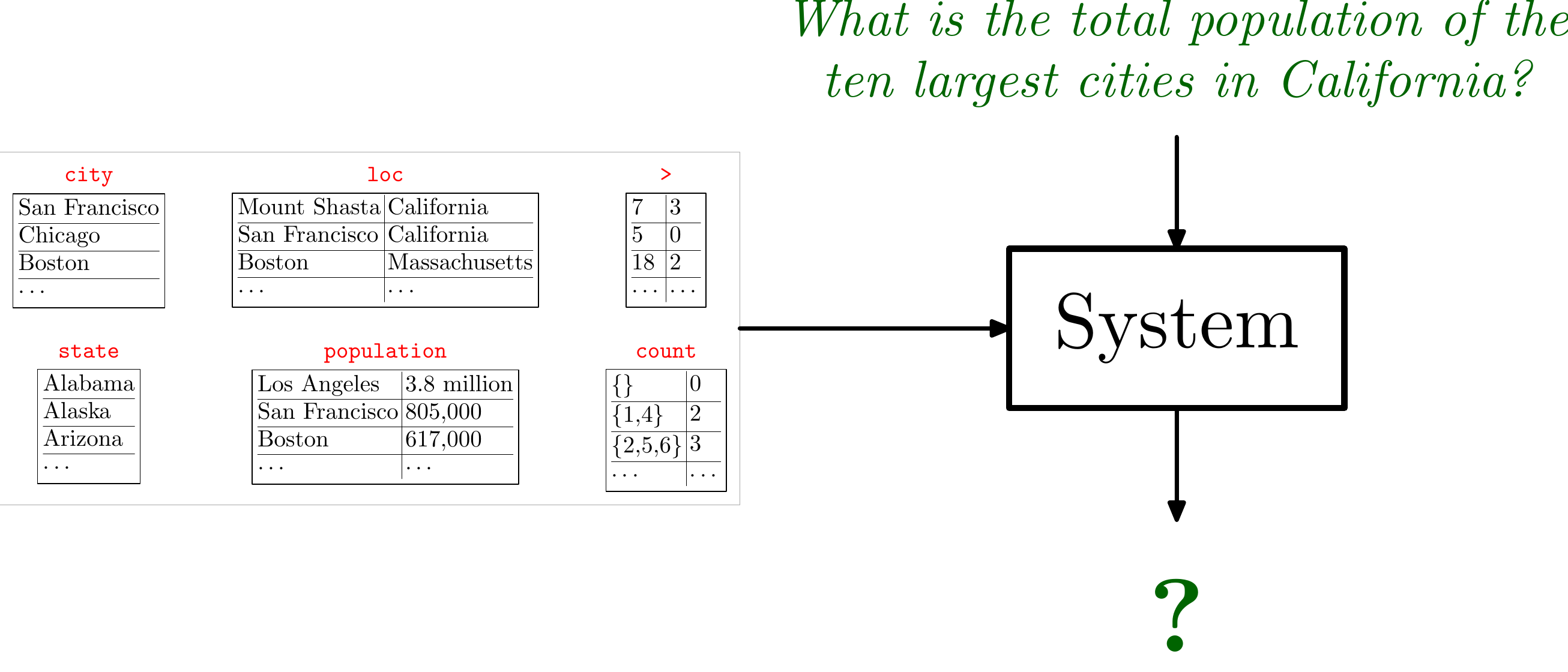}{0.35}{introExample}{
The concrete objective: a system that answers natural language questions
given a structured database of facts.
An example is shown in the domain of US geography.
}

% Problem statement, NLIDB history: rule-based -> statistical-based
We demonstrate our techniques on the concrete task of building a system to answer
questions given a structured database of facts--see \reffig{introExample} for
an example in the domain of US geography.
The problem of building {\em natural language interfaces to databases} (NLIDBs)
has a long history in NLP, starting from the early days of AI with
systems such as {\sc Lunar} \citep{woods72lunar}, {\sc Chat-80} \citep{warren82chat80}, and many others
(see \citet{androutsopoulos95nlidb} for an overview).
While quite successful in their respective limited domains,
because these systems were constructed from manually-built rules, they became difficult to scale up,
both to other domains and to more complex utterances.
In response, against the backdrop of a statistical revolution in NLP during the 1990s, researchers
began to build systems that could learn from examples, with the hope of overcoming the limitations
of rule-based methods.
One of the earliest statistical efforts was the {\sc Chill} system \citep{zelle96geoquery},
which learned a shift-reduce semantic parser.
Since then, there has been a healthy line of work yielding increasingly more accurate semantic parsers
by using new semantic representations and machine learning techniques
\citep{zelle96geoquery,miller96statistical,tang01ilp,ge05scissor,kate05funql,zettlemoyer05ccg,kate06krisp,wong06mt,kate07krisper,zettlemoyer07relaxed,wong07synchronous,kwiatkowski10ccg,kwiatkowski11lex}.

% Problem: annotation costs
However, while statistical methods provided advantages such as robustness
and portability, their application in semantic parsing achieved only limited
success.  One of the main obstacles was that these methods depended crucially on having
examples of utterances paired with logical forms,
and this requires substantial human effort to obtain.
Furthermore, the annotators must be proficient in some formal language, which drastically
reduces the size of the annotator pool, dampening any hope
of acquiring enough data to fulfill the vision of learning highly accurate systems.

% Solution: reduce supervision: latent logical form
In response to these concerns, researchers have recently
begun to explore the possibility of learning a semantic parser without any annotated logical forms
\citep{clarke10world,liang11dcs,goldwasser11confidence,artzi11conversations}.
It is in this vein that we develop our present work.
Specifically, given a
set of $(\bx,y)$ example pairs, where $\bx$ is an utterance (e.g., a question)
and $y$ is the corresponding answer, we wish to learn a mapping from $\bx$ to $y$.
What makes this mapping particularly interesting is it passes through a
latent logical form $z$, which is necessary to capture the semantic
complexities of natural language.  Also note that while the logical form $z$ was the end goal in past
work on semantic parsing, for us, it is just an intermediate variable---a means towards an end.
\reffig{model} shows the graphical model which captures the learning setting we
just described:~The question $\bx$, answer $y$, and world/database $w$ are all
observed.  We want to infer the logical forms $z$ and the parameters
$\theta$ of the semantic parser, which are unknown quantities.

\FigTop{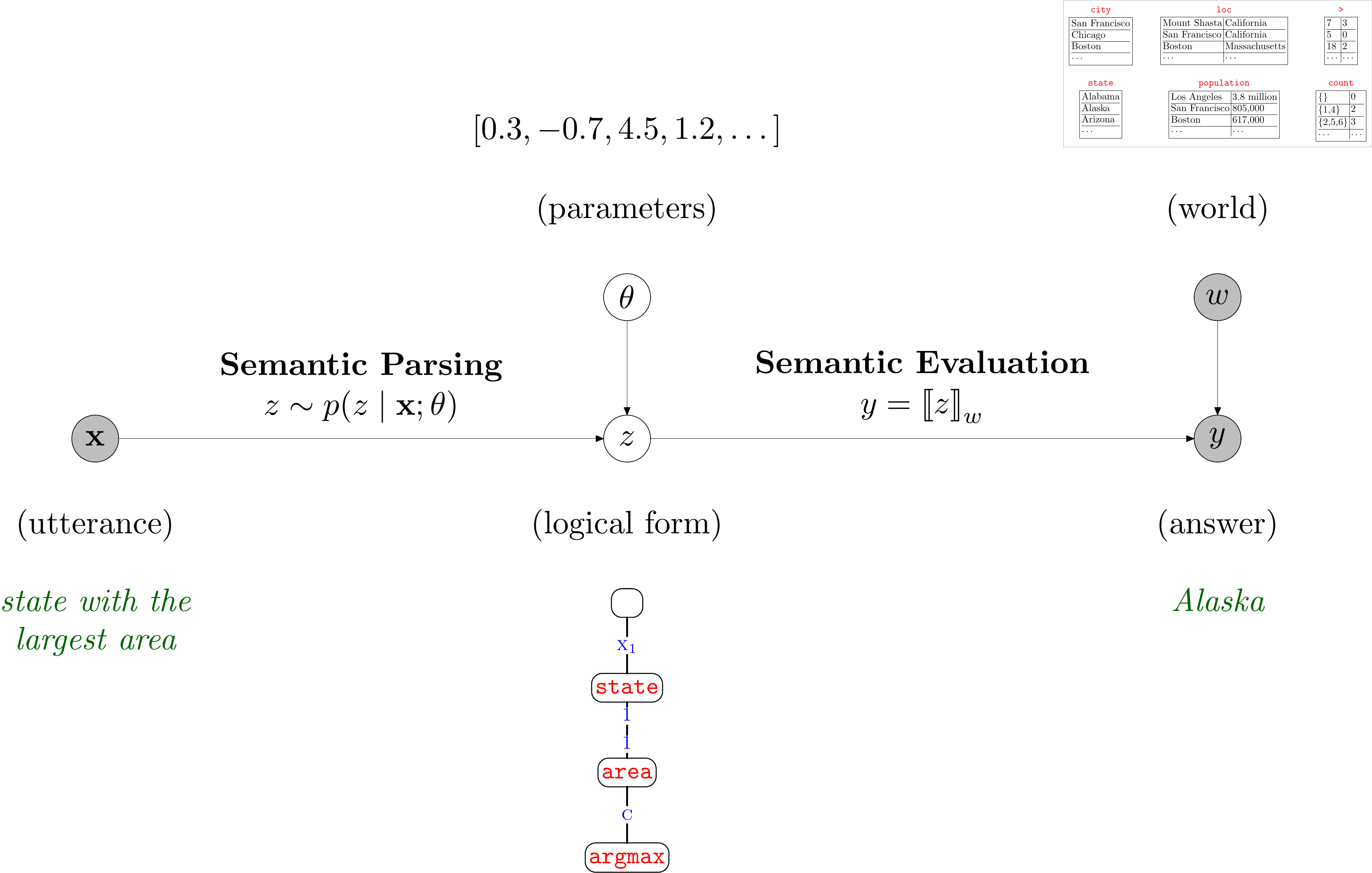}{0.35}{model}{
Our statistical methodology consists of two steps:
(i) semantic parsing: an utterance $\bx$ is mapped to a logical form $z$ by drawing from a log-linear distribution parametrized by
a vector $\theta$;
and (ii) evaluation: the logical form $z$ is evaluated with respect to the
world $w$ (database of facts) to deterministically produce an answer $y = \den{z}$.
The figure also shows an example configuration of the variables around the graphical model.
Logical forms $z$ as represented as labeled trees.
During learning, we are given $w$ and $(\bx,y)$ pairs (shaded nodes)
and try to infer the latent logical forms $z$ and parameters $\theta$.
}

% Challenge -> representation?
While liberating ourselves from annotated logical forms reduces cost, it does
increase the difficulty of the learning problem.  The core challenge here is
{\em program induction}: on each example $(\bx,y)$, we need to efficiently
search over the exponential space of possible logical forms $z$ and find ones
that produces the target answer $y$, a computationally daunting task.
There is also a statistical challenge: how do we parametrize the mapping from
utterance $\bx$ to logical form $z$ so that it can be learned from
only the indirect signal $y$?  To address these two challenges, we must first discuss
the issue of {\em semantic representation}.  There are two basic questions here: (i) what
should the formal language for the logical forms $z$ be, and (ii) what are the
compositional mechanisms for constructing those logical forms?  

% Formal language, construction mechanism, scoring
The semantic parsing literature is quite multilingual with respect to the
formal language used for the logical form:~Researchers have
used SQL \citep{giordani09sql},
Prolog \citep{zelle96geoquery,tang01ilp},
a simple functional query language called FunQL \citep{kate05funql},
and lambda calculus \citep{zettlemoyer05ccg}, just to name a few.
The construction mechanisms are equally diverse, including
synchronous grammars \citep{wong07synchronous}, hybrid trees \citep{lu08generative},
Combinatorial Categorial Grammars (CCG) \citep{zettlemoyer05ccg},
and shift-reduce derivations \citep{zelle96geoquery}.
It is worth pointing out that the choice of formal language and the construction mechanism
are decisions which are really more orthogonal than is often assumed---the former 
is concerned with what the logical forms look like; the latter, with how to 
generate a set of possible logical forms compositionally given an utterance.  
(How to score these logical forms is yet another dimension.)
%left to the domain of the learning algorithm.)

% Weaknesses of exiting formal languages and construction mechanisms
Existing systems are rarely based on the joint design of the formal
language and the construction mechanism; one or the other is often
chosen for convenience from existing implementations.  For example, 
Prolog and SQL have often been chosen as formal languages for convenience
in end applications, but they were not designed for representing the semantics
of natural language, and, as a result, the construction mechanism that 
bridges the gap between natural language and formal language is generally
complex and difficult to learn.  CCG \citep{steedman00ccg} and more 
generally, categorial grammar, is the de facto standard in linguistics.  
In CCG, logical forms are constructed compositionally using a small 
handful of combinators (function application, function composition, and type raising).
For a wide range of canonical examples, CCG produces elegant, streamlined analyses,
but its success really depends on having a good, clean lexicon.  During 
learning, there is often large amounts of uncertainty over the lexical entries,
which makes CCG more cumbersome.  Furthermore, in real-world applications, 
we would like to handle disfluent utterances, and this further strains 
CCG by demanding either extra type-raising rules and disharmonic combinators 
\citep{zettlemoyer07relaxed} or a proliferation of redundant lexical entries 
for each word \citep{kwiatkowski10ccg}.

% New formalism: DCS
To cope with the challenging demands of program induction, we break away from
%the Montague tradition which has dominated formal semantics
tradition
in favor of a new formal language and construction mechanism, which we call {\em dependency-based
compositional semantics} (DCS).  The guiding principle behind DCS is to provide a
simple and intuitive framework for constructing and representing logical forms.
Logical forms in DCS are tree structures called \emph{DCS trees}.
The motivation is two-fold: (i) DCS trees are meant to parallel syntactic
dependency trees, which facilitates parsing; and (ii) a DCS tree essentially encodes a constraint
satisfaction problem, which can be solved efficiently using dynamic programming.
In addition, DCS provides a {\em mark-execute} construct,
which provides a uniform way of dealing with scope variation, 
a major source of trouble in any semantic formalism.
The construction mechanism in DCS is a generalization of labeled dependency
parsing, which leads to simple and natural algorithms.  To a linguist, DCS
might appear unorthodox, but it is important to keep in mind that our primary
goal is effective program induction, not necessarily to model new linguistic
phenomena in the tradition of formal semantics.

% Learning
Armed with our new semantic formalism, DCS, we then define a discriminative probabilistic
model, which is depicted in \reffig{model}.  The semantic parser is a log-linear
distribution over DCS trees $z$ given an utterance $\bx$.  Notably, $z$ is
unobserved, and we instead observe only the answer $y$, which is obtained by
evaluating $z$ on a
world/database $w$.  There are an exponential number of possible trees $z$,
and usually dynamic programming is employed for efficiently searching
over the space of these combinatorial objects.
However, in our case, we must enforce the global constraint that the tree generates
the correct answer $y$, which makes dynamic programming infeasible.  Therefore, we resort to beam search
and learn our model with a simple procedure which alternates between
beam search and optimizing a likelihood objective restricted to those beams.
This yields a natural bootstrapping procedure in which learning and search are
integrated.

% Experiments
We evaluated our DCS-based approach on two standard benchmarks,
\geo, a US geography domain \citep{zelle96geoquery} and \job, a job queries domain \citep{tang01ilp}.
On \geo, we found that our system significantly outperforms previous work that
also learns from answers instead of logical forms \citep{clarke10world}.
What is perhaps a more significant result is that 
our system even outperforms
state-of-the-art systems that do rely on annotated logical forms.
This demonstrates that the viability of training accurate systems
with much less supervision than before.

% Outline
The rest of this \paper\ is organized as follows:
\refchp{representation} introduces dependency-based compositional semantics
(DCS), our new semantic formalism.
\refchp{learning} presents our probabilistic model and learning algorithm.
\refchp{experiments} provides an empirical evaluation of our methods.
Finally, \refchp{discussion} situates this work in a broader context.

\SecOne{representation}{Representation}

In this \chp, we present the main conceptual contribution of this work,
dependency-based compositional semantics (DCS),
using the US geography domain \citep{zelle96geoquery} as a running example.
To do this, we need to define the syntax and semantics of the formal language.
The syntax is defined in \refsec{syntax} and is quite straightforward:~The
logical forms in the formal language are simply trees, which we call {\em DCS
trees}.
In \refsec{worlds},
we give a type-theoretic definition of {\em worlds} (also known as databases or models)
with respect to which we can define the semantics of DCS trees.

The semantics, which is the heart of this \paper, contains two main ideas:
(i) using trees to represent logical forms as constraint satisfaction problems or extensions thereof,
and (ii) dealing with cases when syntactic and semantic scope diverge
(e.g., for generalized quantification and superlative constructions)
using a new construct which we call {\em mark-execute}.
We start in \refsec{basic} by introducing the semantics of a basic version of DCS which focuses only on (i)
and then extend it to the full version (\refsec{full}) to account for (ii).

Finally, having fully specified the formal language,
we describe a construction mechanism for mapping a natural language utterance
to a set of candidate DCS trees (\refsec{construction}).
%\refchp{learning} will then focus on how to learn a distribution over these DCS trees.

%%%%%%%%%%%%%%%%%%%%%%%%%%%%%%
\SecTwo{notation}{Notation}

%Let us first establish some basic notation. 
Operations on tuples will play a prominent role in this \paper.
For a sequence\footnote{We use the {\em sequence} to include
both tuples $(v_1, \dots, v_k)$ and arrays $\arr{v_1, \dots, v_k}$.
For our purposes, there is no functional difference between tuples and arrays;
the distinction is convenient when we start to talk about arrays of tuples.
}
$v = (v_1, \dots, v_k)$, we use $|v| = k$ to denote the length of the sequence.
For two sequences $u$ and $v$, we use $u + v = (u_1, \dots, u_{|u|}, v_1,
\dots, v_{|v|})$ to denote their concatenation.

For a sequence of positive indices $\bi = (i_1, \dots, i_m)$, let $v_\bi = (v_{i_1}, \dots, v_{i_m})$
consist of the components of $v$ specified by $\bi$;
we call $v_\bi$ the projection of $v$ onto $\bi$.
We use negative indices to exclude components: $v_{-\bi} = (v_{(1,\dots,|v|) \backslash \bi})$.
We can also combine sequences of indices by concatenation:
$v_{\bi,\bj} = v_\bi + v_\bj$.
Some examples: if $v = (a, b, c, d)$,
then $v_2 = b$,
$v_{3,1} = (c,a)$,
$v_{-3} = (a,b,d)$,
$v_{3,-3} = (c,a,b,d)$.

%%%%%%%%%%%%%%%%%%%%%%%%%%%%%%
\SecTwo{syntax}{Syntax of DCS Trees}

% Syntax
The syntax of the DCS formal language is built from two ingredients, {\em predicates} and {\em relations}:
\begin{itemize}
\item Let $\sP$ be a set of {\em predicates}.
We assume that $\sP$ always contains a special null predicate $\nil$
and several domain-independent predicates
(e.g., $\wl{count}$, $\wl{<}$, $\wl{>}$, and $\wl{=}$).
In addition, $\sP$ contains domain-specific predicates.
For example, for the US geography domain, $\sP$ would include $\wl{state}$, $\wl{river}$, $\wl{border}$, etc.
Right now, think of predicates as just labels, which have yet to receive formal semantics.

\item Let $\sR$ be the set of {\em relations}.  The full set of relations are shown in \reftab{relations};
note that unlike the predicates $\sP$, the relations $\sR$ are fixed.
%Again, for now, relations are just labels.
\end{itemize}

% Define logical form
The logical forms in DCS are called DCS trees.
A DCS tree is a directed rooted tree in which nodes are labeled with predicates and edges are labeled with relations;
each node also maintains an ordering over its children.
Formally:
\begin{definition}[DCS trees]
\label{def:tree}
Let $\sZ$ be the set of DCS trees,
where each $z \in \sZ$ consists of
(i) a predicate $z.p \in \sP$
and (ii) a sequence of edges $z.\be = (z.e_1, \dots, z.e_m)$.
Each edge $e$ consists of a relation $e.r \in \sR$ (see \reftab{relations}) and a child tree $e.c \in \sZ$.
\end{definition}
We will either draw a DCS tree graphically
or write it compactly as $\syn{p; r_1:c_1; \dots; r_m:c_m}$
where $p$ is the predicate at the root node and $c_1, \dots, c_m$ are its $m$ children connected via edges
labeled with relations $r_1, \dots, r_m$, respectively.
\reffig{simpleExists}(a) shows an example of a DCS tree expressed using both graphical and compact formats.

% Relations
\begin{table}
\begin{center}
\begin{center} {\bf Relations $\sR$} \end{center}
\vspace{0.1in}
\begin{tabular}{lll}
Name                & Relation                                        & Description \\
\hline
\descrip{join}      & $\JoinRel{j}{j'}$ for $j,j' \in \{1,2,\dots\}$  & $j$-th component of parent $=$ $j'$-th component of child \\
\descrip{aggregate} & $\SigmaRel$                                     & parent $=$ set of feasible values of child \\
\hline
\descrip{extract}   & $\ExtractRel$                                   & mark node for extraction \\
\descrip{quantify}  & $\QuantRel$                                     & mark node for quantification, negation \\
\descrip{compare}   & $\CompareRel$                                   & mark node for superlatives, comparatives \\
\descrip{execute}   & $\X_\bi$ for $\bi \in \{1,2\dots\}^*$           & process marked nodes specified by $\bi$
\end{tabular}
\end{center}
\caption{\label{tab:relations}
Possible relations that appear on edges of DCS trees.
Basic DCS uses only the join and aggregate relations;
the full version of DCS uses all of them.
}
\end{table}

% Intuition
A DCS tree is a logical form, but it is designed to look like a
syntactic dependency tree, only with predicates in place of words.
As we'll see over the course of this \chp,
it is this transparency between syntax and semantics provided by DCS which leads to a simple and
streamlined compositional semantics suitable for program induction.

\FigTop{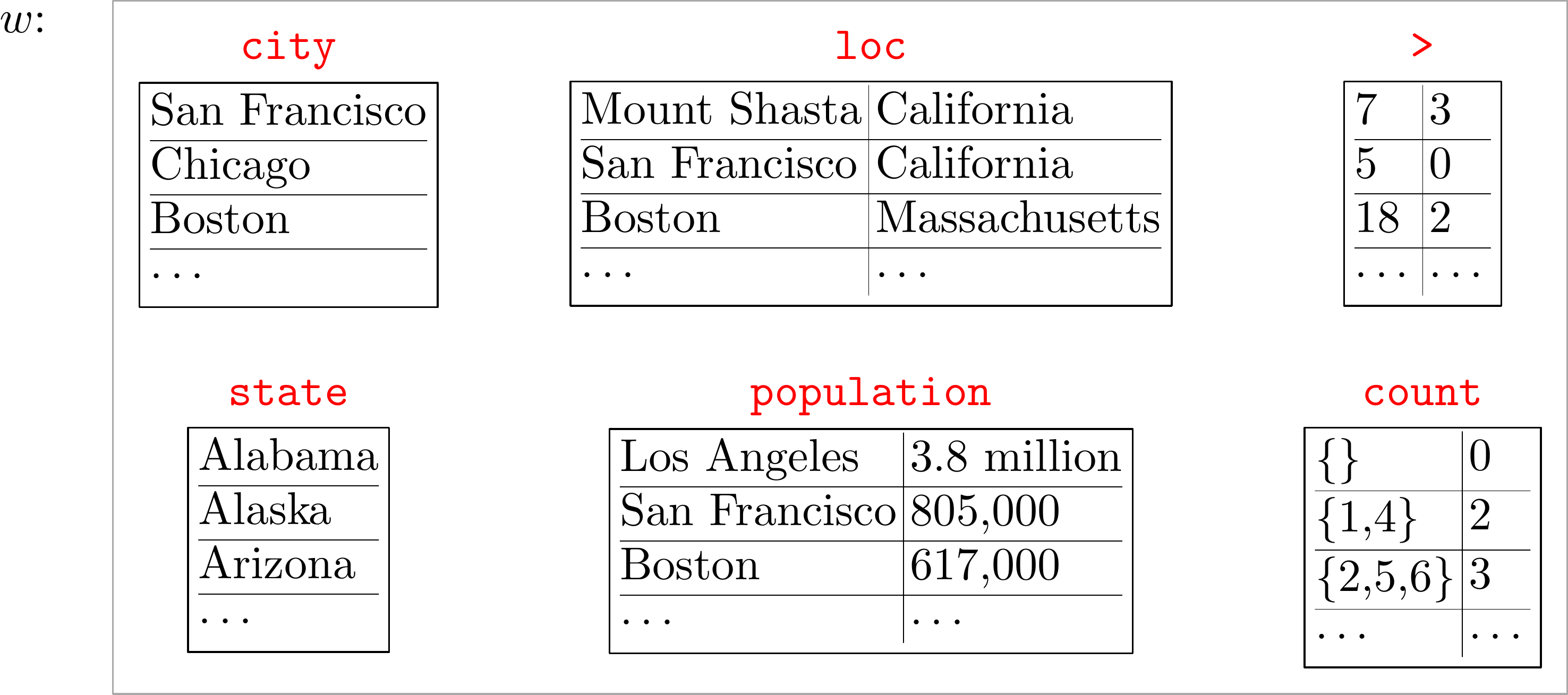}{0.35}{worldExample}{
We use the domain of US geography as a running example.
The figure presents an example of a world $w$ (database)
in this domain. A world maps each predicate to a set of tuples.
For example, the depicted world $w$ maps the predicate \wl{loc} to the set of
pairs of places and their containers.  Note that functions (e.g.,
\wl{population}) are also represented as predicates for uniformity.
Some predicates (e.g., \wl{count}) map to an infinite number of tuples
and would be represented implicitly.
}

%%%%%%%%%%%%%%%%%%%%%%%%%%%%%%
\SecTwo{worlds}{Worlds}

In the context of question answering,
the DCS tree is a formal specification of the question.
To obtain an answer, we still need to evaluate the DCS tree with respect to a
database of facts (see \reffig{worldExample} for an example).
We will use the term {\em world} to refer to this database
(it is sometimes also called model, but we avoid this term to avoid confusion
with the probabilistic model for learning that we will present in \refsec{model}).
%In other words, we need to 
%In order to define the semantics of a DCS tree,
%we need to make reference to a world.

%The semantics, or denotation, of a DCS tree $z \in \sZ$, is defined with
%respect to a {\em world} $w$, also known as a model or database (see
%\reffig{worldExample} for an example).

%%%%%%%%%%%%%%%%%%%%%%%%%%%%%%
\SecThree{values}{Types and Values}

% Values
To define a world, 
we start by constructing a set of {\em values} $\sV$.
The exact set of values depend on the domain (we will continue to use US
geography as a running example).
Briefly, $\sV$ contains numbers (e.g., $3 \in \sV$),
strings (e.g., $\nl{Washington} \in \sV$),
tuples (e.g., $(3,\nl{Washington}) \in \sV$),
sets (e.g., $\{3,\nl{Washington}\} \in \sV$),
and other higher-order entities.

To be more precise, we construct $\sV$ recursively.  First,
define a set of primitive values $\sV_\entity$, which includes the following:
\begin{itemize}
\item Numeric values: each value has the form $x \cto t \in \sV_\entity$,
where
$x \in \R$ is a real number and $t \in \{ \wl{number}, \wl{ordinal}, \wl{percent}, \wl{length}, \dots \}$ is a tag.
The tag allows us to differentiate 3, 3rd, 3\%, and 3 miles---this will be important in \refsec{filtering}.
%Note that some tags are generic and others depend on the domain.
We simply write $x$ for the value $x\cto\wl{number}$.

\item Symbolic values: each value has the form $x \cto t \in \sV_\entity$,
where $x$ is a string (e.g., \nl{Washington})
and $t \in \{ \wl{string}, \wl{city}, \wl{state}, \wl{river}, \dots \}$ is a tag.
Again, the tag 
allows us to differentiate, for example, the entities $\nl{Washington}\cto\wl{city}$ and $\nl{Washington}\cto\wl{state}$.
\end{itemize}

Now we build the full set of values $\sV$ from the primitive values $\sV_\entity$.
To define $\sV$, we need a bit more machinery:~To avoid logical
paradoxes, we construct $\sV$ in increasing order of complexity using
types (see \citet{carpenter98type} for a similar construction).
The casual reader can skip this construction without losing any intuition.

% Types
Define the set of types $\sT$ to be the smallest set that satisfies the following properties:
\begin{enumerate}
\item The primitive type $\entity \in \sT$;
\item The tuple type $(t_1, \dots, t_k) \in \sT$ for each $k \ge 0$ and each non-tuple type $t_i \in \sT$ for $i = 1, \dots, k$; and
\item The set type $\{ t \} \in \sT$ for each tuple type $t \in \sT$.
\end{enumerate}
Note that $\{\entity\}$, $\{ \{ \entity \} \}$, and $((\entity))$ are not valid types.

% Values
For each type $t \in \sT$, we construct a corresponding set of values $\sV_t$:
\begin{enumerate}
\item For the primitive type $t = \entity$, the primitive values $\sV_\entity$ have already been specified.
Note that these types are rather coarse:~Primitive values with different tags are considered to have the same type $\entity$.
\item For a tuple type $t = (t_1, \dots, t_k)$, $\sV_t$ is the cross product of the values of its component types:
\begin{align}
\sV_t = \{ (v_1, \dots, v_k) : \forall i, v_i \in \sV_{t_i} \}.
\end{align}
\item For a set type $t = \{ t' \}$, $\sV_t$ contains all subsets of its element type $t'$:
\begin{align}
\sV_t = \{ s : s \subset \sV_{t'} \}.
\end{align}
Note that all elements of the set must have the same type.
\end{enumerate}
Let $\sV = \cup_{t \in \sT} \sV_t$ be the set of all possible values.

% World
A world maps each predicate
to its {\em semantics}, which is a set of tuples (see \reffig{worldExample} for an example).
First, let $\sTtuple \subset \sT$ be the tuple types, which are the ones of the form $(t_1, \dots, t_k)$.
Let $\sVsettuple$ denote all the sets of tuples (with the same type):
\begin{align}
\sVsettuple &\eqdef \bigcup_{t \in \sTtuple} \sV_{\{t\}}.
\end{align}
Now we define a world formally:
\begin{definition}[World]
\label{def:world}
A {\em world} $w : \sP \mapsto \sVsettuple \cup \{\sV\}$ is a function that
maps each non-null predicate $p \in \sP \backslash \{\nil\}$ to a set of tuples $w(p) \in \sVsettuple$
and maps the null predicate $\nil$ to the set of all values ($w(\nil) = \sV$).
\end{definition}

% Arity
For a set of tuples $A$ with the same arity, let $\arity{A} = |x|$, where $x \in A$ is arbitrary;
if $A$ is empty, then $\arity{A}$ is undefined.
Now for a predicate $p \in \sP$ and world $w$,
define $\arityw{p}$, the arity of predicate $p$ with respect to $w$,
as follows:
\begin{align}
\arityw{p} = \begin{cases}
1            & \text{if $p = \nil$}, \\
\arity{w(p)} & \text{if $p \neq \nil$}.
\end{cases}
\end{align}
The null predicate has arity 1 by fiat;
the arity of a non-null predicate $p$ is inherited from the tuples in $w(p)$.

% Functions
\paragraph{Remarks}
In higher-order logic and lambda calculus, we construct
function types and values, whereas in DCS, we construct tuple types and values.
The two are equivalent in representational power, but this discrepancy does
point at the fact that lambda calculus is based on function application, whereas DCS, as we will see, is
based on declarative constraints.  The set type $\{(\entity,\entity)\}$ in DCS
corresponds to the function type $\entity \to (\entity \to \wl{bool})$.
In DCS, there is no explicit $\wl{bool}$ type---it is implicitly represented by using sets.

%%%%%%%%%%%%%%%%%%%%%%%%%%%%%%
\SecThree{predExamples}{Examples}

The world $w$ maps each domain-specific predicate to a finite set of tuples.
For the US geography domain, $w$ has a predicate that maps to the set of US states (\wl{state}),
another predicate that maps to the set of pairs of entities and where they are located (\wl{loc}),
and so on:
\begin{align}
w(\wl{state}) &= \{ (\nl{California}\cto\wl{state}), (\nl{Oregon}\cto\wl{state}), \dots \}, \\
w(\wl{loc}) &= \{ (\nl{San Francisco}\cto\wl{city}, \nl{California}\cto\wl{state}), \dots \} \\
      & \dots
\end{align}
To shorten notation, we use state abbreviations (e.g., $\wl{CA} = \nl{California}\cto\wl{state}$).

The world $w$ also specifies the semantics of several domain-independent predicates (think of these as helper functions),
which usually correspond to an infinite set of tuples.
%Therefore, $w$ is only conceptually a relational database.
Functions are represented in DCS by a set of input-output pairs.
For example, the semantics of the $\Count_t$ predicate (for each type $t \in \sT$)
contains pairs of sets $S$ and their cardinalities $|S|$:
\begin{align}
w(\Count_t) = \{ (S,|S|) : S \in \sV_{\{(t)\}} \} \in \sV_{\{(\{(t)\},\entity)\}}.
\end{align}

As another example, consider the predicate $\average_t$ (for each $t \in \sT$), which takes
a set of key-value pairs (with keys of type $t$) and returns the average value.
For notational convenience, we treat an arbitrary set of pairs $S$ as a
set-valued function:~We let $S_1 = \{ x : (x,y) \in S \}$ denote the domain of
the function,
and abusing notation slightly, we define the function 
$S(x) = \{ y : (x,y) \in S \}$ to be the set of values $y$ that co-occur with the given $x$.
The semantics of $\average_t$ contains pairs of sets and their averages:
\begin{align}
w(\average_t) = \pc{ (S,z) : S \in \sV_{\{(t,\entity)\}}, z = |S_1|^{-1} \sum_{x \in S_1} \pb{ |S(x)|^{-1} \sum_{y \in S(x)} y } } \in \sV_{\{(\{(t,\entity)\},\entity)\}}.
\end{align}
Similarly, we can define the semantics of $\Argmin_t$ and $\Argmax_t$,
which each takes a set of key-value pairs and returns the keys that attain the smallest (largest) value:
\begin{align}
w(\Argmin_t) &= \pc{ (S,z) : S \in \sV_{\{(t,\entity)\}}, z \in \argmin_{x \in S_1} \min S(x) } \in \sV_{\{(\{(t,\entity)\},t)\}}, \\
w(\Argmax_t) &= \pc{ (S,z) : S \in \sV_{\{(t,\entity)\}}, z \in \argmax_{x \in S_1} \max S(x) } \in \sV_{\{(\{(t,\entity)\},t)\}}.
\end{align}

These helper functions are monomorphic:~For example, $\Count_t$ only computes cardinalities of sets of type $\{(t)\}$.
%We made this design decision to avoid the complications of polymorphism,
%but the price we pay is that we need one predicate for each type $t$.
In practice, we mostly operate on sets of primitives ($t = \entity$).
To reduce notation, we omit $t$ to refer to this version:
$\Count = \Count_{\entity}$, $\average = \average_{\entity}$, etc.

%%%%%%%%%%%%%%%%%%%%%%%%%%%%%%
\SecTwo{basic}{Semantics of DCS Trees: Basic Version}

%Having defined the syntax of DCS trees (\refsec{syntax}) and given a formal definition of worlds (\refsec{worlds}),
%we are now in position to define the semantics of DCS trees.
The semantics or {\em denotation} of a DCS tree $z$ with respect to a world $w$ is denoted $\den{z}$.
First, we define the semantics of DCS trees with only join relations (\refsec{csp}).
In this case, a DCS tree encodes a constraint satisfaction problem (CSP);
this is important because it highlights the constraint-based nature of DCS
and also naturally leads to a computationally efficient way of computing denotations (\refsec{computation}).
We then allow DCS trees to have aggregate relations (\refsec{aggregation}).
The fragment of DCS which has only join and aggregate relations is called {\em basic DCS}.

\SecThree{csp}{DCS Trees as Constraint Satisfaction Problems}

% Definition
Let $z$ be a DCS tree with only join relations on its edges.
In this case, $z$ encodes a constraint satisfaction problem (CSP) as follows:
For each node $x$ in $z$, the CSP has a variable $a(x)$;
the collection of these variables is referred to as an {\em assignment} $a$.
The predicates and relations of $z$ introduce constraints:
\begin{enumerate}
\item $a(x) \in w(p)$ for each node $x$ labeled with predicate $p \in \sP$; and
\item $a(x)_j = a(y)_{j'}$ for each edge $(x,y)$ labeled with $\JoinRel{j}{j'} \in \sR$,
which says that the $j$-th component of $a(x)$ must equal the $j'$-th
component of $a(y)$.
\end{enumerate}
We say that an assignment $a$ is {\em feasible} if it satisfies all the above constraints.
Next, for a node $x$, define $V(x) = \{ a(x) : \text{assignment $a$ is feasible} \}$
as the set of feasible values for $x$---these are the ones which are consistent with at least one feasible assignment.
Finally, we define the denotation of the DCS tree $z$ with respect to the world $w$
to be $\den{z} = V(x_0)$, where $x_0$ is the root node of $z$.

% Example
\reffig{simpleExists}(a) shows an example of a DCS tree.
The corresponding CSP has four variables $c,m,\ell,s$.\footnote{Technically, the node is $c$ and the variable is $a(c)$,
but we use $c$ to denote the variable to simplify notation.}
In \reffig{simpleExists}(b), we have written the equivalent lambda calculus formula.
The non-root nodes are existentially quantified, the root node $c$ is $\lambda$-abstracted,
and all constraints introduced by predicates and relations are conjoined.
The $\lambda$-abstraction of $c$ represents the fact that the denotation is the set of feasible values for $c$
(note the equivalence between the boolean function $\lambda c . p(c)$ and the set $\{ c : p(c) \}$).

\FigTop{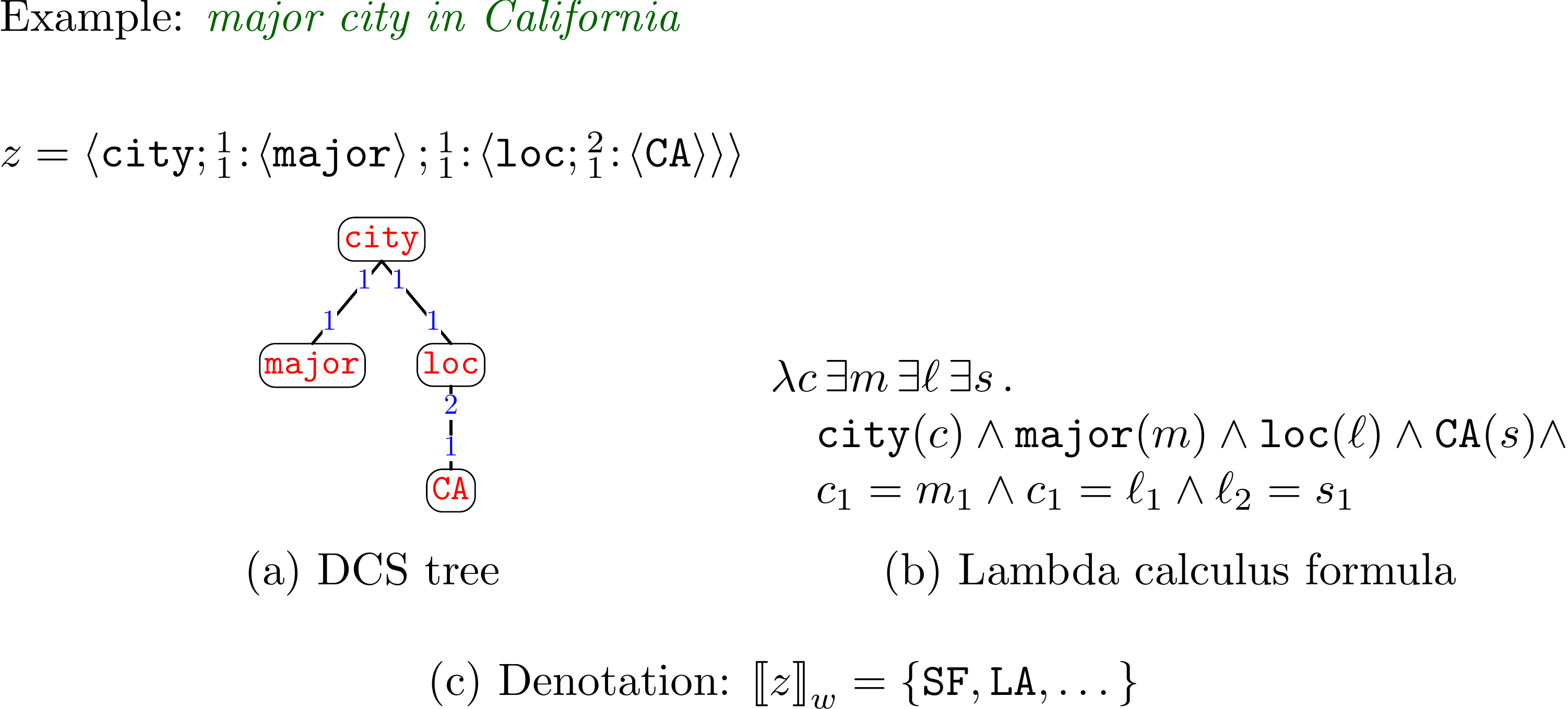}{0.4}{simpleExists}{(a) An example of a DCS tree
(written in both the mathematical and graphical notation).  Each node is labeled with a predicate,
and each edge is labeled with a relation.
(b) A DCS tree $z$ with only join relations encodes a constraint
satisfaction problem. (c) The denotation of $z$ is
the set of feasible values for the root node.
}

\paragraph{Remarks}

% Conjunction and existential
Note that CSPs only allow existential quantification and conjunction.
Why did we choose this particular logical subset as a starting point,
rather than allowing universal quantification, negation, or disjunction?
There seems to be something fundamental about this subset,
which also appears in Discourse Representation Theory (DRT) \citep{kamp93drt,kamp05drt}.
Briefly, logical forms in DRT are called Discourse Representation Structures (DRSes),
each of which contains 
(i) a set of existentially-quantified discourse referents (variables), (ii) a set of conjoined discourse conditions (constraints),
and (iii) nested DRSes.  If we exclude nested DRSes,
a DRS is exactly a CSP.\footnote{DRSes are not necessarily tree-structured, though economical DRT \citep{bos09economical} imposes a tree-like restriction on DRSes
for computational reasons.}
The default existential quantification and conjunction are quite natural 
for modeling cross-sentential anaphora:~New variables can be added to a DRS and connected to other variables.
Indeed, DRT was originally motivated by these phenomena (see \citet{kamp93drt} for more details).

% Representational power - tree, anaphora
Tree-structured CSPs can capture unboundedly complex recursive
structures---such as \nl{cities in states that border states that have rivers
that\dots}.  Trees are limited, however, in that they are unable to
capture long-distance dependencies such as those arising from anaphora.
For example, consider the phrase \nl{a state with a river that traverses 
its capital}.  Here, {\em its} binds to {\em state}, but this dependence 
cannot be captured in a tree structure.  A solution to this problem 
can be pursued within the fuller formalism of CSPs and their realization
as graphical models; we simply introduce an edge between the {\em its} 
node and the {\em state} node which introduces a CSP constraint that the 
two nodes must be equal.  We will not pursue an exploration of non-tree 
structures in the current \paper, but it should be noted that such an 
extension is possible and quite natural.

%%%%%%%%%%%%%%%%%%%%%%%%%%%%%%
\SecThree{computation}{Computation}

% Recurrence - dynamic programming
So far, we have given a declarative definition of the denotation $\den{z}$ of a
DCS tree $z$ with only join relations.
Now we will show how to compute $\den{z}$
efficiently. 
Recall that the denotation is the set of feasible values for the root node.
In general, finding the solution to a CSP is NP-hard, but for trees,
we can exploit dynamic programming~\citep{dechter03constraint}.
The key is that the denotation of a tree depends on its subtrees
only through their denotations:
\begin{align}
\den{\syn{p;\JoinRel{j_1}{j_1'} \cto c_1; \cdots; \JoinRel{j_m}{j_m'} \cto c_m}} 
= w(p) \,\cap\, \bigcap_{i=1}^m \{ v : v_{j_i} = t_{j_i'}, t \in \den{c_i} \}. \label{eqn:joinDenSet}
\end{align}
On the right-hand side of \refeqn{joinDenSet}, the first term $w(p)$ is the set of values that satisfy the node constraint,
and the second term consists of an intersection across all $m$ edges of $\{ v : v_{j_i} = t_{j_i'}, t
\in \den{c_i} \}$, which is the set of values $v$ which satisfy the edge constraint with respect to
some value $t$ for the child $c_i$.

% Join and project
To further flesh out this computation,
we express \refeqn{joinDenSet} in terms of two operations: {\em join} and {\em project}.
Join takes a cross product of two sets of tuples and retains the resulting tuples that match the join constraint:
\begin{align}
A \joinOp{j,j'} B &= \{ u + v : u \in A, v \in B, u_j = v'_j \}. \label{eqn:simpleJoin}
\end{align}
Project takes a set of tuples and retains a fixed subset of the components:
\begin{align}
\ProjOp{\bi}{A} &= \{ v_\bi : v \in A \}. \label{eqn:simpleProj}
\end{align}
The denotation in \refeqn{joinDenSet} can now be expressed in terms of these join and project operations:
\begin{align}
\den{\syn{p;\JoinRel{j_1}{j_1'} \cto c_1; \cdots; \JoinRel{j_m}{j_m'} \cto c_m}} 
= \ProjOp{\bi}{(\ProjOp{\bi}{(w(p) \joinOp{j_1,j_1'} \den{c_1})} \cdots \joinOp{j_m,j'_m} \den{c_m})}, \label{eqn:joinDen}
\end{align}
where $\bi = (1,\dots,\arityw{p})$. Projecting onto $\bi$ retains only components corresponding to $p$.

% Complexity
The time complexity for computing the denotation of a DCS tree $\den{z}$ scales
linearly with the number of nodes, but there is also a dependence on the cost
of performing the join and project operations.  For details on how we optimize
these operations and handle infinite sets of tuples,
see \ifjournal
\citet{liang11thesis}
\else
\refapp{computationDetails}
\fi.

% Shift in focus
The denotation of DCS trees is defined in terms of the feasible values of a CSP,
and the recurrence in \refeqn{joinDenSet} is only one way of computing this denotation.
However, in light of the extensions to come, we now consider \refeqn{joinDenSet}
as the actual definition rather than just a computational mechanism.
It will still be useful to refer to the CSP in order to access the intuition of using declarative constraints.

% MJ:  omit to save space...
%\paragraph{Remarks}
%
%The fact that trees enable efficient computation is a general theme which appears in many algorithmic contexts,
%notably, in probabilistic inference in graphical models.  Indeed, a CSP can be
%viewed as a deterministic version of a graphical model where we maintain only supports of distributions (sets of feasible values)
%rather than the full distribution.
%% Connection
%Recall that trees---dependency trees, in particular---also naturally capture the
%syntactic locality of natural language utterances.  This is not a coincidence for
%the two are intimately connected, and DCS trees
%highlight their connection.\footnote{ %the connection between syntactic locality in the dependency structure and efficient semantic evaluation.
%As a side note, if we introduce $k$ non-tree edges, we obtain a graph with tree-width at most $k$.
%We could then modify our recurrences to compute the denotation of those graphs,
%akin to the junction tree algorithm in graphical models.}

%%%%%%%%%%%%%%%%%%%%%%%%%%%%%%
\SecThree{aggregation}{Aggregate Relation}

\FigTop{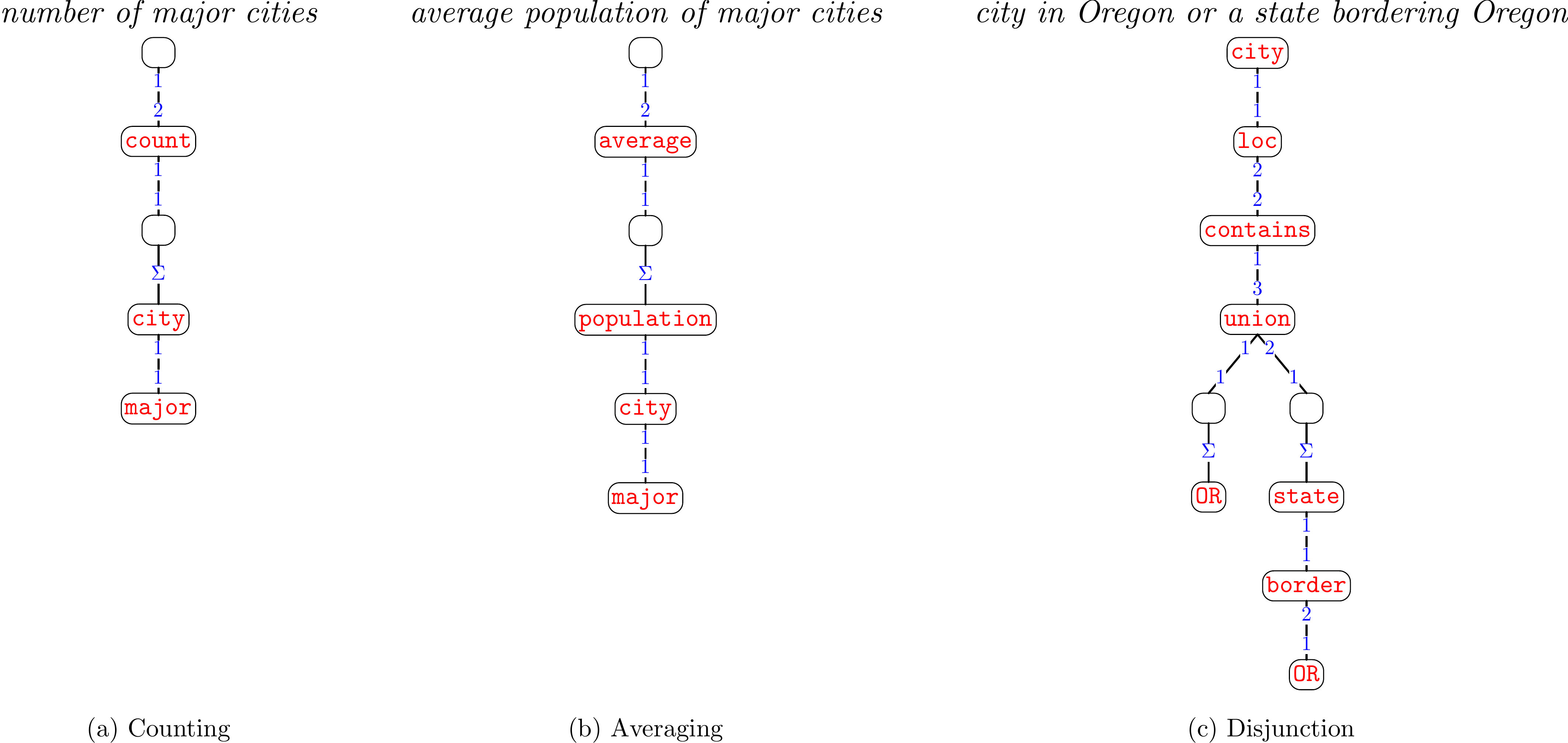}{0.3}{aggregate}{
Examples of DCS trees that use the aggregate relation ($\SigmaRel$) to
(a) compute the cardinality of a set, (b) take the average over a set,
(c) represent a disjunction over two conditions.
The aggregate relation sets the parent node deterministically to the denotation of the child node.
}

Thus far, we have focused on DCS trees that only use join relations,
%These DCS trees encode CSPs and have denotations that can be computed efficiently using dynamic programming.
%These restricted DCS trees can represent arbitrarily complex compositional structures,
which are insufficient for capturing higher-order phenomena in language.
% Example
For example, consider the phrase \nl{number of major cities}.
Suppose that \nl{number} corresponds to the $\Count$ predicate,
and that \nl{major cities} maps to the DCS tree $\syn{\wl{city}; \JoinRel{1}{1} \cto \syn{\wl{major}}}$.
We cannot simply join $\Count$ with the root of this DCS tree because
$\Count$ needs to be joined with the {\em set} of major cities (the denotation of
$\syn{\wl{city}; \JoinRel{1}{1} \cto \syn{\wl{major}}}$),
not just a single city.

%We will argue that it is impossible to use CSPs alone to capture the meaning of
%this phrase.
%First, as can be seen from \refeqn{joinDenSet}, if the denotation of any subtree is empty,
%then the denotation of the entire DCS tree is also empty.
%Now suppose there do not exist any major cities.
%The part of the DCS tree that corresponds to \nl{major cities}
%would then have an empty denotation, and therefore,
%the overall denotation would also be $\emptyset$,
%rather than the desired $\{ (0) \}$.

% Solution
We therefore introduce the {\em aggregate relation} ($\SigmaRel$) that takes a DCS subtree
and reifies its denotation so that it can be accessed by other nodes in its entirety.
Consider a tree $\syn{\SigmaRel \cto c}$, where the root is connected to a child $c$ via $\SigmaRel$.
The denotation of the root is simply the singleton set containing the denotation of $c$:
\begin{align}
\den{\syn{\SigmaRel \cto c}} = \{ (\den{c}) \}. \label{eqn:sigmaDen}
\end{align}

\reffig{aggregate}(a) shows the DCS tree for our running example.
The denotation of the middle node is $\{(s)\}$, where $s$ is all major cities.
Everything above this node is an ordinary CSP:
$s$ constrains the $\Count$ node, which in turns constrains the root node to $|s|$.
\reffig{aggregate}(b) shows another example of using the aggregate relation $\SigmaRel$.
Here, the node right above $\SigmaRel$ is
constrained to be a set of pairs of major cities and their populations.
The $\average$ predicate then computes the desired answer.

% Disjunction
%Recall that by default, all nodes in a DCS tree are existentially quantified
%and all conditions are combined using conjunction.
To represent
logical disjunction in natural language, we use the aggregate relation
and two predicates, \wl{union} and \wl{contains}, which are defined in the expected way:
\begin{align}
w(\wl{union}) &= \{ (S,B,C) : C = A \cup B \}, \\
w(\wl{contains}) &= \{ (A,x) : x \in A \}.
\end{align}
\reffig{aggregate}(c) shows an example of a disjunctive construction:~We use the aggregate relations
to construct two sets, one containing Oregon, and the other containing states bordering Oregon.
We take the union of these two sets; $\wl{contains}$ takes the set
and reads out an element, which then constrains the \wl{city} node.

\paragraph{Remarks}

% Collection of CSPs
A DCS tree that contains only join and aggregate relations can be viewed as a
collection of tree-structured CSPs connected via aggregate relations.
The tree structure still enables us to compute denotations
efficiently based on the recurrences in \refeqns{joinDen}{sigmaDen}.

% DRT analogy
Recall that a DCS tree with only join relations is a DRS without nested DRSes.
The aggregate relation corresponds to the abstraction operator in DRT
and is one way of making nested DRSes.  It turns out that the abstraction operator is sufficient
to obtain the full representational power of DRT,
and subsumes generalized quantification and disjunction constructs in DRT.
By analogy, we use the aggregate relation to handle disjunction
(\reffig{aggregate}(c)) and generalized quantification (\refsec{execute}).

% Expressive power
DCS restricted to join relations is less expressive than first-order logic because it does not
have universal quantification, negation, and disjunction. 
The aggregate relation is analogous to lambda abstraction,
and in basic DCS we use the aggregate relation to implement 
those basic constructs using higher-order predicates
such as \wl{not},\wl{every}, and \wl{union}.
We can also express logical statements such as generalized quantification,
which go beyond first-order logic.

%%%%%%%%%%%%%%%%%%%%%%%%%%%%%%
\SecTwo{full}{Semantics of DCS Trees: Full Version}

\FigTop{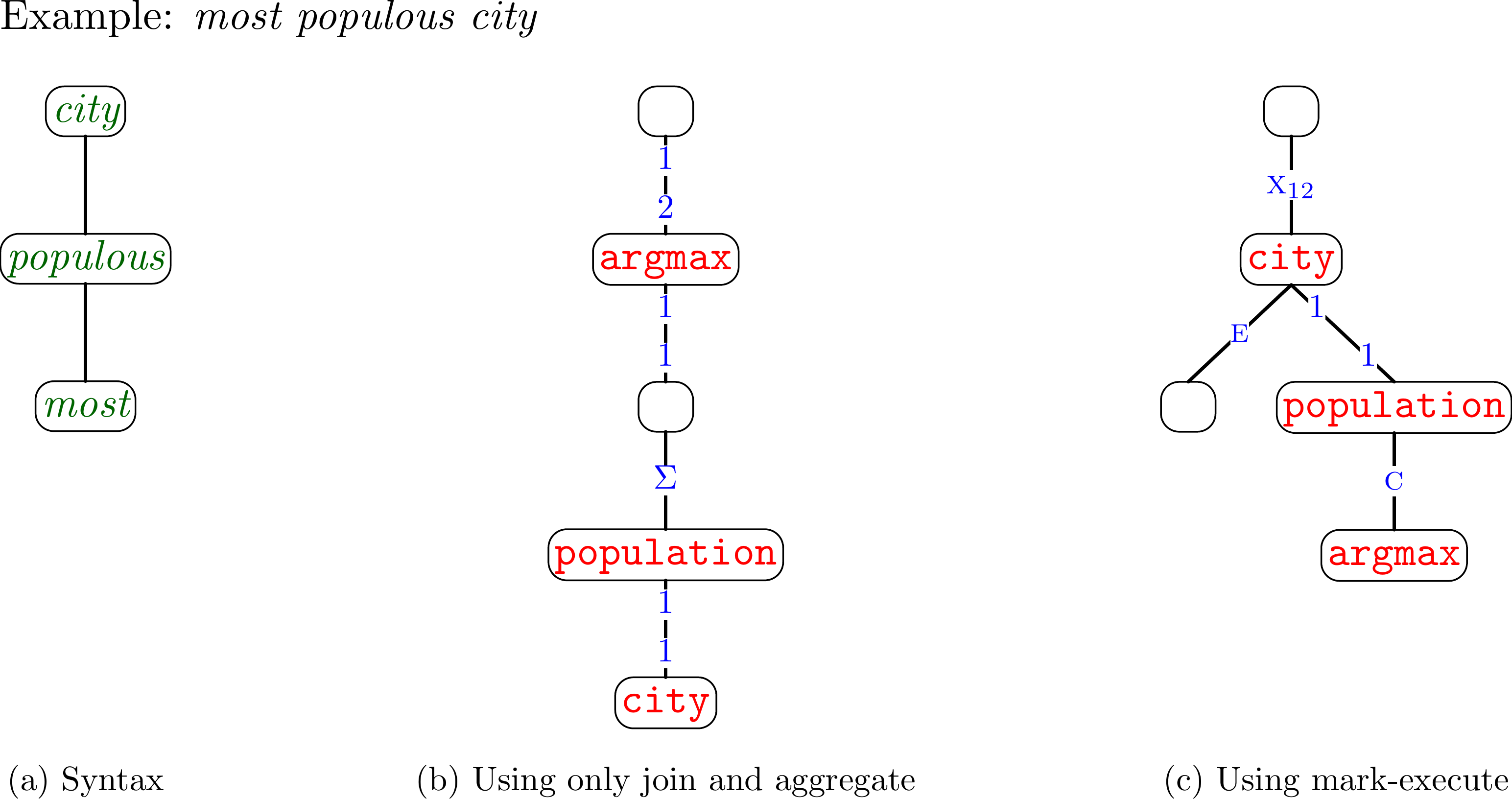}{0.35}{basicToFull}{
Two semantically-equivalent DCS trees are shown in (b) and (c).
The DCS tree in (b),
which uses the join and aggregate relations in the basic DCS,
does not align well with the syntactic structure of \nl{most populous city} (a),
and thus is undesirable.
The DCS tree in (c), by using the mark-execute construct, aligns much better,
with \wl{city} rightfully dominating its modifiers.
The full version of DCS allows us to construct (c), which is preferable to (b).
}

% Problem: divergence
Basic DCS allows only join and aggregate relations,
but is already quite expressive. % in terms of expressing many possible denotations.
However, it is not enough to simply express a denotation using an arbitrary logical form;
one logical form can be better than another even if the two are semantically equivalent.
For example, consider the superlative construction \nl{most populous city},
which has a basic syntactic dependency structure shown in \reffig{basicToFull}(a).
\reffig{basicToFull}(b) shows a DCS tree with only join and aggregate relations
that expresses the correct semantics.
However, the two structures are quite divergent---the syntactic head
is \nl{city} and the semantic head is \wl{argmax}.
This divergence runs counter to a principal desideratum of DCS, which is to
create a transparent interface between syntax and semantics.

% Solution
In this section, we resolve this dilemma by introducing mark and execute relations, which will allow
us to use the DCS tree in \reffig{basicToFull}(c) to represent the 
semantics associated with  \reffig{basicToFull}(a).
The focus of this section is on this {\em mark-execute} construct---using
mark and execute relations to give proper semantically-scoped denotations
to syntactically-scoped tree structures.

% Basic idea
The basic intuition of the mark-execute construct is as follows:
We mark a node low in the tree with a {\em mark relation};
then, higher up in the tree, we invoke it with a corresponding 
\emph{execute relation} (\reffig{markExecute}).
For our example in \reffig{basicToFull}(c),
we mark the \wl{population} node, which puts the child \wl{argmax} in a temporary store;
when we execute the \wl{city} node, we fetch the superlative predicate \wl{argmax} from the store and
invoke it.

% Other cases
This divergence between syntactic and semantic scope
arises in other linguistic contexts besides superlatives such as quantification and negation.
In each of these cases, the general template is the same: a syntactic modifier low in the tree needs to
have semantic force higher in the tree.  A particularly compelling case of this divergence happens
with quantifier scope ambiguity (e.g., \nl{Some river traverses every city.}),
where the quantifiers appear in fixed syntactic positions,
but the wide or narrow reading correspond to different semantically-scoped denotations.
Analogously, a single syntactic structure involving superlatives
can also yield two different semantically-scoped denotations---the absolute and relative readings
(e.g., \nl{state bordering the largest state}).
The mark-execute construct provides a unified framework for dealing all these forms of divergence
between syntactic and semantic scope.
See \reffig{examples} for concrete examples of
this construct.

\FigTop{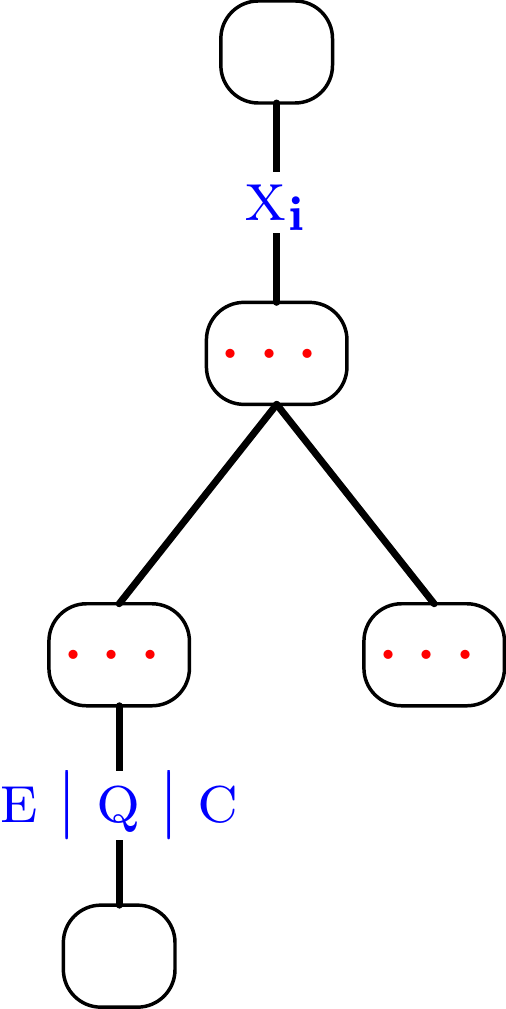}{0.35}{markExecute}{
The template for the mark-execute construct.
A mark relation (one of $\ExtractRel$, $\QuantRel$, $\CompareRel$) ``stores'' the modifier.
Then an execute relation (of the form $\X_\bi$ for indices $\bi$) higher up ``recalls'' the modifier
and applies it at the desired semantic point.
}

\FigStar{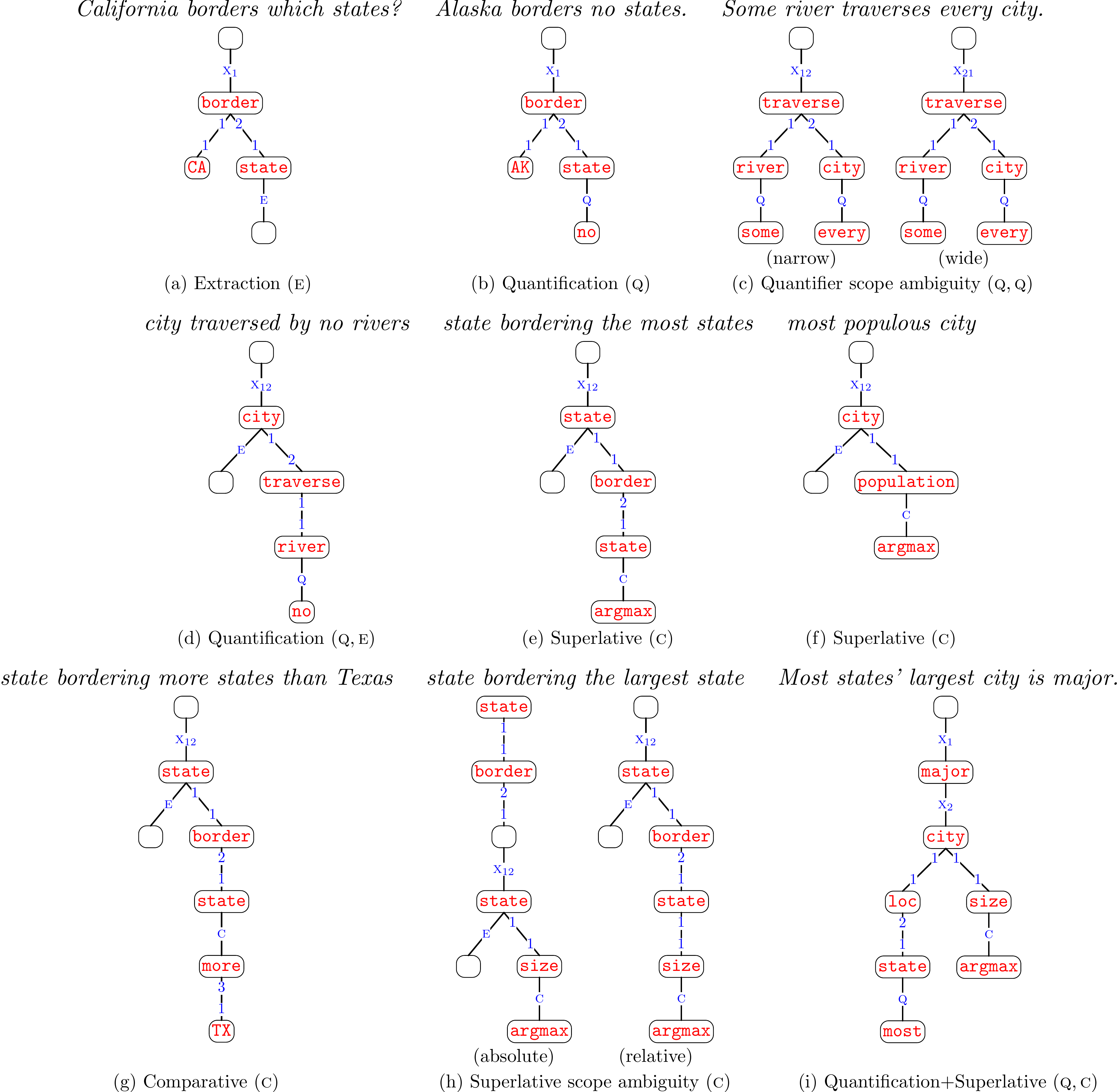}{0.33}{examples}{
Examples of DCS trees that use the mark-execute construct.
(a) The head verb \nl{borders}, which needs to be returned, has
a direct object \nl{states} modified by \nl{which}.
(b) The quantifier \nl{no} is syntactically dominated by \nl{state}
but needs to take wider scope.
(c) Two quantifiers yield two possible readings;
we build the same basic structure, marking both quantifiers;
the choice of execute relation ($\X_{12}$ versus $\X_{21}$) determines the reading.
(d) We employ two mark relations, $\QuantRel$ on \wl{river} for the negation,
and $\ExtractRel$ on \wl{city} to force the quantifier to be computed for each value of \wl{city}.
(e,f,g) Analogous construction but with the $\CompareRel$ relation (for comparatives and superlatives).
(h) Analog of quantifier scope ambiguity for superlatives:
the placement of the execute relation determines an absolute versus relative reading.
(i) Interaction between a quantifier and a superlative.
%The lower execute relation computes the largest city for each state;
%the second execute relation invokes \wl{most} and enforces that the \wl{major} constraint holds for the majority of states.
}

%%%%%%%%%%%%%%%%%%%%%%%%%%%%%%
\SecThree{denotations}{Denotations}

% Formalizing
We now formalize the mark-execute construct.
We saw that the mark-execute construct appears to act non-locally, putting things in a store and retrieving them later.
This means that if we want the denotation of a DCS tree to only depend on the denotations of its subtrees,
the denotations need to contain more than the set of feasible values for the root node, as was the case for basic DCS.
We need to augment denotations to include information about all marked nodes, since 
these can be accessed by an execute relation higher up in the tree.

\FigTop{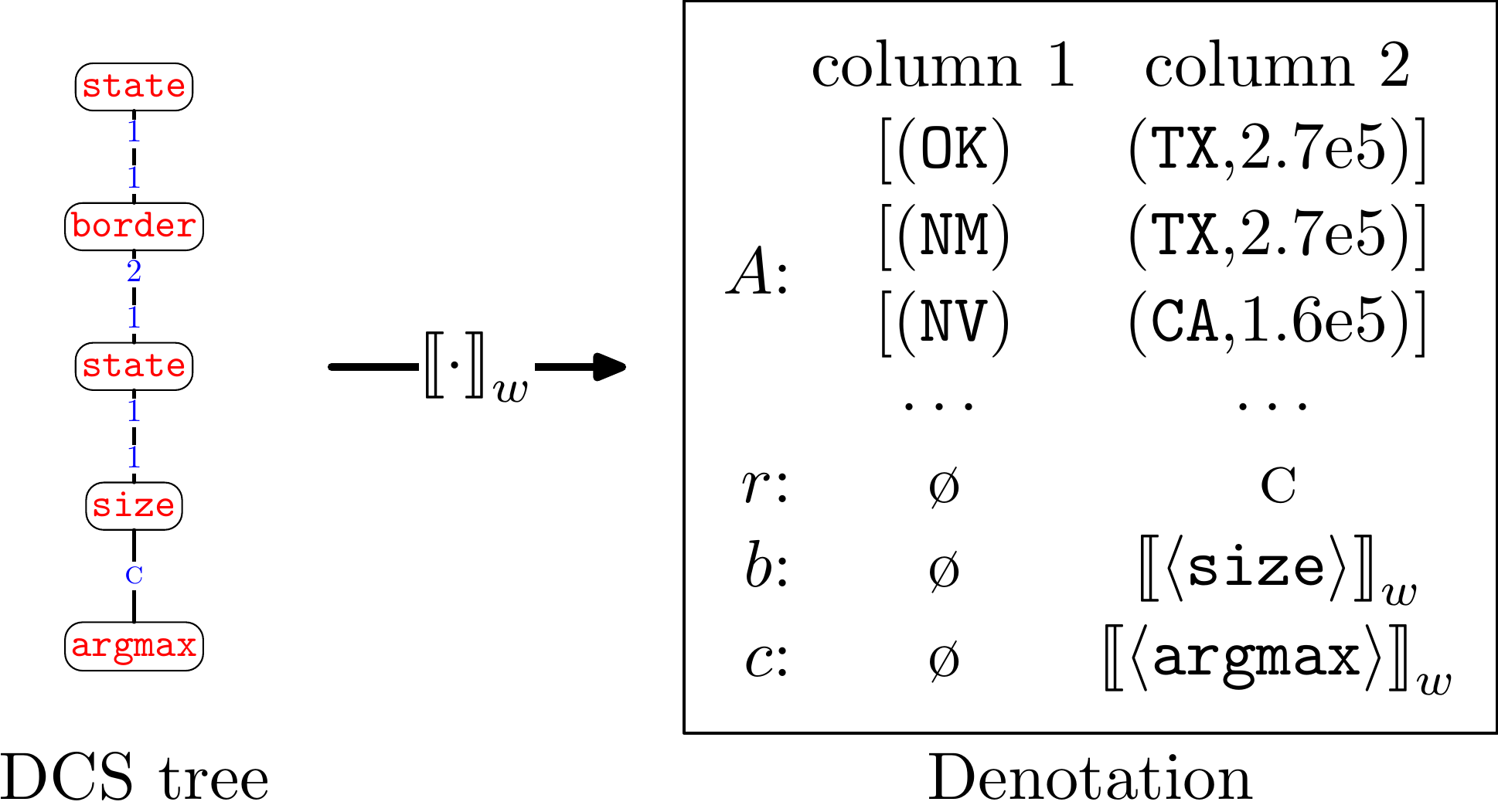}{0.35}{denotation}{
Example of the denotation for a DCS tree with a compare relation $\CompareRel$.
This denotation has two columns, one for each active node---the root node \wl{state} and
the marked node \wl{size}.
}

% Intuitive definition
More specifically, let $z$ be a DCS tree and $d = \den{z}$ be its denotation.
The denotation $d$ consists of $n$ {\em columns},
where each column is either the root node of $z$ or a non-executed marked node in $z$.
In the example in \reffig{denotation}, there are two columns, one for the root \wl{state} node 
and the other for \wl{size} node, which is marked by $\CompareRel$.
The columns are ordered according to a pre-order traversal of $z$,
so column 1 always corresponds to the root node.
The denotation $d$ contains a set of arrays $d.A$, where each array represents a feasible assignment of values to the columns of $d$.
For example, in \reffig{denotation}, the first array in $d.A$ corresponds to
assigning $(\wl{OK})$ to the \wl{state} node (column 1) and
$(\wl{TX},\text{2.7e5})$ to the \wl{size} node (column 2).
If there are no marked nodes, $d.A$ is basically a set of tuples, which corresponds to
a denotation in basic DCS.
For each marked node, 
the denotation $d$ also maintains a {\em store} with information to be retrieved when that marked node is executed.
A store $\store$ for a marked node contains the following:
(i) the mark relation $\store.r$ ($\CompareRel$ in the example),
(ii) the base denotation $\store.b$ which essentially corresponds to denotation of the subtree rooted at the marked node
excluding the mark relation and its subtree ($\den{\syn{\wl{size}}}$ in the example),
and (iii) the denotation of the child of the mark relation ($\den{\syn{\wl{argmax}}}$ in the example).
The store of any non-marked nodes (e.g., the root) is empty ($\store = \nilthree$).
% Formal definition
%Formally, a denotation is defined as follows:
\begin{definition}[Denotations]
Let $\sD$ be the set of denotations, where each denotation $d \in \sD$ consists of
\begin{itemize}
\item a set of arrays $d.A$,
where each array $\ba = \arr{a_1,\dots,a_n} \in d.A$ is a sequence of $n$ tuples; and
\item a sequence of $n$ {\em stores} $d.\bstore = (d.\store_1, \dots, d.\store_n)$,
where each store $\store$
contains a mark relation $\store.r \in \{ \ExtractRel, \QuantRel, \CompareRel, \nil \}$,
a base denotation $\store.b \in \sD \cup \{ \nil \}$,
and a child denotation $\store.c \in \sD \cup \{ \nil \}$.
\end{itemize}
\end{definition}
Note that denotations are formally defined without reference to DCS trees (just
as sets of tuples were in basic DCS), but it is sometimes useful to refer to
the DCS tree that generates that denotation.

% Notation
For notational convenience,
we write $d$ as $\sem{A; (r_1,b_1,c_1); \dots; (r_n,b_n,c_n)}$.
Also let $d.r_i = d.\store_i.r$, $d.b_i = d.\store_i.b$, and $d.c_i = d.\store_i.c$.
Let $d\with{\store_i = x}$ be the denotation which is identical to $d$, except with $d.\store_i = x$;
$d\with{r_i = x}$, $d\with{b_i = x}$, and $d\with{c_i = x}$ are defined analogously.
We also define a project operation for denotations:
$\ProjOp{\bi}{\sem{A;\bstore}} \eqdef \sem{\{\ba_{\bi} : \ba \in A\}; \bstore_{\bi}}$.
Extending this notation further,
we use $\nil$ to denote the indices of the {\em non-initial columns with empty stores}
($i>1$ such that $d.\store_i = \nilthree$).
We can then use $\ProjOp{-\nil}{d}$ to represent projecting away the non-initial columns with empty stores.
For the denotation $d$ in \reffig{denotation}, $\ProjOp{1}{d}$ keeps column 1, $\ProjOp{-\nil}{d}$ keeps both columns,
and $\ProjOp{2,-2}{d}$ swaps the two columns.

%In basic DCS, denotations were simply sets of tuples.
%In the current general framework, such a set $s$ would be represented
%using a one-column denotation: $\sem{\{\arr{v}:v \in s\};\nilthree}$.

% Truth values
In basic DCS, denotations are sets of tuples,
which works quite well for representing the semantics of wh-questions such as \nl{What states border Texas?}
But what about polar questions such as \nl{Does Louisiana border Texas?}
The denotation should be a simple boolean value, which basic DCS does not represent explicitly.
Using our new denotations, we can represent boolean values explicitly using
zero-column structures: {\em true} corresponds to a singleton set containing
just the empty array ($\dtrue = \sem{\{\arr{\,}\}}$) and {\em false} is the
empty set ($\dfalse = \sem{\emptyset}$).

% Basic structure
Having described denotations as $n$-column structures,
we now give the formal mapping from DCS trees to these structures.
As in basic DCS,
this mapping is defined recursively over the structure of the tree.
We have a recurrence for each case (the first line is the base case,
and each of the others handles a different edge relation):
\begin{align}
\den{\syn{p}} &= \sem{\{\arr{v}:v \in w(p)\};\nilthree}, &\aside{base case} \label{eqn:baseFullDen} \\
\den{\syn{p; \bolde; \JoinRel{j}{j'} \cto c}} &= \JoinProjOp{j,j'}{\densyn{p;\bolde}}{\den{c}}, &\aside{join} \label{eqn:joinFullDen} \\
\den{\syn{p; \bolde; \SigmaRel \cto c}} &= \JoinProjOp{*,*}{\densyn{p;\bolde}}{\SumOp{\den{c}}}, &\aside{aggregate} \label{eqn:sigmaFullDen} \\
\den{\syn{p; \bolde; \X_\bi \cto c}} &= \JoinProjOp{*,*}{\densyn{p;\bolde}}{\bX_\bi(\den{c})}, &\aside{execute} \label{eqn:xFullDen} \\
\den{\syn{p; \bolde; \ExtractRel \cto c}} &= \Mark(\densyn{p;\bolde},\ExtractRel,\den{c}), &\aside{extract} \label{eqn:eFullDen} \\
\den{\syn{p; \bolde; \CompareRel \cto c}} &= \Mark(\densyn{p;\bolde},\CompareRel,\den{c}), &\aside{compare} \label{eqn:cFullDen} \\
\den{\syn{p; \QuantRel \cto c; \bolde}} &= \Mark(\densyn{p;\bolde},\QuantRel,\den{c}). &\aside{quantify} \label{eqn:qFullDen}
\end{align}
We define the operations $\joinProjOp{\bj,\bj'},\Sigma,\bX_\bi,\bM$ 
in the remainder of this section.

% Explanation
\SecThree{base}{Base Case}

\refeqn{baseFullDen} defines the denotation for a DCS tree $z$ with a single
node with predicate $p$.  The denotation of $z$ has one column whose arrays
correspond to the tuples $w(p)$; the store for that column is empty.

%Let $z$ be a DCS tree.
%If the last child $c$ of $z$'s root is a join ($\JoinRel{j}{j'}$), aggregate ($\SigmaRel$), or execute ($\X_\bi$) relation (\refeqn{joinFullDen}--\refeqn{xFullDen}),
%then we simply recurse on $z$ with $c$ removed and join it with some transformation (identity, $\Sigma$, or $\X_\bi$) of $c$'s denotation.
%If the last (or first) child is connected via a mark relation $\ExtractRel,\CompareRel$ (or $\QuantRel$),
%then we strip off that child and put the appropriate information in the store by invoking $\bM$.
%We assume that \refeqn{eFullDen}--\refeqn{cFullDen} take precedence over the other rules.
%We now define the operations $\JoinProjOp{\bj,\bj'}{}{},\Sigma,\bX_\bi,\bM$.

%%%%%%%%%%%%%%%%%%%%%%%%%%%%%%
\SecThree{join}{Join Relations}

\refeqn{joinFullDen} defines the recurrence for join relations.
On the left-hand side, $\syn{p;\be;\JoinRel{j}{j'} \cto c}$ is 
a DCS tree with $p$ at the root, a sequence of edges $\be$ followed by a final edge with relation $\JoinRel{j}{j'}$
connected to a child DCS tree $c$.
On the right-hand side, we take the recursively computed denotation of $\syn{p;\be}$, the DCS tree without the final edge,
and perform a {\em join-project-inactive} operation (notated $\joinProjOp{j,j'}$) with the denotation of the child DCS tree $c$.

The join-project-inactive operation joins the arrays of the two denotations (this
is the core of the join operation in basic DCS---see \refeqn{simpleJoin}),
and then projects away the non-initial empty columns:
\begin{align}
\JoinProjOp{\bj,\bj'}{\sem{A;\bstore}}{\sem{A';\bstore'}} &= \ProjOp{-\nil}{\sem{A''; \bstore + \bstore'}}, \text{where} \\
\quad A'' &= \{ \ba + \ba' : \ba \in A, \ba' \in A', a_{1\bj} = a'_{1\bj'} \}.\nonumber
\end{align}
We concatenate all arrays $\ba \in A$ with all arrays $\ba' \in A'$ that satisfy the join condition
$a_{1\bj} = a'_{1\bj'}$.  The sequences of stores are simply concatenated: ($\bstore+\bstore'$).
Finally, any non-initial columns with empty stores are projected away by applying $\ProjOp{-\nil}{\cdot}$.

Note that the join works on column 1; the other columns are carried along for the ride.
As another piece of convenient notation,
we use $*$ to represent all components,
so $\joinProjOp{*,*}$ imposes the join condition that the entire tuple has to agree ($a_1 = a'_1$).

%%%%%%%%%%%%%%%%%%%%%%%%%%%%%%
\SecThree{aggregate}{Aggregate Relations}

\FigTop{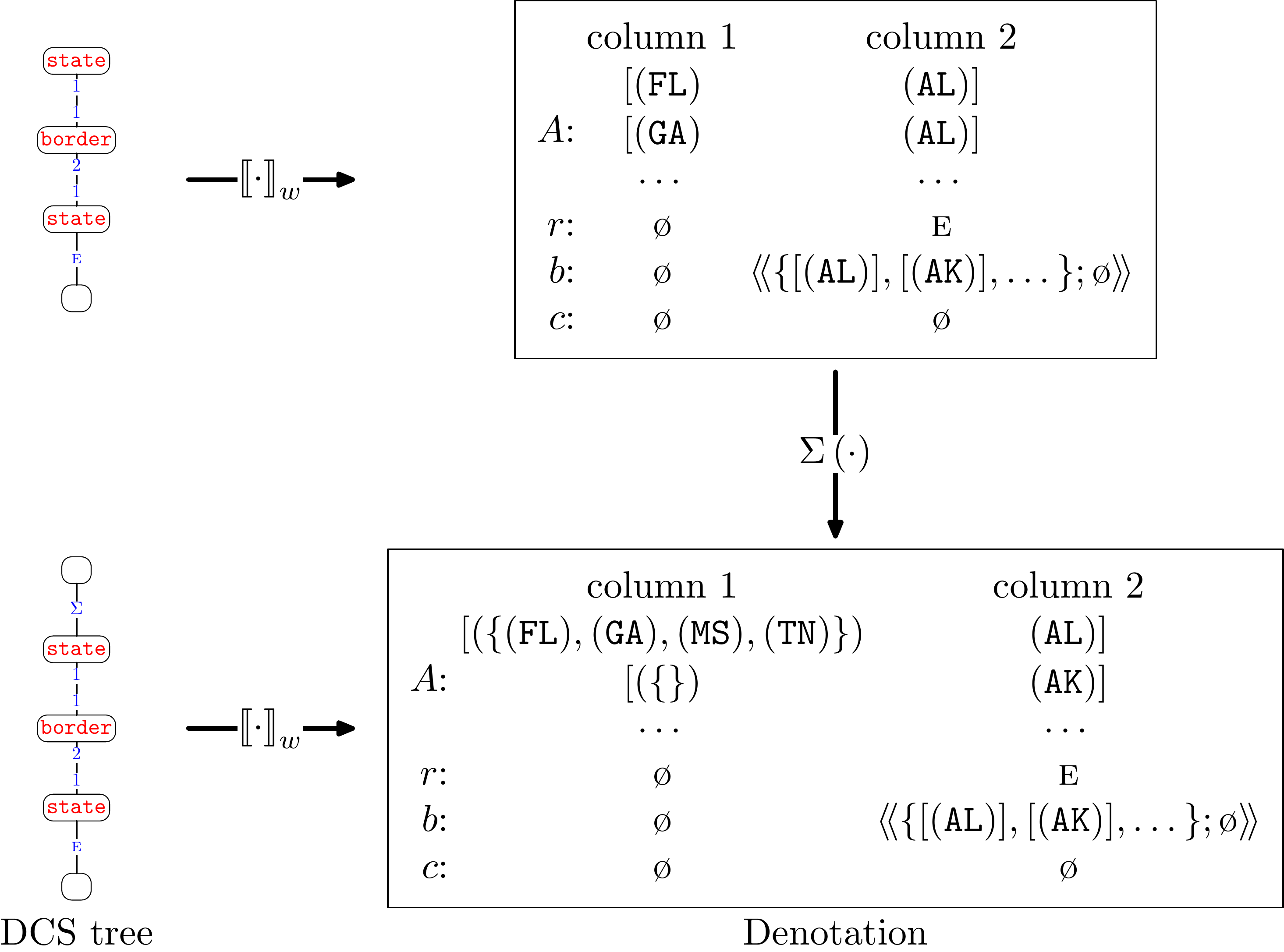}{0.35}{aggregateFull}{
An example of applying the aggregate operation,
which takes a denotation and aggregates the values in column 1 for every setting of the other columns.
The base denotations ($b$) are used to put in $\{\}$ for values that do not appear in $A$.
(in this example, $\wl{AK}$, corresponding to the fact that Alaska does not border any states).
}

\refeqn{sigmaFullDen} defines the recurrence for aggregate relations.
Recall that in basic DCS, aggregate \refeqn{sigmaDen} simply takes the denotation (a set of tuples)
and puts it into a set.
Now, the denotation is not just a set, so we need to generalize this operation.
Specifically, the aggregate operation applied to a denotation forms a set out of the tuples in the first column
for each setting of the rest of the columns:
\begin{align}
\SumOp{\sem{A;\bstore}} &= \sem{A' \cup A'';\bstore}, \\
A' &= \{ \arr{S(\ba),a_2,\dots,a_n} : \ba \in A \}, \nonumber \\
S(\ba) &= \{ a_1' : \arr{a_1',a_2,\dots,a_n} \in A \}, \nonumber \\
A'' &= \{ \arr{\emptyset,a_2,\dots,a_n} : \forall i \in \{2,\dots,n\}, \arr{a_i} \in \ProjOp{1}{\store_i.b.A}, \neg\exists a_1, \ba \in A \}. \nonumber
\end{align}
The aggregate operation takes the set of arrays $A$
and produces two sets of arrays, $A'$ and $A''$, which are unioned (note that the stores do not change).
The set $A'$ is the one that first comes to mind:~For every setting of $a_2,\dots,a_n$, we construct $S(\ba)$,
the set of tuples $a_1'$ in the first column which co-occur with $a_2,\dots,a_n$ in $A$.

However, there is another case: what happens to settings of $a_2,\dots,a_n$ that do not co-occur with any value of $a_1'$ in $A$?
Then, $S(\ba) = \emptyset$, but note that $A'$ by construction will not have the desired array
$\arr{\emptyset,a_2,\dots,a_n}$.
As a concrete example, 
suppose $A = \emptyset$ and we have one column ($n = 1$).
Then $A' = \emptyset$, rather than the desired $\{[\emptyset]\}$.
%which would be the desired answer, to be consistent with \refeqn{sigmaDen}.

Fixing this problem is slightly tricky.  There are an infinite number of $a_2,\dots,a_n$
which do not co-occur with any $a_1'$ in $A$, so for which ones do we actually 
include $\arr{\emptyset,a_2,\dots,a_n}$?  Certainly, the answer to this question cannot come from $A$,
so it must come from the stores.  In particular, for each column $i \in \{2, \dots, n\}$,
we have conveniently stored a base denotation $\store_i.b$.
We consider any $a_i$ that occurs in column 1 of the arrays of this base denotation 
($[a_i] \in \ProjOp{1}{\store_i.b.A}$).
For this $a_2,\dots,a_n$,
we include $\arr{\emptyset,a_2,\dots,a_n}$ in $A''$ 
as long as $a_2,\dots,a_n$ does not co-occur with any $a_1$.
An example is given in \reffig{aggregateFull}.

The reason for storing base denotations is thus partially revealed:
The arrays represent feasible values of a CSP
and can only contain positive information.
When we aggregate, we need to access possibly empty sets of feasible values---a kind of negative information,
which can only be recovered from the base denotations.

%%%%%%%%%%%%%%%%%%%%%%%%%%%%%%
\SecThree{mark}{Mark Relations}

\FigTop{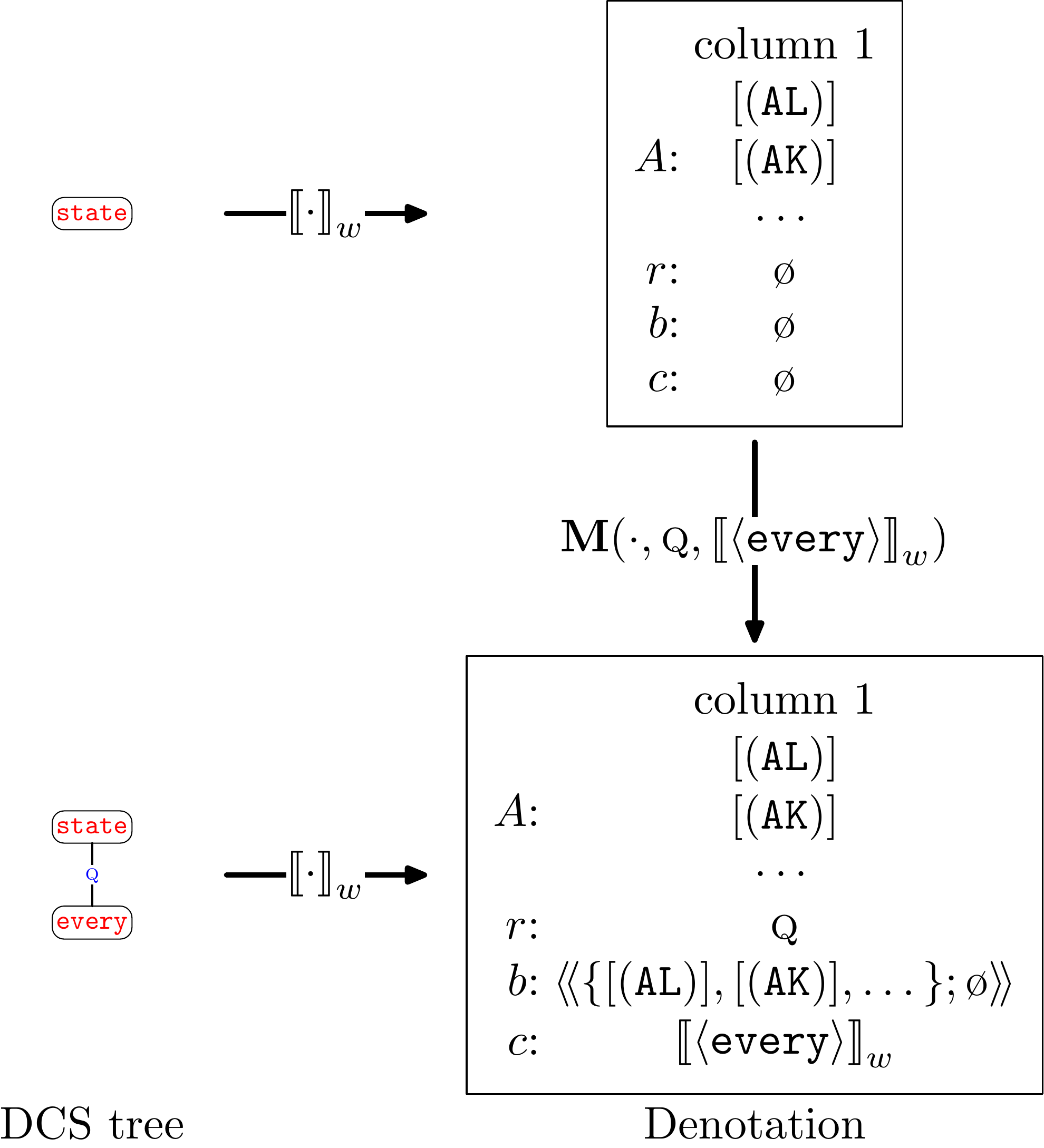}{0.35}{mark}{
An example of applying the mark operation,
which takes a denotation and modifies the store of the column 1.
This information is used by other operations such as aggregate and execute.
}

\refeqn{eFullDen}, \refeqn{cFullDen}, and \refeqn{qFullDen} each processes a different mark relation.
We define a general mark operation, $\Mark(d,r,c)$ which takes a denotation $d$,
a mark relation $r \in \{\ExtractRel,\QuantRel,\CompareRel\}$ and a child denotation $c$,
and sets the store of $d$ in column 1 to be $(r,d,c)$:
\begin{align}
\Mark(d,r,c) &= d\with{r_1 = r, b_1 = d, c_1 = c}.
\end{align}
The base denotation of the first column $b_1$ is set to the current denotation $d$.
This, in some sense, creates a snapshot of the current denotation.
%We already saw that these base denotations are used in aggregate relations. % for determining the values that do not appear in the set of arrays.
\reffig{mark} shows an example of the mark operation.

%%%%%%%%%%%%%%%%%%%%%%%%%%%%%%
\SecThree{execute}{Execute Relations}

\refeqn{xFullDen} defines the denotation of a DCS tree where the last edge of the root is an execute relation.
Similar to the aggregate case \refeqn{sigmaFullDen},
we recurse on the DCS tree without the last edge ($\syn{p;\be}$) and then join it to the result of applying the execute operation $\bX_\bi$
to the denotation of the child ($\den{c}$).

% Complexity goes here
The execute operation $\bX_\bi$ is the most intricate part of DCS and is what does the heavy lifting.
%There is no free lunch: we simplified the DCS trees themselves, but their semantics become more complex.
%However, we believe that this is tradeoff is to our advantage---this somewhat more complex semantics
%apply quite general and is fixed once and for all.  The real difficulty is mapping
%from natural language to DCS trees, and simplifying that part is a big win.
The operation is parametrized by a sequence of distinct indices $\bi$
which specifies the order in which the columns should be processed.
Specifically, $\bi$ indexes into the subsequence of columns with non-empty stores.
%(which are all columns with the possible exception of column 1).
We then process this subsequence of columns in reverse order,
where processing a column means performing some operations depending on the stored relation in that column.
For example,
suppose that columns 2 and 3 are the only non-empty columns.
Then $\bX_{12}$ processes column 3 before column 2.
On the other hand, $\bX_{21}$ processes column 2 before column 3.
We first define the execute operation $\bX_i$ for a single column $i$.
There are three distinct cases, depending on the relation stored in column $i$:

%\paragraph{Extraction ($d.r_i = \ExtractRel$)}
\SecFour{extraction}{Extraction}

\FigTop{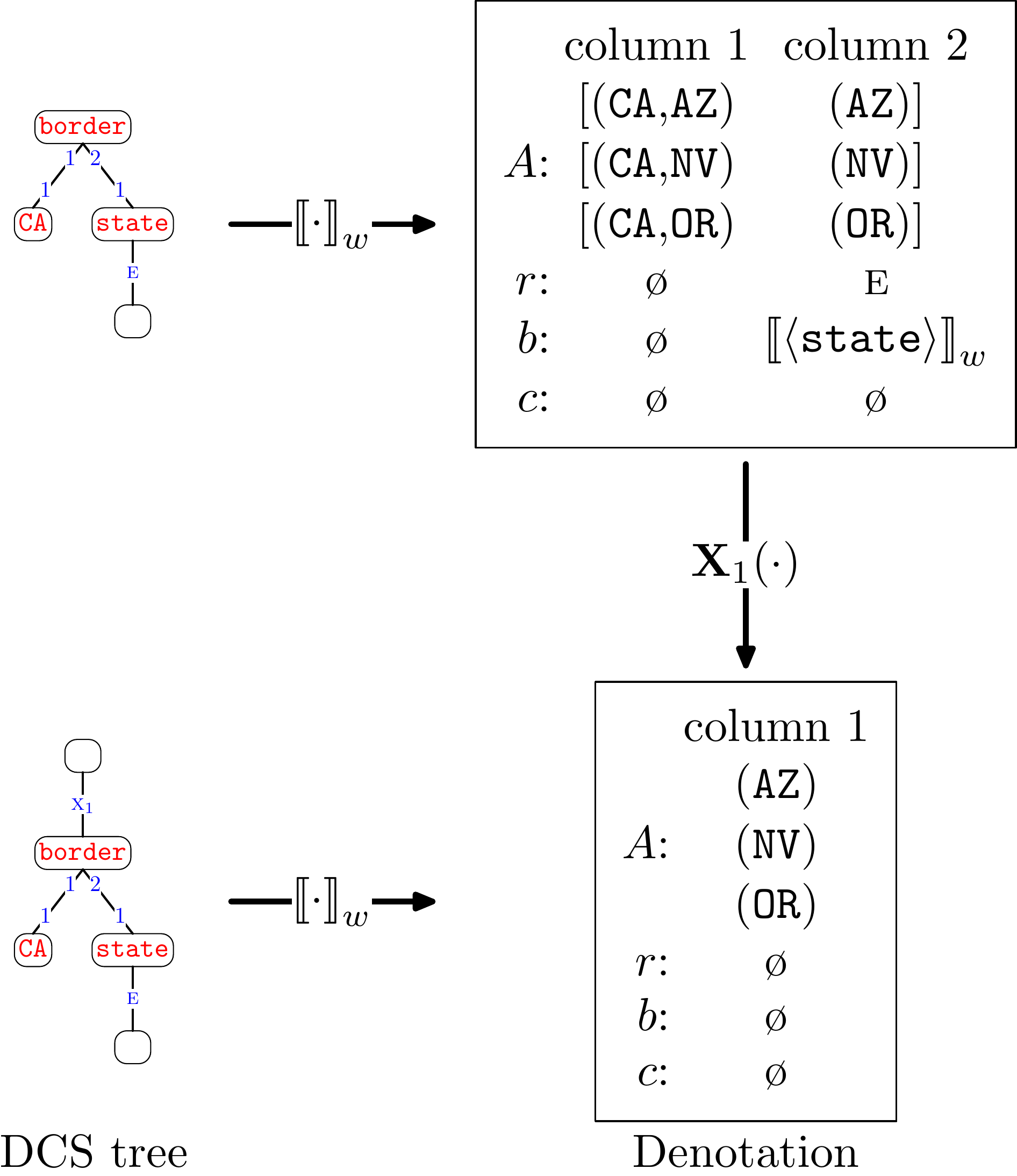}{0.35}{extract}{
An example of applying the execute operation on column $i$ with the extract relation $\ExtractRel$.
}

For a denotation $d$ with the extract relation $\ExtractRel$ in column $i$,
executing $\bX_i(d)$ involves three steps:
(i) moving column $i$ to before column 1 ($\ProjOp{i,-i}{\cdot}$),
(ii) projecting away non-initial empty columns ($\ProjOp{-\nil}{\cdot}$),
and (iii) removing the store ($\cdot\with{\store_1 = \nil}$):
\begin{align}
%\bX_i(d) = \ProjOp{i,-(i,\nil)}{d}\with{\store_1 = \nil} \quad\text{if $d.r_i = \ExtractRel$}.
\bX_i(d) = \ProjOp{-\nil}{\ProjOp{i,-i}{d}}\with{\store_1 = \nil} \quad\text{if $d.r_i = \ExtractRel$}. \label{eqn:executeExtract}
\end{align}
An example is given in \reffig{extract}.
There are two main uses of extraction:
\begin{enumerate}
\item By default, the denotation of a DCS tree is the set of feasible values of the root node (which occupies column 1).
To return the set of feasible values of another node,
we mark that node with $\ExtractRel$.
Upon execution, the feasible values of that node move into column 1.
See \reffig{examples}(a) for an example.

\item
Unmarked nodes are existentially quantified and have narrower scope than all marked nodes.
Therefore, we can make a node $x$ have wider scope than another node $y$
by marking $x$ (with $\ExtractRel$) and executing $y$ before $x$
(see \reffig{examples}(d,e) for examples).
%Then all of the computation will be done conditioned on the value of $x$
The extract relation $\ExtractRel$ (in fact, any mark relation) signifies that we want to control the scope of a node,
and the execute relation allows us to set that scope.
\end{enumerate}

%\paragraph{Generalized Quantification ($d.r_i = \QuantRel$)}
\SecFour{genQuant}{Generalized Quantification}

\FigTop{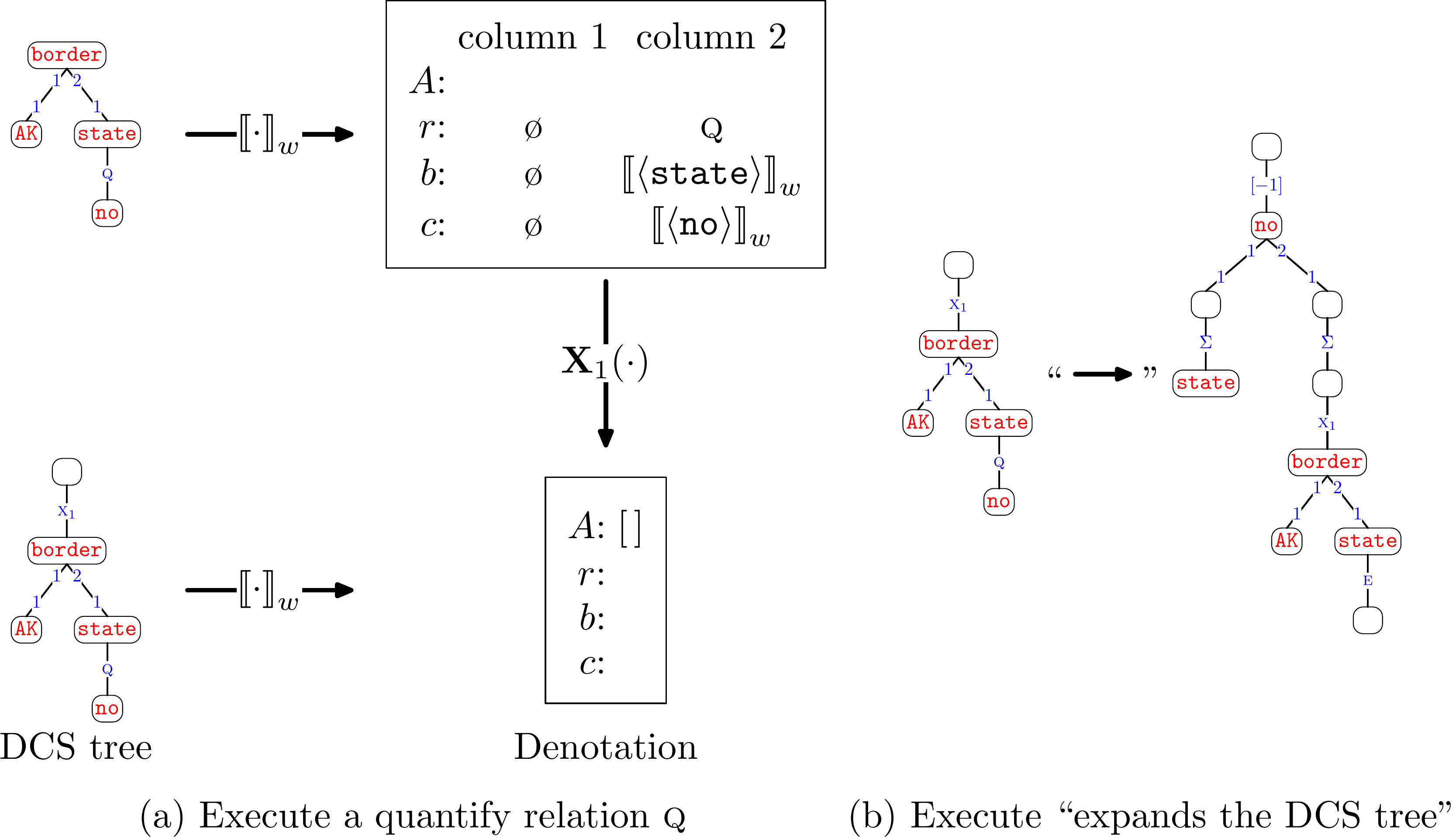}{0.4}{quantify}{
(a) An example of applying the execute operation on column $i$ with the quantify relation $\QuantRel$.
Before executing, note that $A = \{\}$ (because Alaska does not border any states).
The restrictor ($A$) is the set of all states, and the nuclear scope ($B$) is empty.
Since the pair $(A,B)$ does exists in $w(\wl{no})$,
the final denotation, is $\sem{\{\arr{\,}\}}$ (which represents true).
(b)
Although the execute operation actually works on the denotation,
think of it in terms of expanding the DCS tree.
%and is in general cannot be visualized as a DCS tree transformation.
We introduce an extra projection relation $[-1]$, which projects away the first column of the child subtree's denotation.
}

Generalized quantifiers (including negation) are predicates on two sets,
a {\em restrictor} $A$ and a {\em nuclear scope} $B$.  For example,
\begin{align}
w(\wl{some})  &= \{ (A,B) : A \cap B > 0 \}, \\
w(\wl{every}) &= \{ (A,B) : A \subset B \}, \\
w(\wl{no})    &= \{ (A,B) : A \cap B = \emptyset \}, \\
w(\wl{most})  &= \{ (A,B) : |A \cap B| > \half |A| \}.
\end{align}
We think of the quantifier as a modifier
which always appears as the child of a $\QuantRel$ relation;
the restrictor is the parent.
For example, in \reffig{examples}(b), \wl{no} corresponds to the quantifier and \wl{state}
corresponds to the restrictor.  The nuclear scope should be the set of all states that Alaska borders.
More generally, the nuclear scope is the set of feasible values of the restrictor node with respect to the CSP
that includes all nodes between the mark and execute relations.
The restrictor is also the set of feasible values of the restrictor node, but with respect to the CSP
corresponding to the subtree rooted at that node.\footnote{
% Conserativity
Defined this way, we can only handle conservative quantifiers,
since the nuclear scope will always be a subset of the restrictor.
This design decision is inspired by DRT, where it provides a way of modeling donkey anaphora.
We are not treating anaphora in this work, but we can handle it
by allowing pronouns in the nuclear scope to create anaphoric edges into nodes in the restrictor.
These constraints naturally propagate through the nuclear scope's CSP without affecting the restrictor.
}

We implement generalized quantifiers as follows:
Let $d$ be a denotation and suppose we are executing column $i$.
We first construct a denotation for the restrictor $d_A$ and a denotation for the nuclear scope $d_B$.
For the restrictor, we take the base denotation in column $i$ ($d.b_i$)---remember that the base denotation
represents a snapshot of the restrictor node before the nuclear scope constraints are added.
For the nuclear scope, we take the complete denotation $d$ (which includes the nuclear scope constraints)
and extract column $i$ ($\ProjOp{-\nil}{\ProjOp{i,-i}{d}}\with{\store_1 = \nil}$---see \refeqn{executeExtract}).
We then construct $d_A$ and $d_B$ by applying the aggregate operation to each.
Finally, we join these sets with the quantifier denotation, stored in $d.c_i$:
\begin{align}
\bX_i(d) &= \ProjOp{-1}{\p{\JoinProjOp{2,1}{\p{\JoinProjOp{1,1}{d.c_i}{d_A}}}{d_B}}} \quad\text{if $d.r_i = \QuantRel$}, \text{where} \label{eqn:executeQuant} \\
d_A &= \SumOp{d.b_i}, \\
d_B &= \SumOp{\ProjOp{-\nil}{\ProjOp{i,-i}{d}}\with{\store_1 = \nil}}.
\end{align}
When there is one quantifier, think of the
execute relation as performing a syntactic rewriting operation, as shown in
\reffig{quantify}(b).  For more complex cases, we must defer to
\refeqn{executeQuant}.

\reffig{examples}(c) shows an example with two interacting quantifiers.
The denotation of the DCS tree before execution is the same in both readings,
as shown in \reffig{quantify2}.
The quantifier scope ambiguity
is resolved by the choice of execute relation: $\X_{12}$ gives the narrow
reading, $\X_{21}$ gives the wide reading.
\Fig{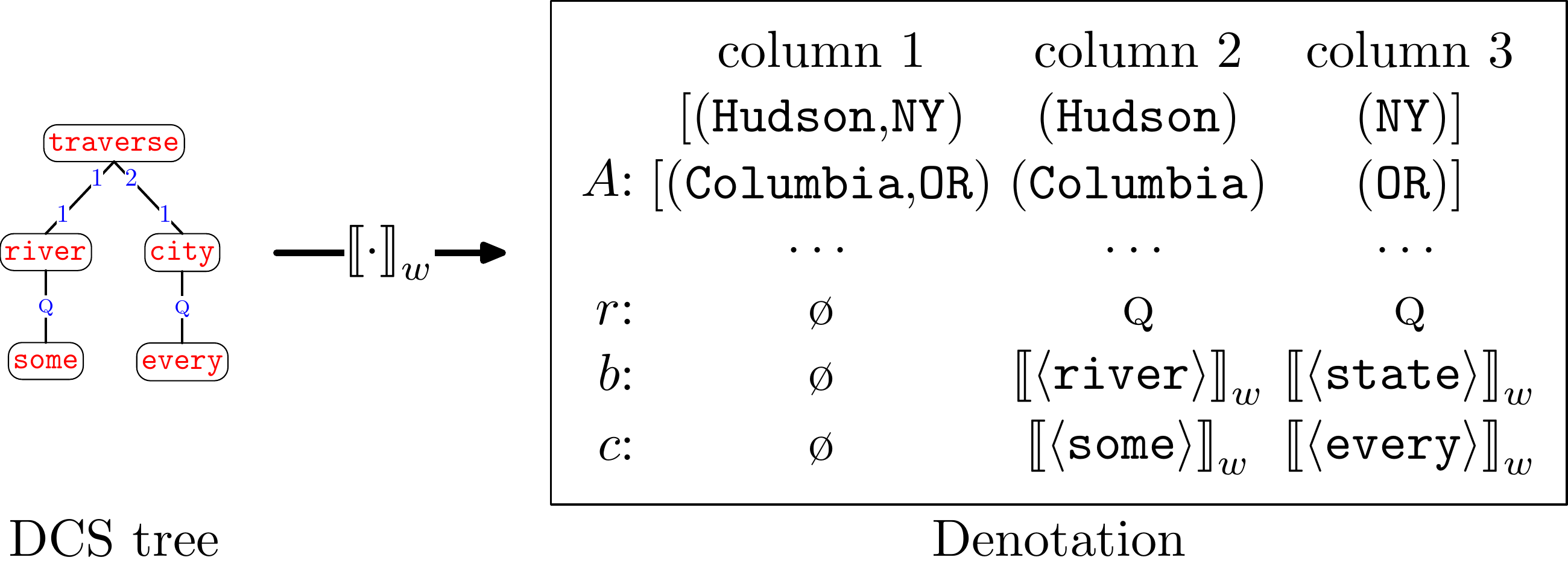}{0.35}{quantify2}{
Denotation of \reffig{examples}(c) before the execute relation is applied.
}

\reffig{examples}(d) shows how extraction and quantification work together.
First, the $\wl{no}$ quantifier is processed for each \wl{city}, which is an unprocessed marked node.
Here, the extract relation is a technical trick to give \wl{city} wider scope.
%since unmarked nodes are 
%with narrower scope than any marked node.
%After that, the actual execution of the extract relation does not do much:
%it merely removes the store.

%\paragraph{Comparatives and Superlatives ($d.r_i = \CompareRel$)}
\SecFour{compare}{Comparatives and Superlatives}

\FigTop{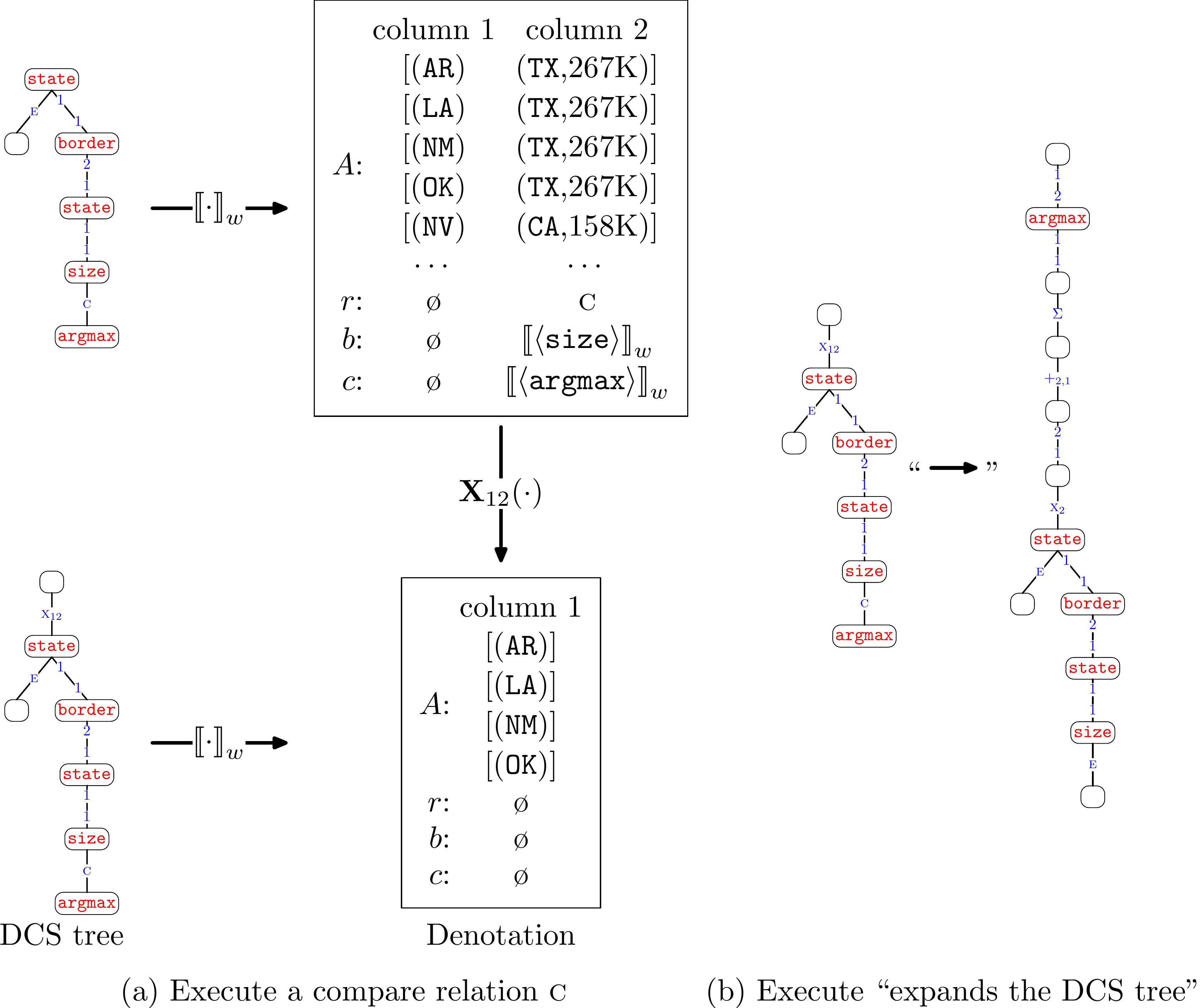}{0.4}{compare}{
(a) Executing a compare relation $\CompareRel$ for an example superlative construction
(relative reading of \nl{state bordering the largest state} from \reffig{examples}(h)).
Before executing, column 1 contains the entity to compare,
and column 2 contains the degree information, of which only the second component is relevant.
After executing, the resulting denotation contains a single column with only the entities
that obtain the highest degree (in this case, the states that border Texas)
(b) For this example, think of the execute operation as expanding the original DCS tree,
although the execute operation actually works on the denotation, not the DCS tree.
The expanded DCS tree has the same denotation as the original DCS tree,
and syntactically captures the essence of the execute-compare operation.
Going through the relations of the expanded DCS tree from bottom to top:
The $\X_2$ relation swaps columns 1 and 2; the join relation keeps only the second component ($(\wl{TX},267K)$ becomes $(267K)$);
$+_{2,1}$ concatenates columns 2 and 1 ($\arr{(267K),(\wl{AR})}$ becomes $\arr{(\wl{AR},267K)}$);
$\Sigma$ aggregates these tuples into a set;
$\wl{argmax}$ operates on this set and returns the elements.
}

Comparative and superlative constructions involve comparing entities,
and for this, we rely on a set $S$ of entity-degree pairs $(x,y)$, where $x$ is an
entity and $y$ is a numeric degree.
Recall that we can treat $S$ as a function, which maps an entity $x$ to the set of degrees $S(x)$ associated with $x$.
Note that this set can contain multiple degrees.
For example, in the relative reading of \nl{state bordering the largest state},
we would have a degree for the size of each neighboring state.

Superlatives use the \wl{argmax} and \wl{argmin} predicates,
which are defined in \refsec{worlds}.
Comparatives use the \wl{more} and \wl{less} predicates:
$w(\wl{more})$ contains triples $(S,x,y)$, where $x$ is ``more than'' $y$ as measured by $S$;
$w(\wl{less})$ is defined analogously:
\begin{align}
w(\wl{more}) &= \{ (S,x,y) : \max S(x) > \max S(y) \}, \\
w(\wl{less}) &= \{ (S,x,y) : \min S(x) < \min S(y) \}.
\end{align}

% Setup (DCS tree, and denotation)
We use the same mark relation $\CompareRel$ for both comparative and superlative constructions.
In terms of the DCS tree, there are three key parts:
(i) the root $x$, which corresponds to the entity to be compared,
(ii) the child $c$ of a $\CompareRel$ relation, which corresponds to the comparative or superlative predicate,
and (iii) $c$'s parent $p$, which contains the ``degree information'' (which will be described later) used for comparison.
We assume that the root is marked (usually with a relation $\ExtractRel$).
This forces us to compute a comparison degree for each value of the root node.
In terms of the denotation $d$ corresponding to the DCS tree prior to execution,
the entity to be compared occurs in column 1 of the arrays $d.A$,
the degree information occurs in column $i$ of the arrays $d.A$,
and the denotation of the comparative or superlative predicate itself is the child denotation at column $i$
($d.c_i$).

First, we define a concatenating function $\ConcatOp{\bi}{d}$, which combines the columns $\bi$ of $d$
by concatenating the corresponding tuples of each array in $d.A$:
\begin{align}
\ConcatOp{\bi}{\sem{A;\bstore}} &= \sem{A';\bstore'}, \text{where} \label{eqn:concat} \\
      A' & = \{ \ba_{(1 \dots i_1})\backslash\bi + \arr{a_{i_1}+\cdots+a_{i_{|\bi|}}} + \ba_{(i_1\dots n)\backslash\bi} : \ba \in A \} \nonumber \\
\bstore' & = \bstore_{(1 \dots i_1})\backslash\bi + \arr{\store_{i_1}} + \bstore_{(i_1\dots n)\backslash\bi}. \nonumber
\end{align}
Note that the store of column $i_1$ is kept and the others are discarded.
As an example:
\begin{align}
\ConcatOp{2,1}{\sem{\{\arr{(1),(2),(3)},\arr{(4),(5),(6)}\};\store_1,\store_2,\store_3}} = 
\sem{\{\arr{(2,1),(3)},\arr{(5,4),(6)}\};\store_2,\store_3}.
\end{align}

We first create a denotation $d'$ where column $i$, which contains the degree information,
is extracted to column 1 (and thus column 2 corresponds to the entity to be compared).
Next, we create a denotation $d_S$ whose column 1 contains a set of entity-degree pairs.
There are two types of degree information:
\begin{enumerate}
\item Suppose the degree information has arity 2 ($\arity{\ProjOp{i}{d.A}} = 2$).
This occurs, for example, in \nl{most populous city} (see \reffig{examples}(f)),
where column $i$ is the \wl{population} node.
In this case, we simply set the degree to the second component of \wl{population}
by projection ($\JoinProjOp{1,2}{\densyn{\nil}}{d'}$).
Now columns 1 and 2 contain the degrees and entities, respectively.
We concatenate columns 2 and 1 ($\ConcatOp{2,1}{\cdot}$) and aggregate to produce a denotation $d_S$ which
contains the set of entity-degree pairs in column 1.

\item Suppose the degree information has arity 1 ($\arity{\ProjOp{i}{d.A}} = 1$).
This occurs, for example, in \nl{state bordering the most states} (see \reffig{examples}(e)),
where column $i$ is the lower marked \wl{state} node.
In this case, the degree of an entity from column 2 is the number of different values that column 1 can take.
To compute this, aggregate the set of values ($\SumOp{d'}$)
and apply the \wl{count} predicate. % (by joining the first component and projecting onto the second component).
Now with the degrees and entities in columns 1 and 2, respectively,
we concatenate the columns and aggregate again to obtain $d_S$.
\end{enumerate}
Having constructed $d_S$, we simply apply the comparative/superlative predicate which has been patiently
waiting in $d.c_i$. % by joining $d_S$ with the first component and projecting onto the second component.
Finally, the store of $d$'s column 1 was destroyed by the concatenation operation $\ConcatOp{2,1}(\cdot)$,
so we must restore it with $\cdot\with{\store_1 = d.\store_1}$.
The complete operation is as follows:
\begin{align}
\bX_i(d) &= \p{\JoinProjOp{1,2}{\den{\syn{\nil}}}{\p{\JoinProjOp{1,1}{d.c_i}{d_S}}}}\with{\store_1=d.\store_1} \text{ if $d.\store_i = \CompareRel, d.\store_1 \neq \nil$, where} \label{eqn:executeCompare} \\
d_S &= \begin{cases}
\SumOp{\ConcatOp{2,1}{\JoinProjOp{1,2}{\densyn{\nil}}{d'}}}                                                & \text{if $\arity{\ProjOp{i}{d.A}} = 2$} \\
\SumOp{\ConcatOp{2,1}{\JoinProjOp{1,2}{\densyn{\nil}}{\p{\JoinProjOp{1,1}{\densyn{\Count}}{\SumOp{d'}}}}}} & \text{if $\arity{\ProjOp{i}{d.A}} = 1$}, \\
\end{cases} \\
d' &= \ProjOp{-\nil}{\ProjOp{i,-i}{d}}\with{\store_1 = \nil}.
\end{align}
An example of executing the $\CompareRel$ relation is shown in \reffig{compare}(a).
As with executing a $\QuantRel$ relation, for simple cases, we can think of executing a $\CompareRel$ relation
as expanding a DCS tree, as shown in \reffig{compare}(b).

% Comparatives - generalized superlatives
\reffig{examples}(e) and \reffig{examples}(f) show examples of superlative constructions
with the arity 1 and arity 2 types of degree information, respectively.
\reffig{examples}(g) shows an example of an comparative construction.
Comparatives and superlatives use the same machinery,
differing only in the predicate:
\wl{argmax} versus \wl{\syn{\wl{more}; \JoinRel{3}{1} \cto {\wl{TX}}}} (\nl{more than Texas}).
%\footnote{
%Note that the DCS tree we construct for \nl{state bordering more states than Texas}
%is no longer projective; we return to this issue in discussing the construction mechanism \refsec{construction}.}).
But both predicates have the same template behavior:~Each takes a set of entity-degree pairs and returns any entity satisfying some property.
For $\wl{argmax}$, the property is obtaining the highest degree;
for $\wl{more}$, it is having a degree higher than a threshold.
We can handle generalized superlatives (\nl{the five largest} or \nl{the fifth largest} or \nl{the 5\% largest})
as well by swapping in a different predicate; the execution mechanisms defined in \refeqn{executeCompare}
remain the same.

% Relative/absolute
We saw that the mark-execute machinery allows decisions regarding quantifier scope to made in a clean and modular fashion.
Superlatives also have scope ambiguities in the form of absolute versus relative readings.
Consider the example in \reffig{examples}(g).
In the absolute reading, we first compute the superlative in a narrow scope (\nl{the largest state} is Alaska),
and then connect it with the rest of the phrase, resulting in the empty set (since no states border Alaska).
In the relative reading, we consider the first \nl{state} as the entity we want to compare,
and its degree is the size of a neighboring state.
In this case, the lower \wl{state} node cannot be set to Alaska because there are no states bordering it.
The result is therefore any state that borders Texas (the largest state that does have neighbors).
The two DCS trees in \reffig{examples}(g) show that 
we can naturally account for this form of superlative ambiguity
based on where the scope-determining execute relation is placed without drastically changing
the underlying tree structure.

\paragraph{Remarks} 

All these issues are not specific to DCS; every serious semantic formalism must address them as well.
Not surprisingly then, the mark-execute construct bears some resemblance to other mechanisms
that operate on categorial grammar and lambda calculus, such as
quantifier raising, Montague's quantifying in, Cooper storage, and Carpenter's scoping constructor \citep{carpenter98type}.
Very broadly speaking, these mechanisms delay application of the divergent element (usually a quantifier),
``marking'' its spot with a dummy pronoun (as in Montague's quantifying in) or in a store (as in Cooper storage),
and then ``executing'' the quantifier at a later point in the derivation.
One subtle but important difference between mark-execute in DCS and the others
is that a DCS tree (which contains the mark and execute relations) is the final logical form,
and all the action happens in the computing of the denotation of this logical form.
In more traditional approaches, the action happens in the construction mechanism for building the logical form;
the actually logical form produced at the end of the day is quite simple.
In other words, we have pushed the inevitable complexity from the construction mechanism into the semantics of the logical from.
This refactoring is important because we want our construction mechanism to focus not on linguistic issues,
but on purely computational and statistical ones, which ultimately determine the 
practical success of our system.

%%%%%%%%%%%%%%%%%%%%%%%%%%%%%%
\SecTwo{construction}{Construction Mechanism}

% Setup
We have thus far defined the syntax (\refsec{syntax}) and semantics
(\refsec{full}) of DCS trees, but we have only vaguely hinted at how these DCS
trees might be connected to natural language utterances by appealing to idealized examples.
In this section, we formally define the construction mechanism for DCS,
which takes an utterance $\bx$ and produces a set of DCS trees $\sZ_L(\bx)$.

% Complications
Since we motivated DCS trees based on dependency syntax, it might be tempting to take a
dependency parse tree of the utterance, replace the words with predicates,
and attach some relations on the edges to produce a DCS tree.
To a first approximation, this is what we will do,
but we need to be a bit more flexible for several reasons:
(i) some nodes in the DCS tree do not have predicates (e.g., children of a $\ExtractRel$ relation
or parent of an $\X_\bi$ relation);
(ii) nodes have predicates that do not correspond to words (e.g., in \nl{California cities},
there is a implicit \wl{loc} predicate that bridges \wl{CA} and \wl{city});
(iii) some words might not correspond to any predicates in our world (e.g., \nl{please});
and (iv) the DCS tree might not always be aligned with the syntactic structure
depending on which syntactic formalism one ascribes to.
While syntax was the inspiration for the DCS formalism,
we will not actually use it in construction.

% Overapproximation
It is also worth stressing the purpose of the construction mechanism.  In
linguistics, the purpose of the construction mechanism is to try to generate
the exact set of valid logical forms for a sentence.
We view the construction mechanism instead as simply
a way of creating a set of candidate logical forms.  A separate step defines
a distribution over this set to favor certain logical forms over others.
The construction mechanism should therefore overapproximate
the set of logical forms.  Settling for an overapproximation
allows us to simplify the construction mechanism.

%This is reflected in
%semantic formalisms such as CCG.  CCG maps words to lambda calculus
%expressions, which are then combined using a small number of combinators such
%as function application.  These combinators operate deterministically and
%therefore the whole construction procedure is quite rigid given the lexicon
%(mapping from words to lambda calculus expressions).  Therefore, if we have the
%right lexicon, CCG will not overgenerate or undergenerate by very much.
%However, in our learning-based setting, we do not start with the right lexicon,
%and furthermore, we must learn it from only question-answer pairs.  Therefore,

%%%%%%%%%%%%%%%%%%%%%%%%%%%%%%
\SecThree{lexicalTriggers}{Lexical Triggers}

The construction mechanism assumes
a fixed set of {\em lexical triggers} $L$.
Each trigger is a pair $(\bs,p)$, where $\bs$ is a sequence of words (usually one) and $p$ is a
predicate (e.g., $\bs = \nl{California}$ and $p = \wl{CA}$).
We use $L(\bs)$ to denote the set of predicates $p$ triggered by $\bs$ ($(\bs,p) \in L$).
We also define a set of {\em trace predicates}, denoted by $L(\epsilon)$,
which can be introduced without an overt lexical trigger.

It is important to think of the lexical triggers $L$ not as pinning down the precise predicate
for each word.  For example, $L$ might contain
$\{ (\nl{city},\wl{city}), (\nl{city},\wl{state}), (\nl{city},\wl{river}), \dots \}$.
\refsec{lexicalTriggerExperiments} describes the lexical triggers that we use in our experiments.

%%%%%%%%%%%%%%%%%%%%%%%%%%%%%%
\SecThree{recursiveConstruction}{Recursive Construction of DCS Trees}

Given a set of lexical triggers $L$,
we will now describe a recursive mechanism
for mapping
an utterance $\bx = (x_1, \dots, x_n)$ to 
$\sZ_L(\bx)$, a set of candidate DCS trees for $\bx$.
The basic approach is reminiscent of projective labeled dependency parsing:
For each span $i..j$ of the utterance, we build a set of trees $C_{i,j}(\bx)$.
The set of trees for the span $0..n$ is the final result:
\begin{align}
\sZ_L(\bx) = C_{0,n}(\bx). \label{eqn:candidateSet}
\end{align}

\FigTop{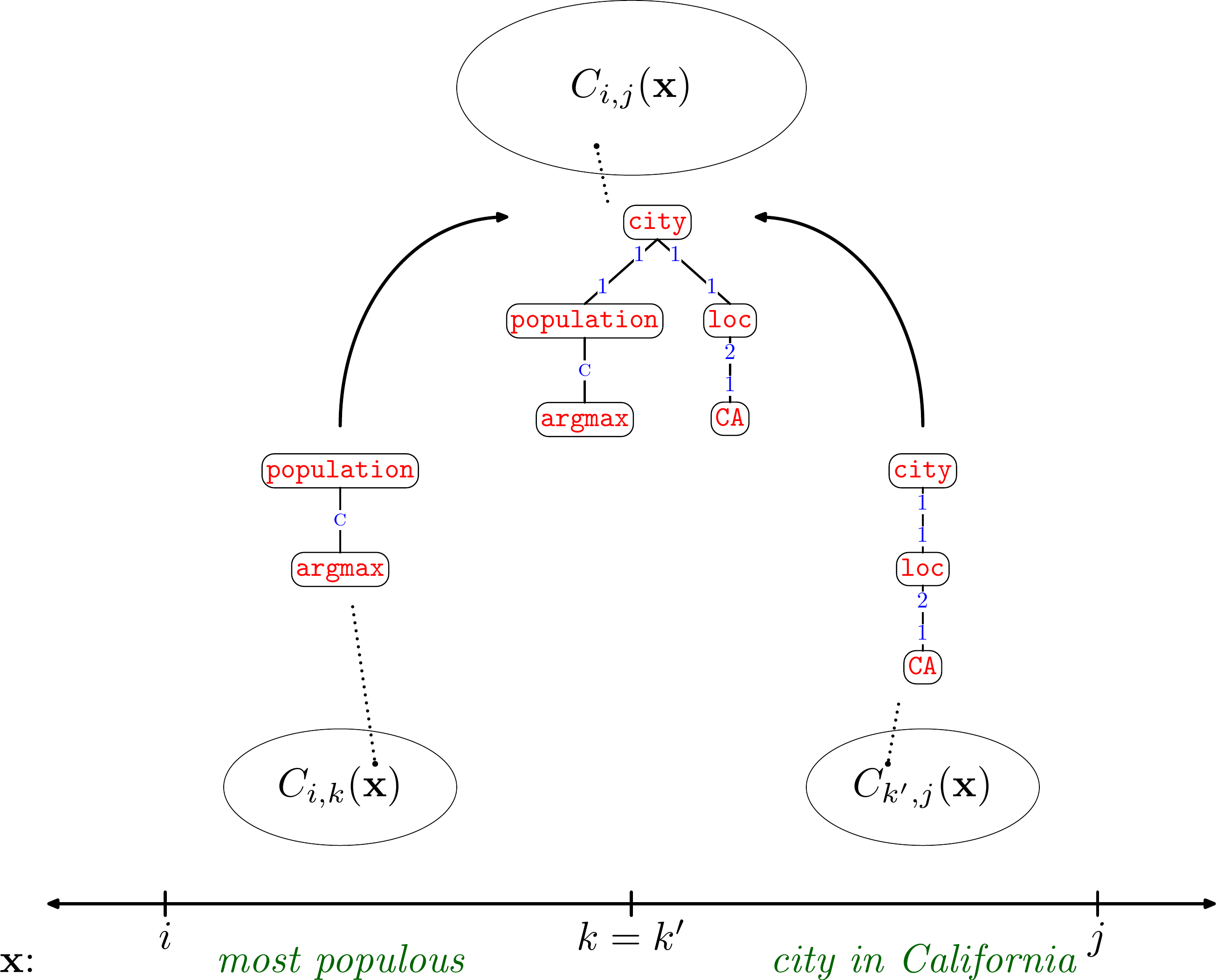}{0.35}{recursiveConstruction}{
Shows an example of the recursive construction of $C_{i,j}(\bx)$, a set of DCS trees for span $i..j$.
}

Each set of DCS trees $C_{i,j}(\bx)$ is constructed
recursively by combining the trees of its subspans $C_{i,k}(\bx)$ and $C_{k',j}(\bx)$
for each pair of split points $k,k'$ (words between $k$ and $k'$ are ignored).
These combinations are then augmented via a function $A$ and filtered via a function $F$;
these functions will be specified later.
Formally, $C_{i,j}(\bx)$ is defined recursively as follows:
\begin{align}
\label{eqn:Cij}
C_{i,j}(\bx) &= F \Big( A\Big( \{ \syn{p}_{i..j} : p \in L(\bx_{i+1..j}) \} \cup \bigcup_{\substack{i \le k \le k' < j \\ a \in C_{i,k}(\bx) \\ b \in C_{k',j}(\bx)}} \cand_1(a,b)) \Big) \Big).
\end{align}
This recurrence has two parts:
\begin{itemize}
\item The base case: we take the phrase (sequence of words) over span $i..j$ and look up the set of predicates $p$ in the set
of lexical triggers.  For each predicate, we construct a one-node DCS tree.
We also extend the definition of DCS trees in \refsec{syntax} to allow each node to store the indices of the span $i..j$ that
triggered the predicate at that node;
this is denoted by $\syn{p}_{i..j}$.
This span information will be useful in \refsec{features},
where we will need to talk about how an utterance $\bx$ is aligned with a DCS tree $z$.

\item The recursive case: $\cand_1(a,b)$, which we will define shortly, that takes two DCS trees,
$a$ and $b$, and returns a set of new DCS trees formed by combining $a$ and $b$.
\reffig{recursiveConstruction} shows this recurrence graphically.
\end{itemize}

We now focus on how to combine two DCS trees.
Define $\cand_d(a,b)$ as the set of DCS trees that
result by making either $a$ or $b$ the root and connecting the other via a chain of relations and at most $d$ trace predicates:
\begin{align}
\cand_d(a,b) = \rcand_d(a,b) \cup \lcand_d(b,a), \label{eqn:candd}
\end{align}
Here, $\rcand_d(a,b)$ is the set of DCS trees where $a$ is the root;
for $\lcand_d(a,b)$, $b$ is the root.
The former is defined recursively as follows:
\begin{align}
\rcand_0(a,b) &= \emptyset, \\
\rcand_d(a,b) &= \bigcup_{\substack{r \in \sR \\p \in L(\epsilon)}} \{ \syn{a; r \cto b}, \syn{a; r \cto \syn{\SigmaRel \cto b}} \} \cup \rcand_{d-1}(a, \syn{p; r \cto b}). \nonumber
\end{align}
First, we consider all possible relations $r \in \sR$ and try putting $r$ between $a$ and $b$ ($\syn{a;r \cto b}$),
possibly with an additional aggregate relation ($\syn{a; r \cto \syn{\SigmaRel \cto b}}$).
Of course, $\sR$ contains an infinite number of join and execute relations,
but only a small finite number of them make sense:
we consider join relations $\JoinRel{j}{j'}$ only for $j \in \{1, \dots, \arity{a.p} \}$ and $j' \in \{1, \dots, \arity{b.p}\}$,
and execute relations $\X_\bi$ for which $\bi$ does not contain indices larger than the number of columns of $\den{b}$.
Next, we further consider all possible trace predicates $p \in L(\epsilon)$,
and recursively try to connect $a$ with the intermediate $\syn{p;r \cto b}$,
now allowing $d-1$ additional predicates.
See \reffig{localConstruction} for an example.
\FigTop{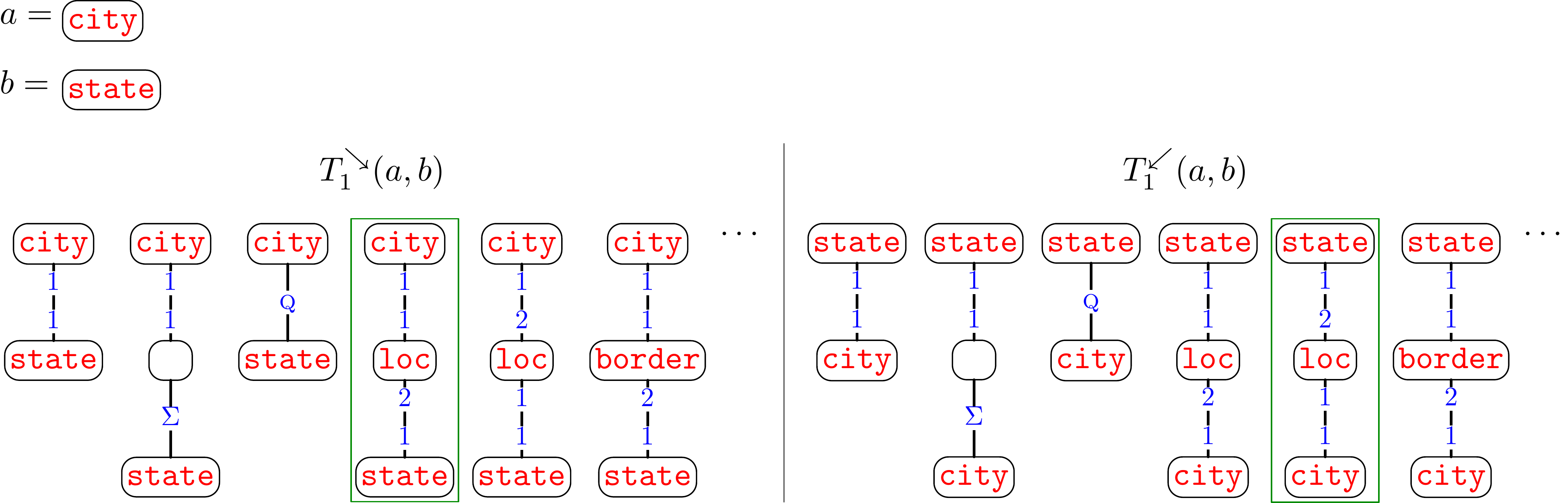}{0.33}{localConstruction}{
Given two DCS trees, $a$ and $b$, $\rcand_1(a,b)$ and $\lcand_1(a,b)$ are the two sets of DCS trees
formed by combining $a$ and $b$ with $a$ at the root and $b$ at the root, respectively;
one trace predicate can be inserted in between.
In this example, the DCS trees which survive filtering (\refsec{filtering}) are shown.
}
In the other direction, $\lcand_d$ is defined similarly:
\begin{align}
\lcand_0(a,b) &= \emptyset, \\
\lcand_d(a,b) &= \bigcup_{\substack{r \in \sR \\p \in L(\epsilon)}} \{ \syn{b.p; r \cto a; b.\be}, \syn{b.p; r \cto \syn{\SigmaRel \cto a}; b.\be} \} \cup \rcand_{d-1}(a, \syn{p; r \cto b}). \nonumber
\end{align}

% Inserting trace predicates
Inserting trace predicates allows us to build logical forms with more predicates
than are explicitly triggered by the words.  This ability is useful for several reasons.
Sometimes, there is a predicate not overtly expressed, especially in noun compounds (e.g., \nl{California cities}).
For semantically light words such as prepositions (e.g., \nl{for})
it is difficult to enumerate all the possible predicates that it might trigger;
it is simpler computationally to try to insert trace predicates.
We can even omit lexical triggers for transitive verbs such as \nl{border}
because the corresponding predicate \wl{border} can be inserted as a trace predicate.
%Indeed, we are being very permissive here and by itself, this is a terrible construction mechanism.
%However, remember that our goal is just to generate an overapproximation, which can be further refined by a probabilistic step.

% Augmentation
The function $\cand_1(a,b)$ connects two DCS trees via a path of relations and trace predicates.
The augmentation function $A$ adds additional relations (specifically, $\ExtractRel$ and/or $\X_\bi$) on a single DCS tree:
\begin{align}
%A(Z) = \bigcup_{z \in Z} \{ \syn{z ; \ExtractRel \cto \syn{\nil}} \} \cup \{ \syn{\X_\bi \cto z} : \X_\bi \in \sR \},
A(Z) = \bigcup_{\substack{z \in Z \\ \X_\bi \in \sR}} \{ z, \syn{z ; \ExtractRel \cto \syn{\nil}}, \syn{\X_\bi \cto z}, \syn{\X_\bi \cto \syn{z; \ExtractRel \cto \syn{\nil}}} \},
\end{align}

%%%%%%%%%%%%%%%%%%%%%%%%%%%%%%
\SecThree{filtering}{Filtering using Abstract Interpretation}

The construction procedure as described thus far is extremely permissive, generating many
DCS trees which are obviously wrong---for example, $\syn{\state; \JoinRel{1}{1}
\cto \syn{\wl{>}; \JoinRel{2}{1} \syn{3}}}$, which tries to compare a state with
the number 3.  There is nothing wrong this expression syntactically: its
denotation will simply be empty (with respect to the world).  But semantically,
this DCS tree is anomalous.

We cannot simply just discard DCS trees with empty denotations,
because we would incorrectly rule out $\syn{\state; \JoinRel{1}{1} \cto
\syn{\wl{border}; \JoinRel{2}{1} \syn{\wl{AK}}}}$.  The difference here is that
even though the denotation is empty in this world, it is possible that it might
not be empty in a different world where history and geology took
another turn, whereas it is simply impossible to compare cities and numbers.

Now let us quickly flesh out this intuition before falling into a philosophical discussion about possible worlds.
Given a world $w$, we define an abstract world $\Abstract(w)$, to be described shortly.
We compute the denotation of a DCS tree $z$ with respect to this abstract world.
If at any point in the computation we create an empty denotation, we judge $z$ to be impossible and throw it away.
The filtering function $F$ is defined as follows:\footnote{
To further reduce the search space,
$F$ imposes a few additional constraints, 
e.g., limiting the number of columns to 2, and
only allowing trace predicates between arity 1 predicates.}
\begin{align}
F(Z) = \{ z \in Z : \forall z' \text{ subtree of z }, \denabs{z'}.A \neq \emptyset \}.
\end{align}

Now we need to define the abstract world $\Abstract(w)$.  The intuition is to map
concrete values to abstract values: $3\cto\wl{length}$ becomes $\wl{length}$, $\nl{Oregon}\cto\wl{state}$ becomes $*\cto\wl{state}$,
and in general, primitive value $x\cto t$ becomes $*\cto t$.
We perform abstraction on tuples componentwise, so that $(\nl{Oregon}\cto\wl{state}, 3\cto\wl{length})$ 
becomes $(*\cto\wl{state},*\cto\wl{length})$.
Our abstraction of sets is slightly more complex:
the empty set maps to the empty set,
a set containing values all with the same abstract value $a$ maps to $\{a\}$,
and a set containing values with more than one abstract value maps to a $\{\ts{mixed}\}$.
Finally, a world maps each predicate onto a set of (concrete) tuples;
the corresponding abstract world maps each predicate onto the set of abstract tuples.
Formally, the abstraction function is defined as follows:
\begin{align}
\Abstract(x:t) &= *:t, &\aside{primitive values} \\
\Abstract((v_1,\dots,v_n)) &= (\Abstract(v_1),\dots,\Abstract(v_n)), &\aside{tuples} \\
\Abstract(A) &= \begin{cases}
\emptyset & \text{if $A = \emptyset$}, \\
\{\Abstract(x) : x \in A\} & \text{if $|\{\Abstract(x) : x \in A\}| = 1$}, \\
\{\text{\sc{mixed}}\} & \text{otherwise}.
\end{cases} &\aside{sets} \\
\Abstract(w) &= \lambda p . \{ \Abstract(x) : x \in w(p)  \}. &\aside{worlds}
\end{align}

% Examples
As an example, the abstract world might look like this:
\begin{align}
\Abstract(w)(\wl{>}) &= \{ (*\cto\wl{number}, *\cto\wl{number}, *\cto\wl{number}), (*\cto\wl{length}, *\cto\wl{length}, *\cto\wl{length}), \dots \}, \\
\Abstract(w)(\wl{state}) &= \{ (*\cto\wl{state}) \}, \\
\Abstract(w)(\wl{AK}) &= \{ (*\cto\wl{state}) \}, \\
\Abstract(w)(\wl{border}) &= \{ (*\cto\wl{state},*\cto\wl{state}) \}.
\end{align}
Now returning our motivating example at the beginning of this section,
we see that the bad DCS tree has an empty abstract denotation
$\denabs{\syn{\state; \JoinRel{1}{1} \cto \syn{\wl{>}; \JoinRel{2}{1} \syn{3}}}} = \sem{\emptyset;\nil}$.
The good DCS tree has an non-empty abstract denotation:
$\denabs{\syn{\state; \JoinRel{1}{1} \cto
\syn{\wl{border}; \JoinRel{2}{1} \syn{\wl{AK}}}}} = \sem{\{(*\cto\wl{state})\};\nil}$,
as desired.

\paragraph{Remarks} 
% Abstract interpretation
Computing denotations on an abstract world is called {\em abstract
interpretation} \citep{cousot77abstract} and is very powerful framework commonly
used in the programming languages community.  The idea is to obtain information
about a program (in our case, a DCS tree) without running it concretely, but
rather just by running it abstractly.  It is closely related to type systems, but
the type of abstractions one uses is often much richer than standard type
systems.

%%%%%%%%%%%%%%%%%%%%%%%%%%%%%%
\SecThree{ccgComparison}{Comparison with CCG}

\FigTop{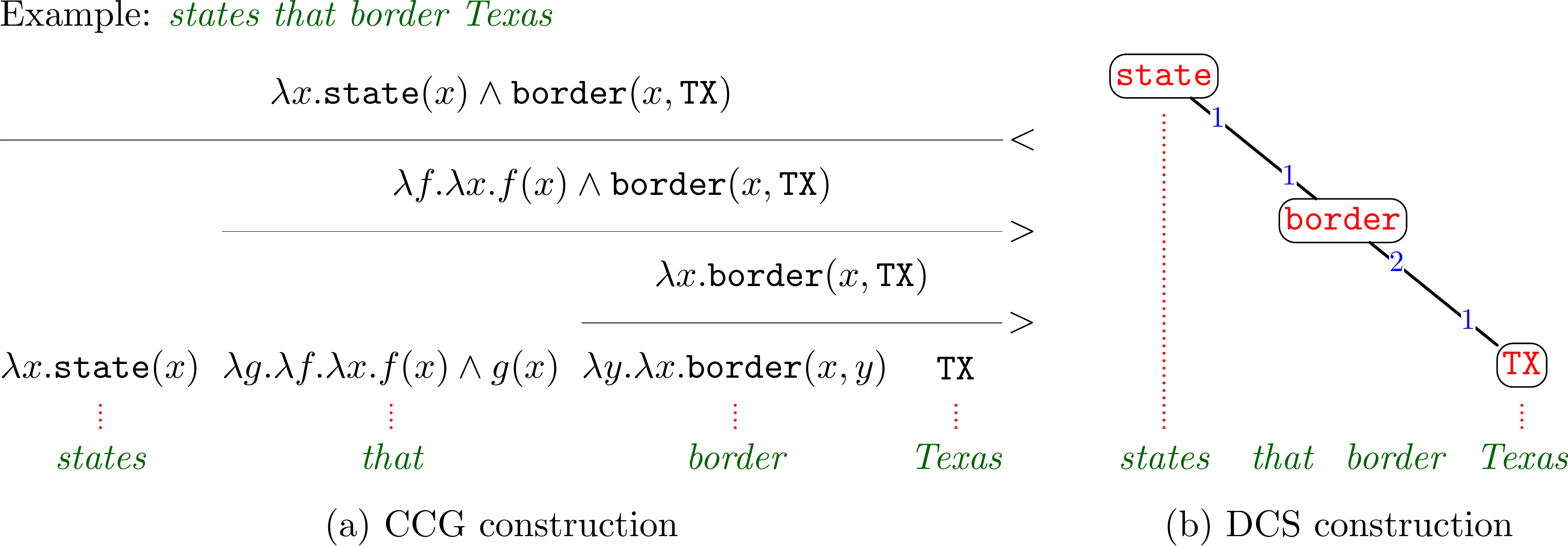}{0.35}{ccgdcs}{
Comparison between the construction mechanisms of CCG and DCS.
There are three principal differences:
First, in CCG, words are mapped onto a lambda calculus expression; in DCS,
words are just mapped onto a predicate.
Second, in CCG, lambda calculus expressions are built by combining (e.g., via function application)
two smaller expressions;
in DCS, trees are combined by inserting relations (and possibly other predicates between them).
Third, in CCG, all words map to a logical expression;
in DCS, only a small subset of words (e.g., \nl{state} and \wl{Texas}) map to predicates;
the rest participate in features for scoring DCS trees.
}

We now compare our construction mechanism with CCG (see \reffig{ccgdcs} for an example).
% Lexical triggers
The main difference is that our lexical triggers contain less information than a lexicon in a CCG.
In CCG, the lexicon would have an entry such as
\begin{align}
\nl{major} \vdash \ts{n} / \ts{n} : \lambda f . \lambda x . \wl{major}(x) \wedge f(x), \label{eqn:ccgLexiconEntry}
\end{align}
which gives detailed information about how this word should interact with its context.
However, in DCS construction, each lexical trigger only has the minimal amount of information:
\begin{align}
\nl{major} \vdash \wl{major}. \label{eqn:dcsLexicalTrigger}
\end{align} 
A lexical trigger specifies a pre-theoretic ``meaning'' of a word which
does not commit to any formalisms.  One advantage of this minimality is that
lexical triggers could be easily obtained from non-expert supervision: One
would only have to associate words with database table names (predicates).

In some sense, the DCS construction mechanism pushes the complexity out of the lexicon.
In linguistics, this complexity usually would
end up in the grammar, which would be undesirable.
However, we do not have to respect this tradeoff, because 
the construction mechanism only produces an overapproximation, which means it
is possible to have both a simple ``lexicon'' and a simple ``grammar.''

% Underspecification - important
There is an important but subtle rationale for this design decision.
During learning, we never just have one clean lexical entry per word (as is typically assumed in formal linguistics).
Rather, there are often many possible lexical entries
(and to handle disfluent utterances or utterances in free word-order languages,
we might actually need many of them \citep{kwiatkowski10ccg,kwiatkowski11lex}):
\begin{align}
& \nl{major} \vdash \ts{n} : \lambda x . \wl{major}(x) \\
& \nl{major} \vdash \ts{n} / \ts{n} : \lambda f . \lambda x . \wl{major}(x) \wedge f(x) \\
& \nl{major} \vdash \ts{n} \backslash \ts{n} : \lambda f . \lambda x . \wl{major}(x) \wedge f(x) \\
& \dots
\end{align}
Now think of a DCS lexical trigger $\nl{major} \vdash \wl{major}$
as simply a {\em compact representation} for a set of CCG lexical entries.
Furthermore, the choice of the lexical entry is made not at the initial lexical base case,
but rather during the recursive construction
by inserting relations between DCS subtrees.
It is exactly at this point that the choice {\em can} be made, because after all,
the choice is one that depends on context. %, which is available during the recursive construction.
The general principle is to compactly represent the indeterminacy until one can resolve it.

%Our construction mechanism for DCS starts with the words and builds DCS trees in a bottom-up fashion.
%In this way, the construction mechanism works like that of CCG, which also constructs
%lambda calculus expressions bottom-up.  However, CCG is based on function application (and generalizations thereof)
%while DCS is based on joining predicates---again highlighting the applicative versus declarative nature of the two formalisms.

% Type raising
Type raising is a combinator in CCG that turns one logical form into another.
It can be used to turn one entity into a related entity (a kind of generalized metonymy).  
For example, \citet{zettlemoyer07relaxed}
used it to allow conversion from $\wl{Boston}$ to $\lambda x . \wl{from}(x, \wl{Boston})$.
Type raising in CCG is analogous to inserting trace predicates in DCS, but there is an important distinction:
Type raising is a unary operation and is unconstrained in that it changes logical forms
into new ones without regard for how they will interact with the context.
Inserting trace predicates is a binary operation which is constrained by the two predicates that it is mediating.
In the example, \wl{from} would only be inserted to combine \wl{Boston} with \wl{flight}.
This is another instance of the general principle of delaying uncertain decisions until there is more information.

\SecOne{learning}{Learning}

In \refchp{representation}, we defined DCS trees and a construction mechanism for producing
a set of candidate DCS trees given an utterance.
We now define a probability distribution over that set (\refsec{model})
and an algorithm for estimating the parameters (\refsec{parameterEstimation}).
%This semantic parsing model is a standard log-linear distribution
%defined in terms of local features over the utterance and DCS tree.
%Then in \refsec{parameterEstimation}, we present an algorithm for estimating the parameters of the model.
The number of candidate DCS trees grows exponentially, so we use beam search to control this growth.
The final learning algorithm alternates between beam search and
optimization of the parameters, leading to a natural bootstrapping
procedure which integrates learning and search.

%%%%%%%%%%%%%%%%%%%%%%%%%%%%%%%%%%%%%%%%%%%%%%%%%%%%%%%%%%%%
\SecTwo{model}{Semantic Parsing Model}

The semantic parsing model specifies a conditional distribution over a set of candidate
DCS trees $C(\bx)$ given an utterance $\bx$.
This distribution depends on a function $\phi(\bx,z) \in \R^d$,
which takes a $(\bx,z)$ pair and extracts a set of local features (see \refsec{features} for a full specification).
Associated with this feature vector is a parameter vector $\theta \in \R^d$.
The inner product between the two vectors, $\phi(\bx,z)^\top\theta$, yields
a numerical score, which intuitively measures the compatibility of the utterance $\bx$ with the DCS tree $z$.
%Let $C(\bx)$ be a function which maps an utterance $\bx$ to a set of candidate DCS trees;
%for example, $C(\bx) = \sZ_L(\bx)$ (defined in \refeqn{Cij}), but later we will consider a different
%one for beam search.
We exponentiate the score and normalize over $C(\bx)$ to obtain a proper probability distribution:
\begin{align}
p(z \mid \bx; C, \theta) &= \exp \{ \phi(\bx,z)^\top \theta - \LogZ(\theta; \bx, C) \}, \\
\LogZ(\theta; \bx, C) &= \log \sum_{z \in C(\bx)} \exp \{ \phi(\bx,z)^\top \theta \},
\end{align}
where $\LogZ(\theta; \bx, C)$ is the log-partition function with respect to the 
candidate set function $C(\bx)$.

%%%%%%%%%%%%%%%%%%%%%%%%%%%%%%
\SecThree{features}{Features}

% Features: generalize
We now define the feature vector $\phi(\bx,z) \in \R^d$, the core part of the semantic parsing model.
Each component $j = 1, \dots, d$ of this vector is a feature,
and $\phi(\bx,z)_j$ is the number of times that feature occurs in $(\bx,z)$.
Rather than working with indices,
we treat features as symbols (e.g., $\WordTriggerPred[\nl{states}, \wl{state}]$).
Each feature captures some property about $(\bx,z)$
which abstracts away from the details of the specific instance
and allow us to generalize to new instances that share common features.

\FigTop{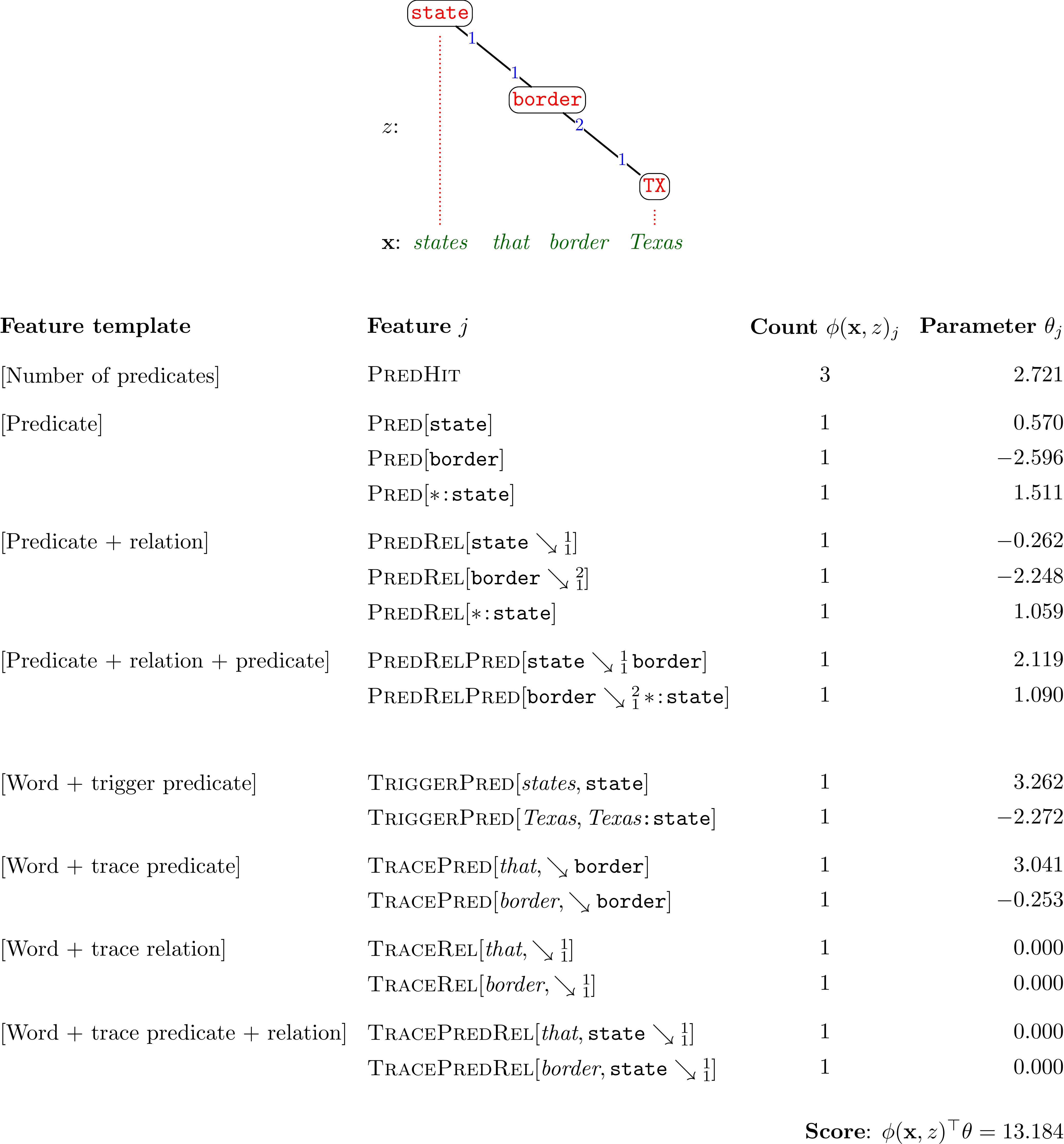}{0.32}{features}{For each utterance-DCS tree pair $(\bx,z)$,
we define a feature vector $\phi(\bx,z)$, whose $j$-th component is the number of times
a feature $j$ occurs in $(\bx,z)$.  Each feature has an associated parameter $\theta_j$,
which is estimated from data in \refsec{parameterEstimation}.
The inner product of the feature vector and parameter vector yields a compatibility score.
}
% Feature templates
The features are organized into feature templates,
where each feature template instantiates a set of features. 
%For example, the $\Pred$ feature template corresponds to 
%the features $\{ \Pred[p] : p \in \sP \}$, one for each predicate $p$.
\reffig{features} shows all the feature templates for a concrete example.
The feature templates are as follows:
\begin{itemize}
\item $\PredCount$ contains the single feature $\PredCount$, which fires for each predicate in $z$.
\item $\Pred$ contains features $\{ \Pred[\alpha(p)] : p \in \sP \}$,
each of which fires on $\alpha(p)$, the abstraction of predicate $p$, where
\begin{align}
\alpha(p) = \begin{cases}
* \cto t & \text{if $p = x \cto t$} \\
p & \text{otherwise}.
\end{cases}
\end{align}
The purpose of the abstraction is to abstract away the details of
concrete values such as $\wl{TX} = \nl{Texas}\cto\wl{state}$.

\item $\PredRel$ contains features $\{ \PredRel[\alpha(p),\bq] : p \in \sP, \bq \in (\{ \swarrow, \searrow \} \times \sR)^* \}$.
A feature fires when a node $x$ has predicate $p$ and is connected via some path $\bq = (d_1,r_1), \dots, (d_m,r_m)$
to the lowest descendant node $y$ with the property that each node between $x$ and $y$ has a null predicate.
Each $(d,r)$ on the path represents an edge labeled with relation $r$ 
connecting to a left ($d=\swarrow$) or right ($d=\searrow$) child.
If $x$ has no children, then $m = 0$.  The most common case is when $m=1$,
but $m=2$ also occurs with the aggregate and execute relations
(e.g., $\PredRel[\wl{count}, \searrow \JoinRel{1}{1} \searrow \SigmaRel]$ fires for \reffig{aggregate}(a)).

\item $\PredPred$ contains features $\{ \PredPred[\alpha(p), \bq, \alpha(p')] :
p, p' \in \sP, \bq \in (\{ \swarrow, \searrow \} \times \sR)^* \}$, which are
the same as $\PredRel$, except that we include both the predicate $p$ of $x$ and the
predicate $p'$ of the descendant node $y$.  These features do not fire if $m =
0$.

\item $\WordTriggerPred$ contains features $\{ \WordTriggerPred[\bs,p] : \bs \in W^*, p \in \sP \}$,
where $W = \{ \nl{it}, \nl{Texas}, \dots \}$ is the set of words.
Each of these features fires when a span of the utterance with words $\bs$
triggers the predicate $p$---more precisely, when a subtree
$\syn{p;\be}_{i..j}$ exists with $\bs = \bx_{i\!+\!1..j}$.  Note that these
lexicalized features use the predicate $p$ rather than the abstracted version
$\alpha(p)$.

\item $\WordTracePred$ contains features $\{ \WordTracePred[s,p,d] : s \in W^*, p \in \sP, d \in \{\swarrow,\searrow\} \}$,
each of which fires when a trace predicate $p$ has been inserted over a word $s$.
The situation is the following:~Suppose we have a subtree $a$ that ends at position $k$ (there is a predicate in $a$
that is triggered by a phrase with right endpoint $k$)
and another subtree $b$ that begins at $k'$.
Recall that in the construction mechanism \refeqn{candd}, we can insert a trace predicate $p \in L(\epsilon)$ between the roots of $a$ and $b$.
Then, for every word $\bx_j$ in between the spans of the two subtrees ($j = \{ k+1, \dots, k' \}$),
the feature $\WordTracePred[\bx_j,p,d]$ fires ($d = \swarrow$ if $b$ dominates $a$ and $d = \searrow$ if $a$ dominates $b$).

\item $\WordTraceRel$ contains features $\{ \WordTraceRel[s,d,r] : s \in W^*, d \in \{\swarrow,\searrow\}, r \in \sR \}$,
each of which fires when some trace predicate with parent relation $r$ has been inserted over a word $s$.

\item $\WordTracePredRel$ contains features $\{ \WordTracePredRel[s,p,d,r] : s \in W^*, p \in \sP, d \in \{\swarrow,\searrow\}, r \in \sR \}$,
each of which fires when a predicate $p$ is connected via child relation $r$ to some trace predicate over a word $s$.
\end{itemize}

% Similarity to dependency parsing
These features are simple generic patterns which can be applied for
modeling essentially any distribution over sequences and labeled trees---there is nothing specific to DCS at all.  
The first half of the feature templates (\PredCount, \Pred, \PredRel, \PredPred)
capture properties of the tree independent of the utterance,
and are similar to ones used for syntactic dependency parsing.
The other feature templates (\WordTriggerPred, \WordTracePred, \WordTraceRel, \WordTracePredRel)
connect predicates in the DCS tree with words in the utterance,
similar to those in a model of machine translation.
%We believe one advantage of using labeled trees to represent logical forms is that
%it is easy to define features over them, especially given the strong precedent in NLP.
%While all our features are divorced from the semantics of DCS trees,
%an interesting possibility is to define features on the denotations of subtrees,
%which we must compute in any case.
%For example, we could define features that $\wl{city}

% Soft encoding of lexicon
%Recall

%%%%%%%%%%%%%%%%%%%%%%%%%%%%%%%%%%%%%%%%%%%%%%%%%%%%%%%%%%%%
\SecTwo{parameterEstimation}{Parameter Estimation}

We have now fully specified the details of the graphical model in
\reffig{model}: \refsec{model} described semantic parsing and 
\refchp{representation} described semantic evaluation.
Next, we focus on the inferential problem of estimating the 
parameters $\theta$ of the model from data. 

%%%%%%%%%%%%%%%%%%%%%%%%%%%%%%
\SecThree{objective}{Objective Function}

We assume that our learning algorithm is
given a training dataset $\sD$ containing question-answer pairs $(\bx,y)$.
%importantly, we do not observe the logical forms $z$.
Because the logical forms are unobserved,
we work with $\log p(y \mid \bx; C, \theta)$, the marginal log-likelihood of obtaining the correct answer $y$ given an utterance $\bx$.
This marginal log-likelihood sums over all $z \in C(\bx)$ that evaluate to $y$:
%For a candidate set function $C(\bx)$ and a dataset $\sD$,
%let $C^{\sD}(\bx)$ be the restriction to only DCS trees that yield the correct answer
%(that is, $z$ such that $(\bx,\den{z}) \in \sD$):
%\begin{align}
%C^{\sD}(\bx) &= \{ z \in C(\bx) : (\bx,\den{z}) \in \sD \}.
%\end{align}
\begin{align}
\log p(y \mid \bx; C, \theta) &= \log p(z \in C^y(\bx) \mid \bx; C, \theta) \\
&= \LogZ(\theta; \bx, C^y) - \LogZ(\theta, \bx, C), \text{ where} \\
%C^y(\bx) &= \begin{cases} \label{eqn:Cy}
%\{ z \in C(\bx) : \den{z} = y \} & \text{if $\exists z \in C(\bx) \wedge \den{z} = y$}, \\
%C(\bx) & \text{otherwise}.
C^y(\bx) &\eqdef \{ z \in C(\bx) : \den{z} = y \}.
\end{align}
Here, $C^y(\bx)$ is the set of DCS trees $z$ with denotation $y$.
%However, if there are no DCS trees in $C(\bx)$ that produce $y$,
%then we back off to the original candidate set $C^y(\bx) = C(\bx)$.

We call an example $(\bx,y) \in \sD$ {\em feasible} if
the candidate set of $\bx$ contains a DCS tree that evaluates to $y$ ($C^y(\bx) \neq \emptyset$).
Define an objective function $\sO(\theta,C)$ containing two terms:~The first
term is the sum of the marginal log-likelihood over all feasible training
examples.  The second term is a quadratic penalty on the parameters
$\theta$ with regularization parameter $\lambda$.
Formally:
\begin{align}
\sO(\theta,C) & \eqdef \sum_{\substack{(\bx,y) \in \sD \\ C^y(\bx) \neq \emptyset}} \log p(y \mid \bx; C, \theta) - \frac{\lambda}{2} \|\theta\|^2_2 \label{eqn:objective} \\
              & = \sum_{\substack{(\bx,y) \in \sD \\ C^y(\bx) \neq \emptyset}} \p{ \LogZ(\theta; \bx, C^y) - \LogZ(\theta; \bx, C) } - \frac{\lambda}{2} \|\theta\|^2_2. \nonumber
\end{align}

We would like to maximize $\sO(\theta,C)$.
The log-partition function $\LogZ(\theta; \cdot,\cdot)$ is convex,
but $\sO(\theta,C)$ is the difference of two log-partition functions and hence is not concave (nor convex).
Thus we resort to gradient-based optimization.
A standard result is that the derivative of the log-partition function is the expected feature vector \citep{wainwright08varinf}.
Using this, we obtain the gradient of our objective function:
\begin{align}
\label{eqn:gradient}
\frac{\partial \sO(\theta,C)}{\partial \theta} &=
\sum_{\substack{(\bx,y) \in \sD \\ C^y(\bx) \neq \emptyset}} \p{ \E_{p(z \mid \bx; C^y, \theta)}[\phi(\bx,z)] - \E_{p(z \mid \bx; C, \theta)}[\phi(\bx,z)] } - \lambda\theta.
\end{align}
Updating the parameters in the direction of the gradient
would move the parameters towards
the DCS trees that yield the correct answer ($C^y$)
and away from over all candidate DCS trees ($C$).
We can use any standard numerical optimization algorithm that
requires only black-box access to a gradient.
\refsec{effectSettings} will discuss the empirical
ramifications of the choice of optimization algorithm.
%Most of these algorithms are only guaranteed to converge to a local maximum of

% Skip example
%Recall from the definition of \refeqn{Cy} that for a training example
%$(\bx,y)$, when $C(\bx)$ does not contain any DCS tree that produces the
%correct answer $y$, then $C^y(\bx) = C(\bx)$.  In this case, the contribution
%of that example to the gradient \refeqn{gradient} is exactly zero.
%In other words, if the candidate set $C(\bx)$ is not good enough to get the right answer,
%then we just skip that example.

%%%%%%%%%%%%%%%%%%%%%%%%%%%%%%
\SecThree{algorithm}{Algorithm}

Given a candidate set function $C(\bx)$, we can optimize \refeqn{objective} 
to obtain estimates of the parameters $\theta$.
Ideally, we would use $C(\bx) = \sZ_L(\bx)$, the candidate sets from our construction mechanism
in \refsec{construction}, but we quickly run into the problem of computing \refeqn{gradient} efficiently.
Note that $\sZ_L(\bx)$ (defined in \refeqn{candidateSet}) grows exponentially with the length of $\bx$.
This by itself is not a show stopper.  Our features (\refsec{features}) decompose along the edges
of the DCS tree, so it is possible to use dynamic programming\footnote{
The state of the dynamic program would be the span $i..j$ and the head predicate over that span.}
to compute the second expectation $\E_{p(z \mid \bx; \sZ_L, \theta)}[\phi(\bx,z)]$ of \refeqn{gradient}.
The problem is computing the first expectation
$\E_{p(z \mid \bx; \sZ_L^y, \theta)}[\phi(\bx,z)]$, which sums over the subset of candidate DCS trees $z$
satisfying the constraint $\den{z} = y$.  Though this is a smaller set, 
there is no efficient dynamic program for this set since the constraint does
not decompose along the structure of the DCS tree.
Therefore, we need to approximate $\sZ_L^y$, and in fact, we will approximate $\sZ_L$ as well
so that the two expectations in \refeqn{gradient} are coherent.

% Training = test
%Therefore, we need to approximate $\sZ_L^y$.
%In fact, we will approximate $\sZ_L$ as well
%so that the candidate sets 
%The rationale is that $\sZ_L^y$ is available only at training time; only $\sZ_L$ is available at test time.
%There has been some theoretical and empirical evidence (e.g., \citet{wainwright06wrong})
%that suggests that the same inference algorithm (for us, determining the set of candidate DCS trees)
%should be employed at training time as at test time.

Recall that $\sZ_L(\bx)$ was built by recursively constructing a set of DCS trees $C_{i,j}(\bx)$ for each span $i..j$.
In our approximation, we simply use beam search, which truncates each $C_{i,j}(\bx)$ to include the (at most) $K$ DCS trees
with the highest score $\phi(\bx,z)^\top\theta$.
We let $\tilde C_{i,j,\theta}(\bx)$ denote this approximation
and define the set of candidate DCS trees with respect to the beam search:
\begin{align}
\tilde \sZ_{L,\theta}(\bx) = \tilde C_{0,n,\theta}(\bx).
\end{align}

%Specifically, define a function $T_{K,\theta}(Z)$ that takes a set of DCS trees
%and returns the (at most) $K$ highest scoring DCS trees based on some parameter vector $\theta$:
%\begin{align}
%T_{K,\bx,\theta}(Z) = \text{sort-decreasing}(Z, \lambda z . \phi(\bx, z)^\top\theta)[1..K].
%\end{align}
%Now define $\tilde C_{i,j,\theta}(\bx
%\tilde C_{i,j,\theta}(\bx)

%the truncated candidate sets for each span, paralleling \refeqn{Cij}:
%Now define the truncated candidate sets for each span, paralleling \refeqn{Cij}:
%\begin{align}
%\tilde C_{i,j,\theta}(\bx)
%&= T_{K,\bx,\theta}\Big( F \Big( A\Big( \{ \syn{p}_{i..j} : p \in L(\bx_{i+1..j}) \} \cup \bigcup_{\substack{i \le k \le k' < j \\ a \in \tilde C_{i,k,\theta}(\bx) \\ b \in \tilde C_{k',j,\theta}(\bx)}} \cand_1(a,b)) \Big) \Big) \Big), \\
%& \tilde \sZ_{L,\theta}(\bx) = \tilde C_{0,n,\theta}(\bx).
%\end{align}

\begin{figure}
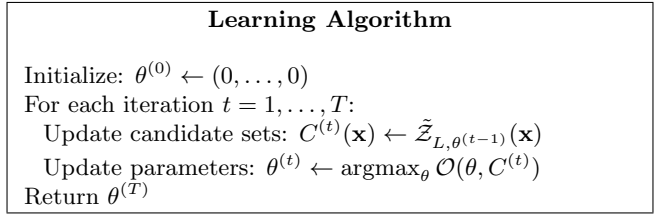

\begin{center} \framebox{ \begin{minipage}{3.2in} % Begin model
\small
\begin{center} {\bf Learning Algorithm} \end{center}
Initialize: $\theta^{(0)} \leftarrow (0, \dots, 0)$ \\
For each iteration $t = 1, \dots, T$: \\
\ind Update candidate sets: $C^{(t)}(\bx) \leftarrow \tilde\sZ_{L,\theta^{(t-1)}}(\bx)$ \\
\ind Update parameters: $\theta^{(t)} \leftarrow \argmax_{\theta} \sO(\theta, C^{(t)})$ \\
Return $\theta^{(T)}$
\end{minipage} } \end{center} % End model
\caption{\label{fig:learningAlgorithm} The learning algorithm
alternates between updating the candidate sets based on beam search
and updating the parameters using standard numerical optimization.
}
\end{figure}

We now have a chicken-and-egg problem:
If we had good parameters $\theta$, we could generate good candidate sets $C(\bx)$
using beam search $\tilde\sZ_{L,\theta}(\bx)$.
If we had good candidate sets $C(\bx)$, we could generate good parameters
by optimizing our objective $\sO(\theta, C)$ in \refeqn{objective}.
This problem leads to a natural solution: simply alternate between the two steps (\reffig{learningAlgorithm}).
This procedure is not guaranteed to converge, due to the
heuristic nature of the beam search, but we have found it to 
be convergent in practice.

Finally, we use the trained model with parameters $\theta$ to answer new questions $\bx$ by
choosing the most likely answer $y$, summing out the latent logical form $z$:
\begin{align}
F_\theta(\bx) &\eqdef \argmax_y p(y \mid \bx; \theta, \tilde\sZ_{L,\theta}) \\
       &= \argmax_y \sum_{\substack{z \in \tilde\sZ_{L,\theta}(\bx) \\ \den{z} = y}} p(z \mid \bx; \theta, \tilde\sZ_{L,\theta}).
\end{align}

\SecOne{experiments}{Experiments}

We have now completed the conceptual part of this \paper---using DCS trees
to represent logical forms (\refchp{representation}),
and learning a probabilistic model over these trees (\refchp{learning}).
In this section, we evaluate and study our approach empirically.
Our main result is that our system obtains higher accuracies than existing systems,
despite requiring no annotated logical forms. % (\refsec{compareResults}).

%%%%%%%%%%%%%%%%%%%%%%%%%%%%%%%%%%%%%%%%%%%%%%%%%%%%%%%%%%%%
\SecTwo{setup}{Experimental Setup}

We first describe the datasets (\refsec{datasets}) that we use to train and evaluate our system.
We then mention various choices in the model and learning algorithm (\refsec{settings}).
One of these choices is the lexical triggers, which is further discussed in \refsec{lexicalTriggerExperiments}.

%%%%%%%%%%%%%%%%%%%%%%%%%%%%%%
\SecThree{datasets}{Datasets}

We tested our methods on two standard datasets, referred to in this
\paper\ as \geo\ and \job.  These datasets were created by Ray Mooney's group
during the 1990s and have been used to evaluate semantic parsers
for over a decade.

\paragraph{US Geography}

% ec2:13.exec (GEO, FINALTEST, lexMode=0)
% Overall: 280 word types, sentence length = 4/ << 8.564~2.425 >> /23 (880)
% 32 used original lexical entries: -BRIDGE- :JJ :NN :NNS :WRB atleast averag call combin count each exclud great high kilomet larg least low mani maximum meter mile more most mostfew name no not number small squarkilomet total
% 43 used predicates: affirm/2 area/2 argmax/4 argmin/4 capital/1 city/1 compareMore/3 count/2 country/1 density/2 elevation/2 every/3 hasInhabitant/2 kilometer/2 lake/1 len/2 loc/1 loc/2 major/1 max/2 mean/2 meter/2 mile/2 mountain/1 nameObj/2 negArea/2 negDensity/2 negElevation/2 negLen/2 negSize/2 negate/2 next_to/2 not/3 person/1 place/1 population/2 river/1 size/2 squared_kilometer/2 state/1 sum/2 traverse/2 {[usa:countryid/2]}

The \geo\ dataset, originally created by \citet{zelle96geoquery},
contains 880 questions about US Geography and a database of facts encoded in Prolog.
The questions in \geo\ ask about general properties (e.g., area,
elevation, population) of geographical entities (e.g., cities, states, rivers,
mountains).
Across all the questions, there are 280 word types, and the length of
an utterance ranges from 4 to 19 words, with an average of 8.5 words.
The questions involve conjunctions, superlatives, negation, but no generalized quantification.
Each question is annotated with a logical form in Prolog, for example:

\begin{tabular}{rl}
Utterance: & \nl{What is the highest point in Florida?} \\
Logical form: & \wl{\footnotesize answer(A,highest(A,(place(A),loc(A,B),const(B,stateid(florida)))))}
\end{tabular}

Since our approach learns from answers, not logical forms,
we evaluated the annotated logical forms on the provided database to obtain the correct answers.

% Database
Recall that a world/database $w$ maps each predicate $p \in \sP$ to a set of tuples $w(p)$.
Some predicates contain the set of tuples explicitly (e.g., \wl{mountain});
others can be derived (e.g., \wl{higher} takes two entities $x$ and $y$
and returns true if $\wl{elevation}(x) > \wl{elevation}(y)$).
Other predicates are higher-order (e.g., \wl{sum}, \wl{highest}) in that they take other predicates as arguments.
We do not use the provided domain-specific higher-order predicates (e.g., \wl{highest}),
but rather provide domain-independent higher-order predicates (e.g., \wl{argmax})
and the ordinary domain-specific predicates (e.g., \wl{elevation}).
This provides more compositionality and therefore better generalization.
Similarly, we use \wl{more} and \wl{elevation} instead of \wl{higher}.
Altogether, $\sP$ contains 43 predicates plus
one predicate for each value (e.g., \wl{CA}).

\paragraph{Job Queries}

% ec2:78.exec (JOB, FINALTEST, lexMode=0)
% 15 used original lexical entries: -BRIDGE- :NN :NNS atleast doesnt dont great least less more name no not outsid year
% 23 used predicates: affirm/2 application/2 area/2 argmin/4 company/2 compareLess/3 compareMore/3 deg/1 desire/2 exp/1 job/1 language/1 language/2 loc/1 loc/2 nameObj/2 not/3 platform/2 recruiter/2 require/2 salary_greater_than/2 title/2 year/2

The \job\ dataset \citep{tang01ilp} contains 640 natural language queries about job postings.
Most of the questions ask for jobs matching various criteria:
job title, company, recruiter, location, salary, languages and platforms used,
areas of expertise, required/desired degrees, and required/desired years of experience.
Across all utterances, there are 388 word types, and the length of an utterance
ranges from 2 to 23 words, with an average of 9.8 words.
% Overall: 388 word types, sentence length = 2/ << 9.809~3.182 >> /23 (640)
The utterances are mostly based on conjunctions of criteria, with a sprinkling of negation and disjunction.
Here is an example:

\scalebox{0.98}{
\begin{tabular}{rl}
Utterance: & \nl{Are there any jobs using Java that are not with IBM?} \\
Logical form: & \wl{\footnotesize answer(A,(job(A),language(A,'java'),$\neg$company(A,'IBM')))}
\end{tabular}
}
%Logical form: & \wl{\footnotesize answer(A,(job(A),language(A,L),const(L,java),$\neg$(company(A,C),const(C,'IBM'))))}

The \job\ dataset comes with a database, which we can use as the world $w$.
However, when the logical forms are evaluated on this database, close to half of
the answers are empty (no jobs match the requested criteria).  Therefore, there is
a large discrepancy between obtaining the correct logical form (which has been the focus
of most work on semantic parsing) and obtaining the correct answer (our focus).
%given that guessing the empty set yields around 50\% accuracy.
%While answering a query correctly seems easier, learning the correct parameters is harder,
%since there are many spurious logical forms that might produce the same answer.

% Create random
To bring these two into better alignment, we generated a random database as follows:
We created $m=100$ jobs.  For each job $j$,
we go through each predicate $p$ (e.g., \wl{company}) that takes two arguments, a job and a target value.
For each of the possible target values $v$,
we add $(j,v)$ to $w(p)$ independently with probability $\alpha=0.8$.
For example, for $p=\wl{company}$, $j=\wl{job37}$, we might add $(\wl{job37}, \wl{IBM})$ to $w(\wl{company})$.
The result is a database with a total of 23 predicates (which includes the domain-independent ones)
in addition to the value predicates (e.g., \wl{IBM}).

% Justify random
The goal of using randomness is to ensure that two different logical forms
will most likely yield different answers.  For example, consider two logical forms:
\begin{align}
z_1 &= \lambda j . \wl{job}(j) \wedge \wl{company}(j, \wl{IBM}), \\
z_2 &= \lambda j . \wl{job}(j) \wedge \wl{language}(j, \wl{Java}).
\end{align}
Under the random construction,
the denotation of $z_1$ is $S_1$, a random subset of the jobs,
where each job is included in $S_1$ independently with probability $\alpha$,
and the denotation of $z_2$ is $S_2$, which has the same distribution as $S_1$ but importantly is independent of $S_1$.
Therefore, the probability that $S_1 = S_2$ is $[\alpha^2 + (1-\alpha)^2]^m$,
which is exponentially small in $m$.
%with $\alpha=0.8$ and $m=100$, this probability is approximately $1.8 \times 10^{-17}$.
%We can do a similar analysis for logical forms that share conjuncts.  Each shared
%conjunct effectively reduces the size of $m$ by a factor of $\alpha$.
%Therefore, we have set $\alpha$ to be relatively high, so that $m$ does not have to be too large.
This construction yields a world that is not entirely ``realistic''
(a job might have multiple employers), but it ensures that if we get the
correct answer, we probably also obtain the correct logical form.
%serve their principal purpose: making sure different logical forms yield different answers
%for the purposes of evaluation.

% Different complexity
%These two datasets, \geo\ and \job, are very different in the following sense:
%The more complex utterances in \geo\ are built out of nested relative clauses
%(e.g., \nl{states that border states that border states$\dots$}),
%which leads to DCS trees which are narrow and deep.
%The complex utterances in \job\ are built out of a conjunction of multiple constraints (\nl{jobs in New York with IBM using Java$\dots$}),
%which leads to DCS trees which are broad and shallow.

%%%%%%%%%%%%%%%%%%%%%%%%%%%%%%
\SecThree{settings}{Settings}

There are a number of settings which control the tradeoffs between computation,
expressiveness, and generalization power of our model, shown below.   For now, we will use generic settings
chosen rather crudely; \refsec{effectSettings} will explore the effect of changing these settings.

\begin{description}
\item [Lexical Triggers]

The lexical triggers $L$ (\refsec{lexicalTriggers})
define the set of candidate DCS trees for each utterance.  There is a tradeoff
between expressiveness and computational complexity:~The more triggers we have,
the more DCS trees we can consider for a given utterance,
but then either the candidate sets become too large or beam search starts dropping the good DCS trees.
Choosing lexical triggers is important and requires additional supervision
(\refsec{lexicalTriggerExperiments}).

\item [Features]
Our probabilistic semantic parsing model is defined in terms of feature templates (\refsec{features}).
Richer features increase expressiveness but also might lead to overfitting.
By default, we include all the feature templates.
%In \refsec{effectFeatures}, we study the effect of including various subsets of these feature templates,

\item [Number of training examples ($n$)]
An important property of any learning algorithm is its sample
complexity---how many training examples are required to obtain a certain level of
accuracy?  By default, all training examples are used.

\item [Number of training iterations ($T$)]
Our learning algorithm (\reffig{learningAlgorithm}) alternates between updating
candidate sets and updating parameters for $T$ iterations.
We use $T=5$ as the default value.
%More iterations can improve accuracy but increases the risk of overfitting.

\item [Beam size ($K$)]
The computation of the candidate sets in \reffig{learningAlgorithm} is based on beam
search where each intermediate state keeps at most $K$ DCS trees.
The default value is $K=100$.

\item [Optimization algorithm]
To optimize an the objective function $\sO(\theta,C)$ our
default is to use the standard L-BFGS algorithm \citep{nocedal80lbfgs} 
with a backtracking line search for choosing the step size.

\item [Regularization ($\lambda$)]
The regularization parameter $\lambda > 0$ in the objective function
$\sO(\theta,C)$ is another knob for controlling the tradeoff between fitting
and overfitting.
The default is $\lambda = 0.01$.
\end{description}

%%%%%%%%%%%%%%%%%%%%%%%%%%%%%%
\SecThree{lexicalTriggerExperiments}{Lexical Triggers}

The lexical trigger set $L$ (\refsec{lexicalTriggers})
is a set of entries $(\bs,p)$, where $\bs$ is a sequence of words and $p$ is a predicate.
We run experiments on two sets of lexical triggers:
{\em base triggers} $\Lb$ and {\em augmented triggers} $\Lbp$.

\paragraph{Base Triggers}
The base trigger set $\Lb$ includes three types of entries:
\begin{itemize}
\item Domain-independent triggers:
For each domain-independent predicate (e.g., \wl{argmax}),
we manually specify a few words associated with that predicate (e.g., \nl{most}).
The full list is shown at the top of \reffig{lexicalTriggers}.

\item Values:
For each value $x$ that appears in the world (specifically, $x \in v_j \in
w(p)$ for some tuple $v$, index $j$, and predicate $p$), $\Lb$ contains an
entry $(x,x)$ (e.g., $(\nl{Boston},\nl{Boston}\cto\wl{city})$).
Note that this rule implicitly specifies an infinite number of triggers.
%Note that there are an infinite number of these triggers,
%but they have a very compact logical representation based on string matching.

Regarding predicate names, we do {\em not} add entries such as $(\nl{city},\wl{city})$,
because we want our system to be language-independent.
In Turkish, for instance, we would not have the luxury of lexicographical cues that associate \wl{city} with \nl{şehir}.
So we should think of the predicates as just symbols \wl{predicate1}, \wl{predicate2}, etc.
On the other hand, values in the database are generally proper nouns (e.g., city names)
for which there are generally strong cross-linguistic lexicographic similarities.

\item Part-of-speech (POS) triggers:\footnote{
To perform POS tagging, we used the Berkeley Parser \citep{petrov06latent},
trained on the WSJ Treebank \citep{marcus93treebank} and the Question Treebank
\citep{judge06qtb}---thanks to Slav Petrov for providing the trained grammar.}
For each domain-specific predicate $p$, we specify a set of part-of-speech tags $T$.
Implicitly, $\Lb$ contains all pairs $(x,p)$ where the word $x$ has a POS tag $t \in T$.
For example, for \wl{city}, we would specify \ts{nn} and \ts{nns}, which means
that any word which is a singular or plural common noun triggers the
predicate \wl{city}.  Note that \nl{city} triggers \wl{city} as desired,
but \nl{state} also triggers \wl{city}.
%We rely on features and learning favor the former over the latter.

The POS triggers for \geo\ and \job\ domains are shown in the left side of
\reffig{lexicalTriggers}.  Note that that some predicates such as \wl{traverse}
and \wl{loc} are not associated with any POS tags.  Predicates corresponding to
verbs and prepositions are not included as overt lexical triggers, but rather
included as trace predicates $L(\epsilon)$.
In constructing the logical forms, nouns and adjectives serve as anchor points.
Trace predicates can be inserted in between these anchors.  This strategy is more flexible
than requiring each predicate to spring from some word.

\end{itemize}

\paragraph{Augmented Triggers}

We now define the augmented trigger set $\Lbp$, which contains more
domain-specific information than $\Lb$.  Specifically, for each domain-specific
predicate (e.g., \wl{city}), we manually specify a single prototype word (e.g.,
\nl{city}) associated with that predicate.
%For non-prototype words, we backoff to the base triggers $\Lb$.
Under $\Lbp$, \nl{city} would trigger
only \wl{city} because \nl{city} is a prototype word, but \nl{town} would trigger
all the \ts{nn} predicates (\wl{city}, \wl{state}, \wl{country}, etc.)
because it is not a prototype word.
%because we would fall back on the POS trigger.

Prototype triggers require only a modest amount of domain-specific supervision
(see the right side of \reffig{lexicalTriggers} for the entire list for \geo\ and \job).
%In particular, they do not encode how the predicate should combine with its context,
%and thus are much leaner than CCG lexical entries.
In fact, as we'll see in \refsec{compareResults},
prototype triggers are not absolutely required to obtain good accuracies,
but they give an extra boost and also improve computational efficiency by reducing the set of candidate DCS trees.

\FigTop{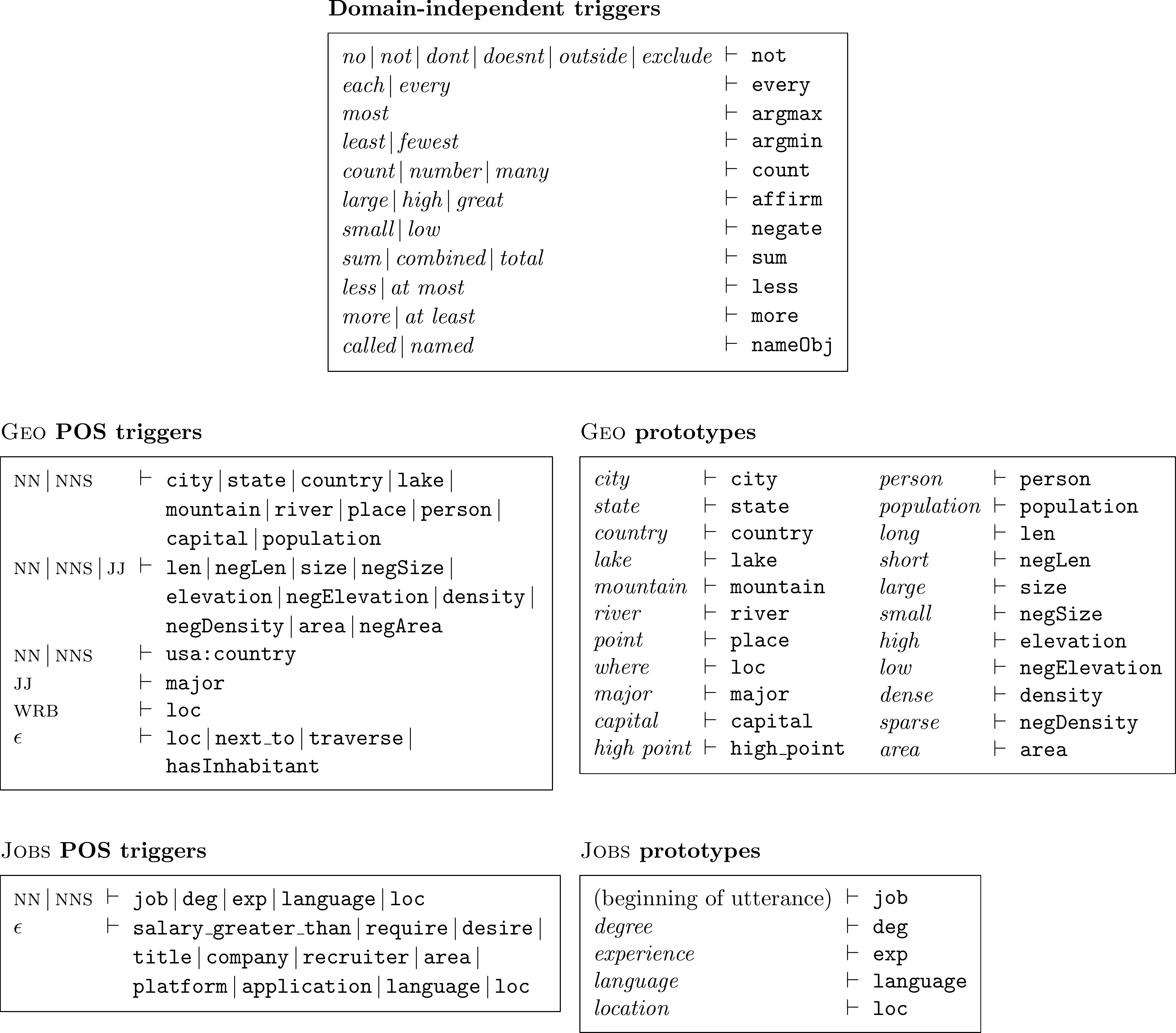}{0.3}{lexicalTriggers}{Lexical triggers
used in our experiments.
}

% Details
Finally, we use a small set of rules that expand morphology (e.g., \nl{largest}
is mapped to \nl{most large}).
To determine triggering, we stem all words using the Porter stemmer \citet{porter80stem},
so that \nl{mountains} triggers the same predicates as \nl{mountain}.

%%%%%%%%%%%%%%%%%%%%%%%%%%%%%%%%%%%%%%%%%%%%%%%%%%%%%%%%%%%%
\SecTwo{compareResults}{Comparison with Other Systems}

We now compare our approach with existing methods (\refsec{compareResults}).
We used the same training-test splits as \citet{zettlemoyer05ccg}
(600 training and 280 test examples for \geo,
500 training and 140 test examples for \job).
For development, we created five random splits of the training data.
For each split, we put 70\% of the examples into a {\em development training set} and the
remaining 30\% into a {\em development test set}.  The actual test set was only
used for obtaining final numbers.

%%%%%%%%%%%%%%%%%%%%%%%%%%%%%%
\SecThree{experimentOne}{Systems that Learn from Question-Answer Pairs}

\newcommand\rcitep\citep

We first compare our system (henceforth, LJK11) with \citet{clarke10world} (henceforth, CGCR10), which is most similar to our work
in that it also learns from question-answer pairs without using annotated logical forms.  CGCR10 works 
with the FunQL language and casts semantic parsing as integer linear programming (ILP).
In each iteration,
the learning algorithm solves the ILP to predict the logical form for each training example.
The examples with correct predictions are fed to a structural SVM and the model parameters are updated.

% Difference in algorithm
Though similar in spirit, there are some important differences between CGCR10 and our approach.
They use ILP instead of beam search and structural SVM instead of log-linear models, but the
main difference is which examples are used for learning.  Our approach learns
on any feasible example (\refsec{objective}), one where the candidate set
contains a logical form that evaluates to the correct answer.  CGCR10 uses a much more stringent
criterion: the highest scoring logical form must evaluate to the correct answer.
Therefore, for their algorithm to progress, the model already must be
non-trivially good before learning even starts.
This is reflected in the amount of prior knowledge and
initialization that CGCR10 employs before learning starts:
WordNet features, and syntactic parse trees, and
a set of lexical triggers with 1.42 words per non-value predicate.
Our system with base triggers requires only
simple indicator features, POS tags, and 0.5 words per non-value predicate.

% Results
CGCR10 created a version of \geo\ which contains 250 training and 250 test examples.
\reftab{resultsSmall} compares the empirical results on this split.
We see that our system (LJK11) with base triggers significantly
outperforms CGCR10 (84\% over 73.2\%), and it even outperforms the version of CGCR10 that is
trained using logical forms (84.0\% over 80.4\%).
If we use augmented triggers, we widen the gap by another 3.6\%.\footnote{Note that
the numbers for LJK11 differ from those presented in \citet{liang11dcs}, which
reports results based on 10 different splits rather than the setup used by CGCR10.}

\begin{table}
\begin{center}
\begin{tabular}{ll|c}
System                     &                              & Accuracy   \\ \hline
CGCR10 w/answers           & \rcitep{clarke10world}       & 73.2       \\
CGCR10 w/logical forms     & \rcitep{clarke10world}       & 80.4       \\        
LJK11 w/base triggers      & \rcitep{liang11dcs}          & 84.0       \\ % lexMode=0 (ec2:15.exec)
LJK11 w/augmented triggers & \rcitep{liang11dcs}          & {\bf 87.6} \\ % lexMode=2 (ec2:17.exec)
\end{tabular}     
\caption{\label{tab:resultsSmall}
Results on \geo\ with 250 training and 250 test examples.
Our system (LJK11 with base triggers and no logical forms) obtains higher test accuracy than
CGCR10, even when CGCR10 is trained using logical forms.
}
\end{center}
\end{table}

%%%%%%%%%%%%%%%%%%%%%%%%%%%%%%
\SecThree{experimentTwo}{State-of-the-Art Systems}

We now compare our system (LJK11) with state-of-the-art systems,
which all require annotated logical forms (except \ts{Precise}).
Here is a brief overview of the systems:
\begin{itemize}

\item \ts{Cocktail} \citep{tang01ilp} uses inductive logic programming to learn rules for driving
the decisions of a shift-reduce semantic parser.  It assumes that a lexicon
(mapping from words to predicates) is provided.
% Trained on 250?

\item \ts{Precise} \citep{popescu03precise}
does not use learning, but instead relies on matching
words to strings in the database using various heuristics based on WordNet and the Charniak parser.
Like our work, it also uses database type constraints to rule out spurious logical forms.
One of the unique features of \ts{Precise} is that it has
100\% precision---it refuses to parse an utterance which it deems {\em semantically intractable}.

\item \ts{Scissor} \citep{ge05scissor}
learns a generative probabilistic model that extends the Collins models \citep{collins99thesis}
with semantic labels, so that syntactic and semantic parsing can be done jointly.

% Main results on 250, also test on 880, compare with Precise

\item \ts{Silt} \citep{kate05funql} 
learns a set of transformation rules for mapping utterances to logical forms.

\item \ts{Krisp} \citep{kate06krisp}
uses SVMs with string kernels to drive the local decisions of a chart-based semantic parser.

\item \ts{Wasp} \citep{wong06mt}
uses log-linear synchronous grammars to transform utterances into logical forms,
starting with word alignments obtained from the IBM models.

\item $\lambda$-\ts{Wasp} \citep{wong07synchronous}
extends \ts{Wasp} to work with logical forms that contain bound variables (lambda abstraction).

\item LNLZ08 \citep{lu08generative} learns a generative model over {\em hybrid trees},
which are logical forms augmented with natural language words.
IBM model 1 is used to initialize the parameters,
and a discriminative reranking step works on top of the generative model.

\item ZC05 \citep{zettlemoyer05ccg} learns a discriminative log-linear model over CCG derivations.
Starting with a manually-constructed domain-independent lexicon,
the training procedure grows the lexicon by
adding lexical entries derived from associating
parts of an utterance with parts of the annotated logical form.

\item ZC07 \citep{zettlemoyer07relaxed} extends ZC05 with extra
(disharmonic) combinators to increase the expressive power of the model.

\item KZGS10 \citep{kwiatkowski10ccg} uses a restricted higher-order unification procedure,
which iteratively breaks up a logical form into smaller pieces.
This approach gradually adds lexical entries of increasing generality,
thus obviating the need for the manually-specified templates used by ZC05 and ZC07
for growing the lexicon.  IBM model 1 is used to initialize the parameters.

\item KZGS11 \citep{kwiatkowski11lex} extends KZGS10 by factoring lexical
entries into a template plus a sequence of predicates which fill the slots of
the template.  This factorization improves generalization.

\end{itemize}

With the exception of \ts{Precise}, all other systems require annotated logical forms,
whereas our system learns from annotated answers.  On the other hand,
many of the later systems require essentially no manually-crafted lexicon
and instead rely on unsupervised word alignment (e.g.,
\citet{wong06mt,wong07synchronous,kwiatkowski10ccg,kwiatkowski11lex}) and/or
lexicon learning (e.g., \citet{zettlemoyer05ccg,zettlemoyer07relaxed,kwiatkowski10ccg,kwiatkowski11lex}).
We cannot use these automatic techniques because they require annotated logical forms.
Our system instead relies on lexical triggers, which does require some manual effort.
These lexical triggers play a crucial role in the initial stages of learning,
because they constrain the set of candidate DCS trees; otherwise we would face
a hopelessly intractable search problem.

\reftab{resultsGeoBig} shows the results for \geo.
Semantic parsers are typically evaluated on the accuracy of the logical forms:
precision (the accuracy on utterances which are successfully parsed) and recall
(the accuracy on all utterances).  We only focus on recall (a lower bound on precision)
and simply use the word {\em accuracy} to refer to recall.\footnote{
Our system produces a logical form for every utterance, and thus our precision is the same as our recall.}
Our system is evaluated only on answer accuracy because our model
marginalizes out the latent logical form.
All other systems are evaluated on the accuracy of logical forms.
To calibrate, we also evaluated KZGS10 on answer accuracy
and found that it was quite similar to its logical form accuracy (88.9\% versus
88.2\%).\footnote{The 88.2\% corresponds to 87.9\% in
\citet{kwiatkowski10ccg}.  The difference is due to using a slightly newer version of the code.}
This does not imply that our system would necessarily have a high logical form accuracy
because multiple logical forms can produce the same answer,
and our system does not receive a training signal to tease them apart.
Even with only base triggers, our system (LJK11) outperforms all but two of the systems,
falling short of KZGS10 by only one point (87.9\% versus 88.9\%).\footnote{The 87.9\% and 91.4\%
correspond to 88.6\% and 91.1\% in \citet{liang11dcs}.  These differences are due to minor differences in the code.}
With augmented triggers, our system takes the lead (91.4\% over 88.9\%).

\begin{table}
\begin{center}
\small
\begin{tabular}{ll|cc}
System                         &                               & LF         & Answer     \\ \hline
\ts{Cocktail}                  & \rcitep{tang01ilp}            & 79.4       & --         \\         
\ts{Precise}                   & \rcitep{popescu03precise}     & 77.5       & 77.5       \\       
\ts{Scissor}                   & \rcitep{ge05scissor}          & 72.3       & --         \\
\ts{Silt}                      & \rcitep{kate05funql}          & 54.1       & --         \\
\ts{Krisp}                     & \rcitep{kate06krisp}          & 71.7       & --         \\
\ts{Wasp}                      & \rcitep{wong06mt}             & 74.8       & --         \\
$\lambda$-\ts{Wasp}            & \rcitep{wong07synchronous}    & 86.6       & --         \\
LNLZ08                         & \rcitep{lu08generative}       & 81.8       & --         \\ \hline
ZC05                           & \rcitep{zettlemoyer05ccg}     & 79.3       & --         \\
ZC07                           & \rcitep{zettlemoyer07relaxed} & 86.1       & --         \\
KZGS10                         & \rcitep{kwiatkowski10ccg}     & 88.2       & 88.9       \\
KZGS11                         & \rcitep{kwiatkowski10ccg}     & {\bf 88.6} & --         \\
LJK11 w/base triggers          & \rcitep{liang11dcs}           & --         & 87.9       \\
LJK11 w/augmented triggers     & \rcitep{liang11dcs}           & --         & {\bf 91.4} \\
\end{tabular}
\caption{\label{tab:resultsGeoBig}
Results on \geo.
Logical form accuracy (LF) and answer accuracy (Answer)
of the various systems.
The first group of systems are evaluated using 10-fold cross-validation on all 880 examples;
the second are evaluated on the $680+200$ split of \citet{zettlemoyer05ccg}.
Our system (LJK11) with base triggers obtains comparable accuracy to past work,
while with augmented triggers, our system obtains the highest overall accuracy.
}
\end{center}
\end{table}

\reftab{resultsJobBig} shows the results for \job.
%Here, there is less previous work because the dataset is less interesting than \geo.
The two learning-based systems (\ts{Cocktail} and ZC05) are actually outperformed
by \ts{Precise}, which is able to use strong database type constraints.
By exploiting this information and doing learning, we obtain the best results.

\begin{table}
\begin{center}
\small
\begin{tabular}{ll|cc}
System                         &                               & LF         & Answer     \\ \hline
\ts{Cocktail}                  & \rcitep{tang01ilp}            & 79.4       & --         \\
\ts{Precise}                   & \rcitep{popescu03precise}     & {\bf 88.0} & 88.0       \\ \hline     
ZC05                           & \rcitep{zettlemoyer05ccg}     & 79.3       & --         \\
LJK11 w/base triggers          & \rcitep{liang11dcs}           & --         & 90.7       \\
LJK11 w/augmented triggers     & \rcitep{liang11dcs}           & --         & {\bf 95.0} \\
\end{tabular}
\caption{\label{tab:resultsJobBig}
Results on \job.   Both \ts{Precise} and our system use database type constraints,
which results in a decisive advantage over the other systems.
In addition, LJK11 incorporates learning and therefore obtains the highest accuracies.
}
\end{center}
\end{table}

%%%%%%%%%%%%%%%%%%%%%%%%%%%%%%%%%%%%%%%%%%%%%%%%%%%%%%%%%%%%
\SecTwo{empiricalAnalysis}{Empirical Properties}

In this section, we try to gain intuition into properties of our approach.
All experiments in this section are performed on random development splits.
Throughout this section, ``accuracy'' means development test accuracy.

%%%%%%%%%%%%%%%%%%%%%%%%%%%%%%
\SecThree{errorAnalysis}{Error Analysis}

To understand the type of errors our system makes, we examined one of the development runs,
which had 34 errors on the test set.
We classified these errors into the following categories
(the number of errors in each category is shown in parentheses):

\begin{itemize}
\item Incorrect POS tags (8): \geo\ is out-of-domain for our POS tagger,
so the tagger makes some basic errors which adversely affect the predicates that can be lexically triggered.
%and thus leave out desired candidate DCS trees.
For example, the question \nl{What states border states \dots} is tagged
as \ts{wp} \ts{vbz} \ts{nn} \ts{nns} \dots,
which means that the first \nl{states} cannot trigger \wl{state}.
In another example, \nl{major river} is tagged as \ts{nnp} \ts{nnp}, so these cannot trigger the appropriate predicates either,
and thus the desired DCS tree cannot even be constructed.
%Training a tagger with a small amount of in-domain data would suffice to correct these errors.

\item Non-projectivity (3): The candidate DCS trees are defined
by a projective construction mechanism (\refsec{construction}) that prohibits edges in the DCS tree from crossing.
This means we cannot handle utterances such as \nl{largest city by area},
since the desired DCS tree would
have \wl{city} dominating \wl{area} dominating \wl{argmax}. 
To construct this DCS tree, we could allow local reordering of the words.
%but this risks over-expanding the set of candidate DCS trees.

\item Unseen words (2): We never saw \nl{at least} or \nl{sea level} at training time.
The former has the correct lexical trigger, but not a sufficiently large feature weight (0)
to encourage its use.  For the latter, the problem is more structural:
We have no lexical triggers for $0\cto\wl{length}$,
and only adding more lexical triggers can solve this problem.

\item Wrong lexical triggers (7): Sometimes the error is localized to a single lexical trigger.
%for example, mapping \nl{area} to \wl{size} for cities.
%This happens because \wl{size} means \wl{population} for cities but \wl{area} for states,
%and the model over-generalizes.
For example, the model incorrectly thinks \nl{Mississippi} is the state rather than the river,
and that \nl{Rochester} is the city in New York rather than the name,
even though there are contextual cues to disambiguate in these cases.

\item Extra words (5): Sometimes, words trigger predicates that should be ignored.
For example, for \nl{population density}, the first word triggers
\wl{population}, which is used rather than \wl{density}.

\item Over-smoothing of DCS tree (9): The first half of our features (\reffig{features})
are defined on the DCS tree alone; these produce a form of smoothing that
encourages DCS trees to look alike regardless of the words.
We found several instances where this essential tool for generalization went too far.
For example, in \nl{state of Nevada}, the trace predicate \wl{border} is inserted between the two nouns,
because it creates a structure more similar to that of the common question \nl{what states border Nevada?}

\end{itemize}

%%%%%%%%%%%%%%%%%%%%%%%%%%%%%%
\SecThree{eyeCandy}{Visualization of Features}

Having analyzed the behavior of our system for individual utterances, let us
move from the token level to the type level and analyze the learned parameters
of our model.  We do not look at raw feature weights,
because there are complex interactions between them not represented by examining individual weights.
Instead, we look at expected feature counts, which we think are more interpretable.

\FigTop{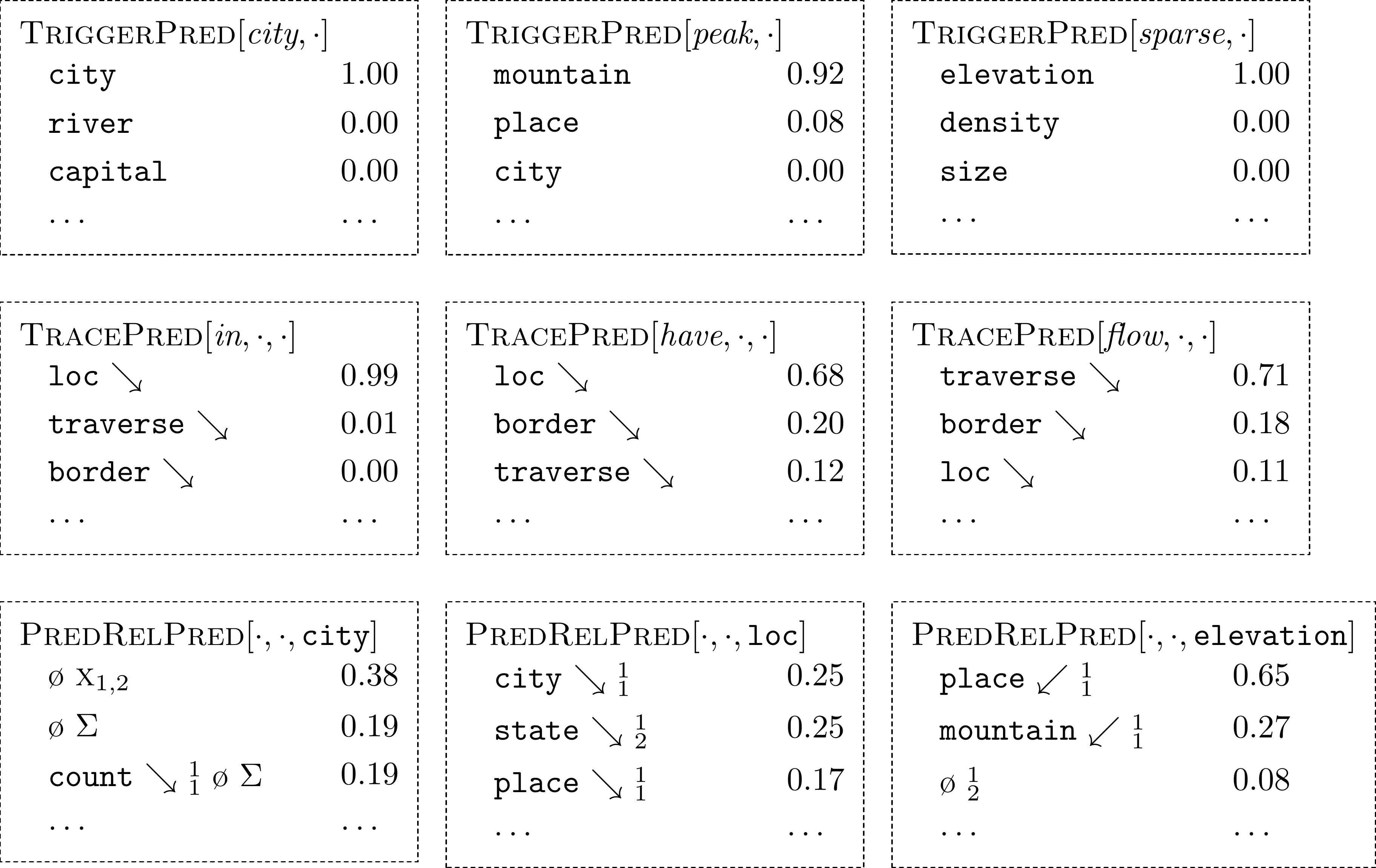}{0.35}{featureDistrib}{
Learned feature distributions.  In a feature group (e.g., $\WordTriggerPred[\nl{city},\cdot]$),
each feature is associated with the marginal probability that the feature fires
according to \refeqn{featureDistrib}.
Note that we have successfully learned that \nl{city} means \wl{city},
but incorrectly learned that \nl{sparse} means \wl{elevation} (due to the confounding fact
that Alaska is the most sparse state and has the highest elevation).
}

Consider a group of ``competing'' features $J$, for 
example $J = \{ \WordTriggerPred[\nl{city},p] : p \in \sP \}$.
We define a distribution $q(\cdot)$ over $J$ as follows:
\begin{align}
q(j) & = \frac{N_j}{\sum_{j' \in J} N_{j'}}, \text{ where} \label{eqn:featureDistrib} \\
N_j &= \sum_{(\bx,y) \in \sD} \E_{p(z \mid \bx, \tilde\sZ_{L,\theta}, \theta)}[\phi(\bx,z)]. \nonumber
\end{align}
Think of $q(j)$ as a marginal distribution (since all our features are positive) which represents
the relative frequencies with which the features $j \in J$ fire
with respect to our training dataset $\sD$ and trained model $p(z \mid \bx, \tilde\sZ_{L,\theta}, \theta)$.
To appreciate the difference between what this distribution 
and raw feature weights capture, suppose we had two features, $j_1$ and $j_2$,
which are identical ($\phi(\bx,z)_{j_1} \equiv \phi(\bx,z)_{j_2}$).
The weights would be split across the two features, but the features would have
the same marginal distribution ($q(j_1) = q(j_2)$).
\reffig{featureDistrib} shows some of the feature distributions learned.
%for three of our feature templates. 
%The lexicalized feature templates,
%$\WordTriggerPred$ and $\WordTracePred$, tend to have sharper distributions
%than the ones defined only on the DCS tree

%%%%%%%%%%%%%%%%%%%%%%%%%%%%%%
\SecThree{bootstrapping}{Learning, Search, Bootstrapping}

\FigTop{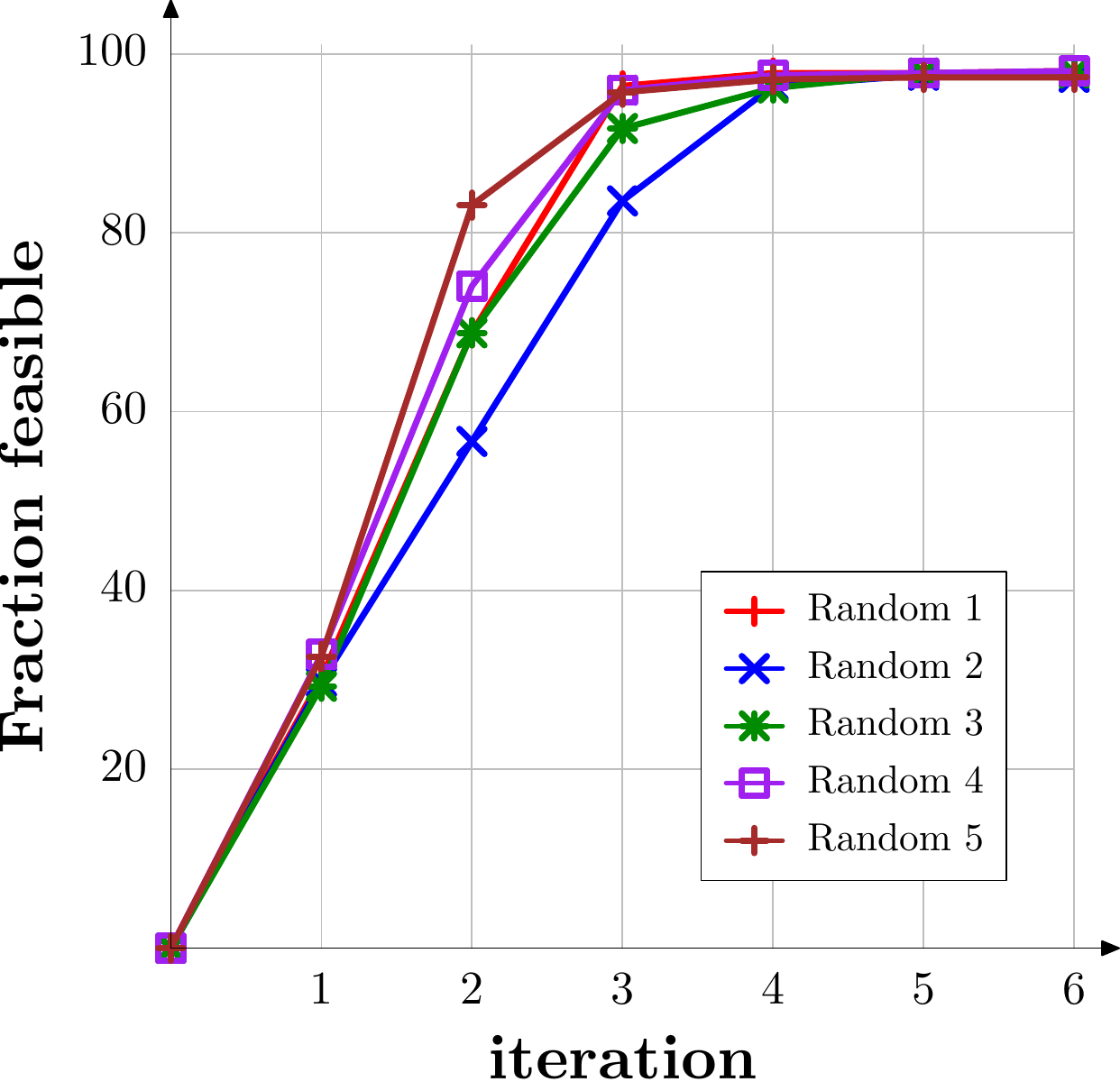}{0.5}{oracleAccuracy}{
The fraction of feasible training examples increases steadily as the parameters, and thus, the beam search,
improves.  Each curve corresponds to a run on a different development split.
%The variation due to the randomness across the five development splits is shown,
%but the general trend is unchanged.
}

Recall from \refsec{objective} that a training example is feasible (with
respect to our beam search) if the resulting candidate set contains a DCS tree
with the correct answer.  Infeasible examples are skipped, but an example may
become feasible in a later iteration.  A natural question is how many training examples
are feasible in each iteration.  \reffig{oracleAccuracy} shows the
answer:~Initially, only around 30\% of the training examples are feasible;
this is not surprising given that all the parameters are zero, so our beam search is essentially unguided.
However, training on just these examples improves the parameters, and over the next few iterations,
the number of feasible examples steadily increases to around 97\%.

% Learning + search
In our algorithm, learning and search are deeply intertwined.
Search is of course needed to learn, but learning also improves search.
The general approach is similar in spirit to Searn \citep{daume09searn},
although we do not have any formal guarantees at this point.
%which works sequentially rather than along a parse chart.

Our algorithm also has a bootstrapping flavor.
The ``easy'' examples are processed first, where easy is
defined by the ability of beam search to generate the
correct answer.  This bootstrapping occurs quite naturally:~Unlike most bootstrapping algorithms,
we do not have to set a confidence threshold for accepting new training examples,
something that can be quite tricky to do.
Instead, our threshold falls out of the discrete nature of the beam search.

%%%%%%%%%%%%%%%%%%%%%%%%%%%%%%
\SecThree{effectSettings}{Effect of Various Settings}
%\SecThree{effectFeatures}{Effect of Features}

So far, we have used our approach with default settings (\refsec{settings}).
How sensitive is the approach to these choices?
\reftab{effectFeatures} shows the impact of the feature templates.
\reffig{effects} shows the effect of the number of training examples,
number of training iterations, beam size, and regularization parameter.
The overall conclusion is that there are no big surprises:~Our default settings
could be improved on slightly, but these differences
are often smaller than the variation across different development splits.

\begin{table}[ht]
\begin{center}
\begin{tabular}{l|r}
Features & Accuracy \\
\hline
$\Pred$ & $13.4 \pm 1.6$ \\
\hline
$\Pred+\PredRel$ & $18.4 \pm 3.5$ \\
$\Pred+\PredRel+\PredPred$ & $23.1 \pm 5.0$ \\
\hline
$\Pred+\WordTriggerPred$ & $61.3 \pm 1.1$ \\
$\Pred+\WordTriggerPred+\WordTraceAll$ & $76.4 \pm 2.3$ \\
\hline
$\Pred+\PredRel+\PredPred+\WordTriggerPred+\WordTraceAll$ & $84.7 \pm 3.5$ \\
\end{tabular}

\end{center}
\caption{\label{tab:effectFeatures}
There are two classes of feature templates: lexical features (\WordTriggerPred,\WordTraceAll)
and non-lexical features (\PredRel,\PredPred).
The lexical features are relatively much more important for obtaining
good accuracy (76.4\% versus 23.1\%),
but adding the non-lexical features makes a significant contribution as well (84.7\% versus 76.4\%).
}
\end{table}

\FigTop{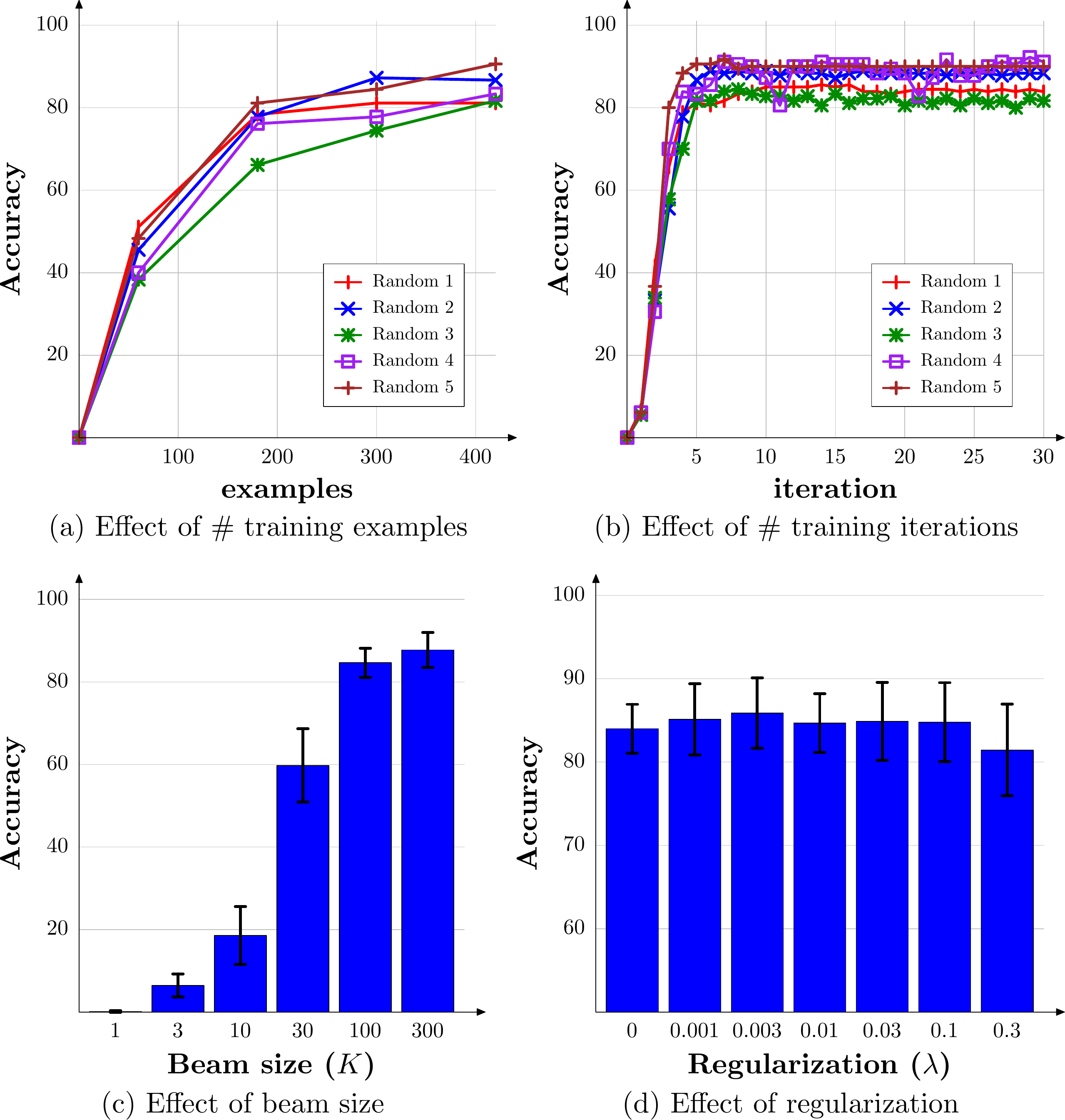}{0.5}{effects}{(a) The learning
curve shows test accuracy as the number of training examples increases;
about 300 examples suffices to get around 80\% accuracy.
(b) Although our algorithm is not guaranteed to converge,
the test accuracy is fairly stable (with one exception)
with more training iterations---hardly any overfitting occurs.
(c) As the beam size increases, the accuracy increases monotonically,
although the computational burden also increases.
There is a small gain from our default setting of $K=100$ to the more expensive $K=300$.
(d) The accuracy is relatively insensitive to the choice of the regularization parameter
for a wide range of values.  In fact, no regularization is also acceptable. 
This is probably because the features are simple, and the lexical triggers and
beam search already provide some helpful biases.
}

\FigTop{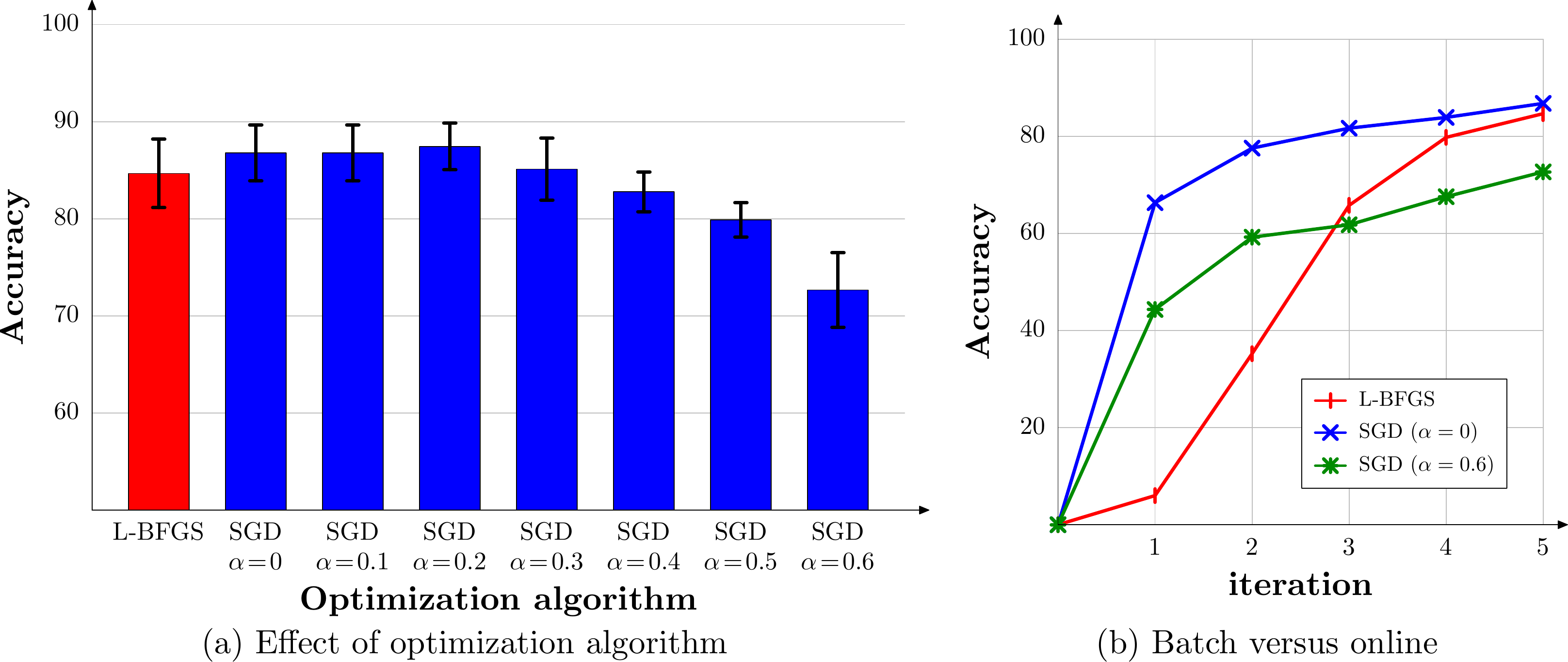}{0.4}{effectsOpt}{
(a) Given the same number of iterations,
compared to default batch algorithm (L-BFGS),
the online algorithm (stochastic gradient descent) is slightly better
for aggressive step sizes (small $\alpha$) and worse for conservative step sizes (large $\alpha$).
(b) The online algorithm (with an appropriate choice of $\alpha$)
obtains a reasonable accuracy much faster than L-BFGS.
}

% Optimization
We now consider the choice of optimization algorithm to update the parameters given candidate sets (see \reffig{learningAlgorithm}).
Thus far, we have been using L-BFGS \citep{nocedal80lbfgs},
which is a batch algorithm:~Each iteration,
we construct the candidate sets $C^{(t)}(\bx)$ for all the training examples
before solving the optimization problem $\argmax_\theta \sO(\theta, C^{(t)})$.
We now consider an online algorithm, stochastic gradient descent (SGD) \citep{munro51stochastic},
which updates the parameters after computing the candidate set for each example.
In particular, we iteratively scan through the training examples in a random order.
%and at each step $t$, 
For each example $(\bx,y)$, we compute the candidate set using beam search.
We then update the parameters in the direction of the gradient of the marginal log-likelihood for that example
(see \refeqn{gradient}) with step size $t^{-\alpha}$:
\begin{align}
\theta^{(t+1)} \leftarrow \theta^{(t)} + t^{-\alpha} \p{ \frac{\partial \log p(y \mid \bx; \tilde\sZ_{L,\theta^{(t)}}, \theta)}{\partial \theta}\Big|_{\theta=\theta^{(t)}}}.
\end{align}
The trickiest aspect of using SGD is selecting the correct step size:
a small $\alpha$ leads to quick progress but also instability;
a large $\alpha$ leads to the opposite.
We let L-BFGS and SGD both take the same number of iterations (passes over the training set).
\reffig{effectsOpt} shows that a very small value of $\alpha$ (less than $0.2$) is best 
for our task, even though only values between $0.5$ and $1$ guarantee convergence.
Our setting is slightly different since we are interleaving the SGD updates with beam search,
which might also lead to unpredictable consequences.
Furthermore, the non-convexity of the objective function exacerbates the
unpredictability \citep{liang09online}.
Nonetheless, with a proper $\alpha$, SGD converges much faster than L-BFGS
and even to a slightly better solution.
%Non-convex objectives share some of the same advantages of fast convergence,
%Online algorithms have been highly successful for convex objectives in NLP.

\SecOne{discussion}{Discussion}

%The ramifications of our empirical success are two-fold.
%The first is that in terms of information,
%answers only somewhat weaker than logical forms;
%of course, computationally, they are harder to learn from.

The work we have presented in this \paper\ addresses three important themes.
The first theme is {\em semantic representation} (\refsec{linguisticFrameworks}):
How do we parametrize the mapping from utterances to their meanings?
The second theme is {\em program induction} (\refsec{programInduction}):
How do we efficiently search through the space of logical structures
given a weak feedback signal?
Finally, the last theme is {\em grounded language} (\refsec{grounded}):
How do we use constraints from the world to guide learning of language
and conversely use language to interact with the world?

%%%%%%%%%%%%%%%%%%%%%%%%%%%%%%%%%%%%%%%%%%%%%%%%%%%%%%%%%%%%
\SecTwo{linguisticFrameworks}{Semantic Representation}

% Principle of compositionality
Since the late nineteenth century, philosophers and linguists have worked on
elucidating the relationship between an utterance and its meaning.  One of the
pillars of formal semantics is Frege's principle of
compositionality, that the meaning of an utterance is built by composing the
meaning of its parts.
What these parts are and how they are composed
is the main question.
The dominant paradigm, which stems from
the seminal work of Richard Montague in the early 1970s \citep{montague73ptq},
states that parts are lambda calculus expressions that correspond to syntactic
constituents, and composition is function application.
%or the more general combinators found in CCG \citep{steedman00ccg}.

% Statistical compositionality
Consider the compositionality principle from a statistical point of view,
where we construe compositionality as factorization.  Factorization,
the way a statistical model breaks into features, is necessary for
generalization: It enables us to learn from previously seen examples and
interpret new utterances.
Projecting back to Frege's original principle,
the parts are the features (\refsec{features}), and composition is the
DCS construction mechanism (\refsec{construction}) driven by parameters learned
from training examples.

% Acquisition
Taking the statistical view of compositionality, finding a good semantic
representation becomes designing a good statistical model.  But statistical modeling
must also deal with the additional issue of language acquisition or learning,
which presents complications:~In absorbing training examples, our learning algorithm must inevitably traverse
through intermediate models that are wrong or incomplete.  The algorithms
must therefore tolerate this degradation, and do so in a computationally
efficient way. 
For example, in the line of work on learning probabilistic CCGs
\citep{zettlemoyer05ccg,zettlemoyer07relaxed,kwiatkowski10ccg},
many candidate lexical entries must be entertained for each word even when
polysemy does not actually exist (\refsec{ccgComparison}).

To improve generalization, the lexicon can be further factorized \citep{kwiatkowski11lex},
but this is all done within the constraints of CCG.
DCS represents a departure from this tradition,
which replaces a heavily-lexicalized constituency-based formalism
with a lightly-lexicalized dependency-based formalism.
We can think of DCS as a shift in linguistic coordinate systems,
which makes certain factorizations or features more accessible.
For example, we can define features on paths between predicates in a DCS tree
which capture certain lexical patterns much more easily than in a lambda
calculus expression or a CCG derivation.
%This expands the types of which allows for new factorizations.
%with have favorable computational and statistical properties.
%Furthermore, from an engineering perspective, we want methods
%which can handle disfluent utterances robustly; this property also ignored in
%semantic theories and is delegated to another part of linguistics.

% NLF - dependency
DCS has a family resemblance to a semantic representation called 
\emph{natural logic form} \citep{alshawi11nlf}, which is also
motivated by the benefits of working with dependency-based logical 
forms.  The goals and the detailed structure of the two semantic 
formalisms are different, however.  \citet{alshawi11nlf} focuses on
parsing complex sentences in an open domain where a structured database or
world does not exist.  While they do equip their logical forms with a full
model-theoretic semantics, the logical forms are actually closer to dependency
trees: quantifier scope is left unspecified, and the predicates are simply the words.

% DRT
Perhaps not immediately apparent is the fact that DCS draws
an important idea from Discourse Representation Theory (DRT) 
\citep{kamp93drt}---not from the treatment of
anaphora and presupposition which it is known for, but something closer to its core.
This is the idea of having a logical form where all variables are existentially
quantified and constraints are combined via conjunction---a Discourse Representation Structure (DRS) in DRT,
or a basic DCS tree with only join relations.
Computationally, these logical structures conveniently encode CSPs.
Linguistically, it appears that existential quantifiers play an important
role and should be treated specially \citep{kamp93drt}.
DCS takes this core and focuses on semantic compositionality and computation,
while DRT focuses more on discourse and pragmatics.

% Programming language design
In addition to the statistical view of DCS as a semantic representation, it is
useful to think about DCS from the perspective of programming language design.
Two programming languages can be equally expressive, but what matters is how
simple it is to express a desired type of computation in a given language.
In some sense, we designed the DCS formal language to make it easy to represent computations
expressed by natural language.  An important part of DCS is the
mark-execute construct, a uniform framework for dealing with the divergence
between syntactic and semantic scope.  This construct allows us to build simple
DCS tree structures and still handle the complexities of phenomena such as
quantifier scope variation.  Compared to lambda calculus, think of DCS as a
higher-level programming language tailored to natural language, which results
in simpler programs (DCS trees).  Simpler programs are easier for us to work with
and easier for an algorithm to learn.

%%%%%%%%%%%%%%%%%%%%%%%%%%%%%%%%%%%%%%%%%%%%%%%%%%%%%%%%%%%%
\SecTwo{programInduction}{Program Induction}

Searching over the space of programs is challenging.
This is the central computational challenge of program induction, that of
inferring programs (logical forms) from their behavior (denotations).
This problem has been tackled by different communities in various forms:
program induction in AI, programming by demonstration in HCI, and
program synthesis in programming languages.  The core computational difficulty is that the
supervision signal---the behavior---is a complex function of the program
which cannot be easily inverted.
What program generated the output \nl{Arizona}, \nl{Nevada}, and \nl{Oregon}?

% Multi-task
Perhaps somewhat counterintuitively, program induction is easier if we
infer programs for not a single task but for multiple tasks.  The intuition
is that when the tasks are related, the solution to one task can help another
task, both computationally in navigating the program space and statistically in
choosing the appropriate program if there are multiple feasible possibilities
\citep{liang10programs}.
In our semantic parsing work, we want to infer a logical form for
each utterance (task).  Clearly the tasks are related
because they use the same vocabulary to talk about the same domain.

Natural language also makes program induction easier by providing side information (words) which can be used to guide
the search.  There have been several papers that induce programs in this setting:
\citet{eisenstein09read} induces conjunctive formulae from natural language instructions,
\citet{piantadosi08compositional} induces first-order logic formulae using CCG in a small domain assuming observed lexical semantics,
and \citet{clarke10world} induces logical forms in semantic parsing.
In the ideal case, the words would determine the program predicates,
and the utterance would determine the entire program compositionally.
But of course, this mapping is not given and must be learned.

%%%%%%%%%%%%%%%%%%%%%%%%%%%%%%%%%%%%%%%%%%%%%%%%%%%%%%%%%%%%
\SecTwo{grounded}{Grounded Language}

% Alignment
In recent years, there has been an increased interest in connecting language
with the world.\footnote{Here, {\em world} need not refer to the physical world,
but could be any virtual world.  The point is that the world has non-trivial structure
and exists extra-linguistically.}
One of the primary issues in grounded language is alignment---figuring out
what fragments of utterances refer to what aspects of the world.
In fact, semantic parsers trained on examples of utterances and annotated logical form (those discussed in \refsec{experimentTwo})
need to solve the task of aligning words to predicates.
Some can learn from utterances paired with a set of logical forms, one of which is correct
\citep{kate07krisper,chen08sportscast}.
\citet{liang09semantics} tackles the even more difficult alignment problem
of segmenting and aligning a discourse to a database of facts,
where many parts on either side are irrelevant.

% Response from the world
If we know how the world relates to language,
we can leverage structure in the world to guide the learning and interpretation of language.
We saw that type constraints from the
database/world reduces the set of candidate logical forms and lead to more accurate systems 
\citep{popescu03precise,liang11dcs}.
Even for syntactic parsing, information from the denotation of an utterance can
be helpful \citep{schuler03interpretation}.

% Less supervision
One of the exciting aspects about using the world for learning language is that it opens
the door to many new types of supervision.
We can obtain answers given a world, which are cheaper to obtain than logical forms \citep{clarke10world,liang11dcs}.
%Other researchers have pushed further in this direction:
\citet{goldwasser11confidence} learns a semantic parser based on bootstrapping
and estimating the confidence of its own predictions.
\citet{artzi11conversations} learns a semantic parser not from annotated
logical forms, but from user interactions with a dialog system.
\citet{branavan09reinforcement,branavan10high,branavan11win} use reinforcement
learning to follow natural language instructions
from a reward signal.
In general, supervision from the world is indirectly related to the learning task,
but it is often much more plentiful and natural to obtain.

% Language helps solve task
The benefits can also flow from language to the world.  For example,
previous work learned to interpret language to troubleshoot a Windows machine
\citep{branavan09reinforcement,branavan10high}, win a game of Civilization
\citep{branavan11win}, play a legal game of solitaire
\citep{eisenstein09read,goldwasser11instructions}, and navigate a map by
following directions \citep{vogel10navigate,chen11navigate}.  Even when the
objective in the world is defined independently of language (e.g., in
Civilization), language can provide a useful bias towards the non-linguistic
end goal.

%Our work pushes the grounded language agenda towards deeper representations of
%language---think grounded compositional semantics.

%%%%%%%%%%%%%%%%%%%%%%%%%%%%%%
\SecTwo{conclusion}{Conclusions}

The main conceptual contribution of this \paper\ is a new semantic formalism,
dependency-based compositional semantics (DCS), which has favorable linguistic,
statistical, and computational properties.  This enabled us to learn a
semantic parser from question-answer pairs where the intermediate logical form
(a DCS tree) is induced in an unsupervised manner.
Our final question-answering system was able to outperform current
state-of-the-art systems despite requiring no annotated logical forms.

%However, \citet{poon09semantic} built a question answering system induce clusters of distributionally similar
%lambda calculus expressions

% Open-domain question
There is currently a significant conceptual gap between our question-answering
system (which are natural language interfaces to databases) and
open-domain question-answering systems.  The former focuses
on understanding a question compositionally and computing the answer
compositionally, while the latter focuses on
retrieving and ranking answers from a large unstructured textual corpus.
The former has depth; the latter has breadth.
Developing methods that can both model the semantic richness of language
and scale up to an open-domain setting remains an open challenge.

We believe that it is possible to push our approach in the open-domain direction.  Neither
DCS nor the learning algorithm is tied to having a clean rigid database, which
could instead be a database generated from a noisy information extraction process.  The key is to
drive the learning with the desired behavior, the question-answer
pairs.  The latent variable is the logical form or program, which just tries to compute the
desired answer by piecing together whatever information is available.  Of course, there are many open challenges ahead,
but with the proper combination of linguistic, statistical, and computational insight, we
hope to eventually build systems with both breadth and depth.

\paragraph{Acknowledgments}
We thank Luke Zettlemoyer and Tom Kwiatkowski for providing us with data and answering questions.
The first author was supported by an NSF Graduate Research Fellowship.

\bibliographystyle{plainnat}
\bibliography{arxiv}

\end{document}